\documentclass[10pt]{article} 

\usepackage[accepted]{rlj}           

\usepackage{amssymb}            
\usepackage{mathtools}          
\usepackage{mathrsfs}           
\usepackage{graphicx}           
\usepackage{subcaption}         
\usepackage[space]{grffile}     
\usepackage{url}                
\usepackage{lipsum}             
\usepackage{booktabs}

\title{Uncovering RL Integration in SSL Loss: Objective-Specific Implications for Data-Efficient RL}

\setrunningtitle{Uncovering RL Integration in SSL Loss: Objective-Specific Implications for Data-Efficient RL}

\author{Ömer Veysel Çağatan\textsuperscript{1,2}, Barış Akgün\textsuperscript{1,2}}

\emails{\ \{ocagatan19,baakgun\}@ku.edu.tr}

\affiliations{
$^{1}$\textbf{Department of Computer Engineering, Koç University, Turkey}\\
$^{2}$\textbf{KUIS AI Center, Koç University, Turkey}\\

\par 

}

\summary{
This paper presents a systematic analysis of the role of self-supervised learning (SSL) objectives and their modifications in data-efficient reinforcement learning. We investigate previously undocumented modifications in the Self-Predictive Representations (SPR)~\citep{Schwarzer2020DataEfficientRL} framework that significantly impact agent performance. We demonstrate that feature decorrelation-based SSL objectives can achieve comparable performance without relying on domain-specific modifications and show that the impact of these modifications persists even in more advanced models.

By conducting extensive experiments on the Atari 100k benchmark and DeepMind Control Suite, we provide insights into how different SSL objectives and their modifications affect learning efficiency across diverse environments. Our findings reveal that the choice and adaptation of SSL objectives play a crucial role in achieving data efficiency in self-predictive reinforcement learning, with implications for the design of future algorithms in this space.}

\contribution{
    We demonstrate that previously undocumented SSL modifications in SPR~\citep{Schwarzer2020DataEfficientRL} - terminal state masking and prioritized replay weighting - are crucial for performance, with their removal leading to an 18\% decrease in IQM score on Atari 100k
    }
    {
    These modifications were silently adopted by subsequent work~\citep{DOro2023SampleEfficientRL,Nikishin2022ThePB, Schwarzer2023BiggerBF} and their impact was not previously analyzed.
    }

\contribution{
    We show that the Barlow Twins SSL objective~\citep{Zbontar2021BarlowTS} can come within 5\% of SPR's performance without using domain-specific modifications, and VICReg~\citep{Bardes2021VICRegVR} can match PlayVirtual's~\citep{yu2021playvirtual} performance in continuous control tasks.
    }
    {
    Prior work on SSL in reinforcement learning relied heavily on problem-specific modifications to achieve strong performance~\citep{Schwarzer2020DataEfficientRL,DOro2023SampleEfficientRL,Schwarzer2023BiggerBF}.
    }

\contribution{
    We establish that the impact of SSL modifications remains proportionally consistent in more sophisticated models, with unmodified versions of SR-SPR and BBF showing similar relative performance degradation despite having base IQM scores 3x and 2x higher than SPR, respectively.
    }
    {
    Previous work on SR-SPR~\citep{DOro2023SampleEfficientRL,Nikishin2022ThePB} and BBF~\citep{Schwarzer2023BiggerBF} did not investigate the role of these modifications in their improved performance.
    }

\begin{document}

\makeCover  
\maketitle  

\begin{abstract}
In this study, we investigate the effect of SSL objective modifications within the SPR framework, focusing on specific adjustments such as terminal state masking and prioritized replay weighting, which were not explicitly addressed in the original design. While these modifications are specific to RL, they are not universally applicable across all RL algorithms. Therefore, we aim to assess their impact on performance and explore other SSL objectives that do not accommodate these adjustments, like Barlow Twins and VICReg. We evaluate six SPR variants on the Atari 100k benchmark, including versions both with and without these modifications. Additionally, we test the performance of these objectives on the DeepMind Control Suite, where such modifications are absent. Our findings reveal that incorporating specific SSL modifications within SPR significantly enhances performance, and this influence extends to subsequent frameworks like SR-SPR and BBF, highlighting the critical importance of SSL objective selection and related adaptations in achieving data efficiency in self-predictive reinforcement learning.
\end{abstract}
\section{Introduction}
\label{intro}


Self-supervised learning (SSL) has become increasingly popular in data-efficient reinforcement learning (RL) due to its benefits in enhancing both efficiency and performance~\citep{Schwarzer2023BiggerBF, ye2021mastering, Hafner2023MasteringDD, Srinivas2020CURLCU, tomar2021learningrepresentationspixelbasedcontrol, li2023doesselfsupervisedlearningreally, cagatan2023barlowrl}. However, the application of SSL methods is often problem/domain-specific to maximize the performance of the RL agents.
Although this approach is rational given the 
nature of these methods, it raises questions about generalization and transferability.

One of the key challenges in Deep RL is understanding the factors driving performance improvements, whether through hyperparameter tuning or novel algorithmic approaches \citep{obandoceron2024consistencyhyperparameterselectionvaluebased}. The lack of transparency in hyperparameter selection often causes issues while algorithmic innovations are usually well-documented. However, our study of different SSL objectives within the Self-Predictive Representations (SPR) framework \citep{Schwarzer2020DataEfficientRL} revealed that the SSL loss used in SPR differs from what is described in the original publication and its following works ~\citep{Nikishin2022ThePB,DOro2023SampleEfficientRL,Schwarzer2023BiggerBF} built upon it. This motivated us to investigate the effects of the undocumented modifications and further evaluate additional SSL objectives.

Unlike conventional SSL methods in RL, which often follow vision pretraining approaches \citep{Chen2020ASF} and directly combine SSL and RL losses \citep{Srinivas2020CURLCU}, SPR modifies the SSL loss before integrating it with the RL objective. To further clarify, SPR employs the BYOL/SimSiam \citep{Grill2020BootstrapYO, Chen2020ExploringSS} auxiliary objective and incorporates two algorithm-specific adjustments to the SSL objective: (i) masking SSL loss with a boolean non-terminal state matrix and (ii) applying prioritized replay weighting to the batch loss. Consequently, this poses an essential question: How do these modifications affect the base performance of SSL objectives in the RL agent, and can they be effectively applied to other SSL techniques in the RL domain? In addition, could this be a recurring phenomenon across the following models ~\citep{Nikishin2022ThePB,Schwarzer2023BiggerBF} that adopt SPR as their baseline?


Concurrently, a plethora of novel self-supervised representation learning objectives has emerged \citep{Zbontar2021BarlowTS, Bardes2021VICRegVR, Ozsoy2022SelfSupervisedLW, caron2021unsupervised}, demonstrating 
performance improvements beyond image pretraining \citep{lee2023importance, goulão2023pretraining, zhou2022noncontrastivelearningmeetslanguageimage, çağatan2024unseeunsupervisednoncontrastivesentence}. These objectives, based on feature decorrelation, do not inherently support the modifications used in SPR because the loss is computed along the feature dimension instead of the batch dimension, which we detail in Section~\ref{app:sprs}.

This divergence raises another important question: How do these alternative objectives perform relative to the original SPR without SSL modifications? This inquiry is particularly significant because the information required to modify SSL objectives may not always be available in the environment. Understanding the performance of these unmodified objectives could provide valuable insights into the generalizability and robustness of different SSL approaches in RL contexts. Towards this end, we incorporate Barlow Twins and VICReg SSL objectives within SPR.

In essence, we frame our investigation around the following questions:
\begin{enumerate}
    \item \textbf{How do these modifications affect the performance of SPR, and do their impacts extend to SPR-based models such as SR-SPR and BBF? Additionally, how do these alternative objectives compare to the original SPR when no SSL modifications are implemented?} 

    
    Our findings reveal that modifications to SSL significantly affect SPR performance, leading to an 18\% decrease in IQM when these modifications are removed. Additionally, SR-SPR and BBF exhibit a similar decline in performance. Among these modifications, prioritized replay weighting stands out as the most influential. Notably, Barlow Twins achieves results comparable to those of the original SPR, while VICReg's performance aligns with that of prioritized replay weighting. This indicates that these problem-specific modifications can be mitigated by employing alternative SSL objectives. 
    Overall, our results underscore the importance of SSL modifications in SPR, which persist in strong models that utilize SPR

    \item \textbf{How effectively do these SSL objectives perform in an environment in which SPR modifications are not applicable?}

    To address this, we examine VICReg, Barlow Twins, and SPR (BYOL/SimSiam) within the DeepMind Control Suite, where the popular SAC agent does not utilize prioritized replay weighting and the environment lacks a terminal state. Unlike in the Atari 100k benchmark, our results show VICReg as the top performer, even outpacing PlayVirtual, a more sophisticated variant of SPR. Meanwhile, SPR and Barlow Twins exhibit comparable performance levels. These findings highlight that algorithms tailored for specific domains may not consistently excel across different problem sets. Therefore, transferability should be a key factor in the design of new Deep RL algorithms. 

\end{enumerate}

\begin{figure*}[t]
  \centering
  \includegraphics[width=1\textwidth]{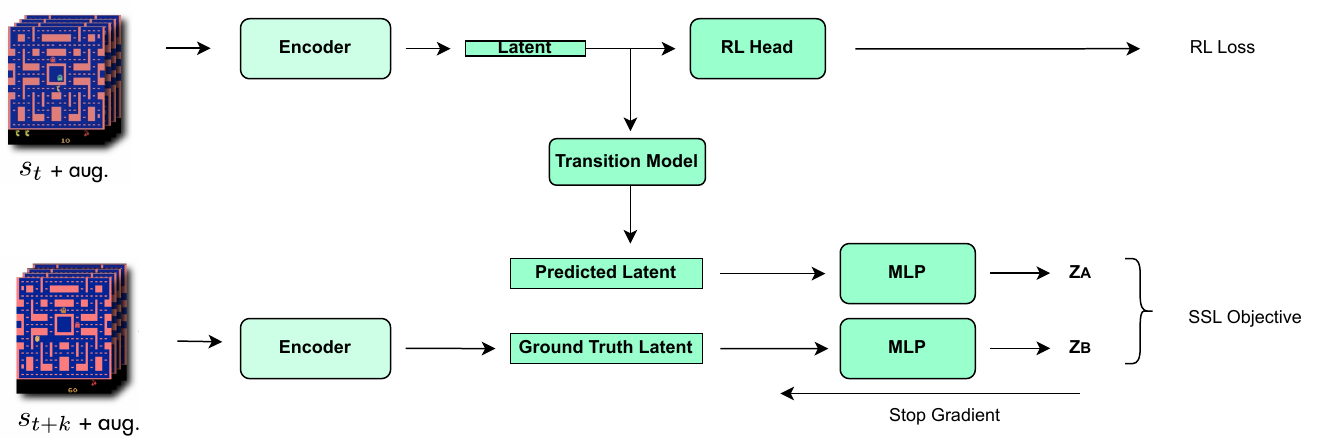} 
  \caption{General flow diagram of SPR based methods. 
  An encoder is used to create representations used for reinforcement learning and predicting future representations via a transition model and ground truth representations are created by the same encoder. MLPs differ when the predictor layer is used as in the case of BYOL/SimSiam. While we show the $k^{th}$ step here, the actual loss computation covers steps 1 to $K$. 
  The SSL objective and RL loss changes between specific methods.}
  \label{fig:spr}
\end{figure*}

\section{Related Work}\label{app:relatedwork}

~\citet{tomar2021learningrepresentationspixelbasedcontrol} tackles a more challenging setting for representation learning within RL with background distractors, using a simple baseline approach that avoids metric-based learning, data augmentations, world-model learning, and contrastive learning. They analyze why previous methods may fail or perform similarly to the baseline in this tougher scenario and stress the importance of detailed benchmarks based on reward density, planning horizon, and task-irrelevant components. They propose new metrics for evaluating algorithms and advocate for a data-centric approach to better apply RL to real-world tasks.

~\citet{li2023doesselfsupervisedlearningreally} explore whether SSL can enhance online RL from pixel data. By extending the contrastive reinforcement learning framework~\citep{Srinivas2020CURLCU} to jointly optimize SSL and RL losses, and experimenting with various SSL losses, they find that the current SSL approaches offer no significant improvement over baselines that use image augmentation alone, given the same data and augmentation. Even after evolutionary searches for optimal SSL loss combinations, these methods do not outperform carefully designed image augmentations. Their evaluation across various environments, including real-world robots, reveals that no single SSL loss or augmentation method consistently excels.

\subsection{Data Efficient RL in Atari 100k}
The introduction of the Atari 100k benchmark~\citep{Kaiser2019ModelBasedRL} has expedited the advancement of sample-efficient reinforcement learning algorithms. Model-based approach, SimPLe~\citep{Kaiser2019ModelBasedRL}, outperformed Rainbow DQN~\citep{Hessel2017RainbowCI}, showcasing superior performance. Building on Rainbow's framework, ~\citet{Hasselt2019WhenTU} enhanced its efficacy through minor hyperparameter adjustments, resulting in Data-Efficient Rainbow (DER), which achieved a higher score compared to SimPLe.

DrQ~\citep{Kostrikov2020ImageAI} employs a multi-augmentation strategy to regularize the value function during training of both Soft Actor-Critic~\citep{Haarnoja2018SoftAO} and Deep Q-Network~\citep{Mnih2015HumanlevelCT}. This approach effectively reduces overfitting and enhances training efficiency, leading to performance improvements for both algorithm

Several prevalent methods adopt the Atari 100k dataset, and these can be classified as follows: Model-Based~\citep{Hafner2023MasteringDD,Robine2023TransformerbasedWM,Micheli2022TransformersAS,Ayton2021WidthBasedPA,robine2021smaller}, Pretraining~\citep{Goulo2022PretrainingTV,Schwarzer2021PretrainingRF,Lee2023EnhancingGA,Liu2021APSAP}, Model-Free~\citep{Schwarzer2023BiggerBF,Huang2022AcceleratingRL,Nikishin2022ThePB,Cetin2022StabilizingOD,Lee2023EnhancingGA,Liang2022ReducingVI}

\subsection{Representation Learning in Atari 100k}

\citet{Cetin2022HyperbolicDR} presents a deep reinforcement learning method using hyperbolic space for latent representations. Their innovative approach tackles optimization challenges in existing hyperbolic deep learning, ensuring stable end-to-end learning through deep hyperbolic representations.

\citet{Huang2022AcceleratingRL} proposes a Multiview Markov Decision Process (MMDP) with View-Consistent Dynamics (VCD), a method that enhances traditional MDPs by considering multiple state perspectives. VCD trains a latent space dynamics model for consistent state representations, achieved through data augmentation.

\citet{Srinivas2020CURLCU} incorporate the InfoNCE~\citep{oord2019representation} as an auxiliary component within DER.~\citet{cagatan2023barlowrl} uses  Barlow Twins~\citep{Zbontar2021BarlowTS} instead of a contrastive objective to further improve results. This integration serves to enhance the learning process. SPR~\citep{Schwarzer2020DataEfficientRL} outperforms all previous model-free approaches by predicting its latent state representations multiple steps into the future with BYOL~\citep{Grill2020BootstrapYO}.

PlayVirtual~\citep{yu2021playvirtual} introduces a novel transition model as an alternative to the simplistic module in SPR. The methodology enriches actual trajectories by incorporating a multitude of cycle-consistent virtual trajectories. These virtual trajectories, generated using both forward and backward dynamics models, collectively form a closed 'trajectory cycle.' The crucial aspect is ensuring the consistency of this cycle, validating the projected states against real states and actions. This approach significantly improves data efficiency by acquiring robust feature representations with reduced reliance on real-world experiences. This method proves particularly advantageous for tasks where obtaining real-world data is costly or challenging.

\section{SPR}\label{app:spr}
SPR is a performant data-efficient agent and a baseline of many other performant agents~\citep{Schwarzer2023BiggerBF, Nikishin2022ThePB, DOro2023SampleEfficientRL,yu2021playvirtual} and its general architecture is depicted in Figure \ref{fig:spr}. The approach trains an agent by having it predict the latent state based on the current state. It encodes the present state, forecasts the latent representation of the next state using a transition model, and calculates loss by measuring the mean squared error between normalized embeddings. Additionally, SPR adjusts its loss through terminal masking and prioritized replay weighting. These two modifications inject RL-specific information into the auxiliary self-supervised learning task. While the utilization of these ideas is not explicitly mentioned by \citet{Schwarzer2020DataEfficientRL}, it is possible that these techniques were considered self-evident and consequently were included in their implementation ~\citep{sprGithub}. We mention them here so as to be able to better differentiate between SPR and other SPR variants.

SSL loss matrix in SPR denoted as $L$, encompasses negative cosine similarities between predicted latent representations and ground truth latent representations, with dimensions of $B \times (K + 1)$, where $B$ is the batch size, and $K$ is the prediction horizon with 1 coming from the current observation. The batch of interactions is drawn from the replay buffer, and their terminal status is known. The terminal mask matrix, $M$, is composed of 0s  and 1s denoting terminal and non-terminal states. The process involves updating $L$ through a Hadamard product with $M$, denoted as $L \circ M$, effectively modifying the loss matrix.

The loss matrix is divided into two components: SPR loss and Model SPR loss. 
SPR loss is between the latent representations of the augmented views of the present state. Model SPR loss is between the latent representations of the augmented views of the future states and the predicted future latent representations, generated by the transition model. Model SPR is averaged across the temporal dimension and as a result, both components have $N \times 1$ dimensionality. 

The loss of each transition is multiplied by the prioritized replay weight, determined by the temporal difference errors. Then the final loss is computed as the weighted sum of the average SPR loss and half the average of the Model SPR loss across a batch as follows:
\begin{equation}\label{eq:spr}
\mathcal{L}_{SPR} =  \frac{1}{N} \sum_{i=1}^{N} \omega_i (\lambda\text{SPR}_i + \gamma\text{Model SPR}_i)
\end{equation}
where $N$ is the batch size, $\omega_i$ is the priority weight ($\sum_i \omega_i = 1$), and $i$ indexes individual transitions, where $\lambda$,$\gamma$ are hyperparameters.

\section{SPR-*} \label{app:sprs}
Despite variations in SSL objectives and RL algorithms across different benchmarks, the architecture remains largely consistent, as depicted in Figure \ref{fig:spr}. SPR employs a BYOL~\citep{Grill2020BootstrapYO} objective with a momentum of 1, essentially adopting the SimSiam~\citep{Chen2020ExploringSS} approach. The primary architectural distinction lies in the inclusion of an extra predictor layer in the online MLP of BYOL or SimSiam to prevent collapse, a feature omitted in the original Barlow Twins and VICReg formulations as their objectives inherently mitigate the risk of collapse.

\textbf{SPR-Nakeds}\quad While SPR demonstrates considerable efficacy, the fundamental question remains unanswered—what is the impact of pure self-supervised learning and potential adaptations leading to SPR? Consequently, we introduce SPR-Naked, representing pure SSL. To assess the effects of prioritized replay weighting and terminal masking, we further establish SPR-Naked+Prio and SPR-Naked+Non, respectively.

In addition to the original SPR and its naked versions, we implement two additional types of agents with different SSL objectives.

\textbf{SPR-Barlow} \quad
To extend the Barlow Twins to future predictions, we compute individual cross-correlation matrices for both the current and predicted latent representations at each time step. This results in a total of $K+1$ matrices, each with dimensions $d \times d$, where $d$ denotes the embedding dimension within a single batch. Subsequently, we calculate the loss for each matrix and average the results. To make it easier to compare, we can define $\overline{\text{SPR}}$ Loss and $\overline{\text{Model SPR}}$ Loss analogously to their SPR counterparts, where the first is about the current state and the latter is about the future states. The final loss is then; 
\begin{equation}\label{eq:spr-barlow}
\mathcal{L}_{SPR-Barlow} = \overline{\text{SPR}} + \frac{1}{K} \sum_{k=1}^{K}\overline{\text{Model SPR}}_k
\end{equation}
where $K$ is the number of predicted future observations.

\textbf{SPR-VICRegs}\quad We employ a parallel procedure as in Barlow Twins for VICReg. We introduce two variations of VICReg-High and VICReg-Low, featuring high or low covariance weights in the VICReg loss (Equation~\ref{eq:vicreg}) while maintaining consistency in other hyperparameters. The primary objective is to observe the impact of feature decorrelation without inducing model collapse.

\textbf{Why not employ replay weighting and terminal state masking in Barlow/VICReg?}
The key limitation preventing the use of replay weighting or terminal masking in feature decorrelation-based methods lies in their reliance on covariance regularization. These methods employ either a cross-correlation matrix or a covariance matrix, both with dimensions matching the feature dimension. This structure prohibits applying the weighting of a feature dimension matrix using a batch dimension matrix. Consequently, these methods produce a unified loss for the entire batch, unlike approaches such as BYOL or SimSiam, which generate losses on a per-sample basis.

\begin{table}[t]
    \centering
    \begin{minipage}{0.5\textwidth}
        \centering
        \footnotesize
        \setlength{\tabcolsep}{4pt}  
        \begin{tabular}{@{}l|l|l|l|l@{}}  
              &Median & IQM & Mean & Opt.Gap \\
             \hline
             Barlow & \textbf{0.324} & \textbf{0.320} & \textbf{0.605} & \textbf{0.593}\\
             VICReg & 0.281 & 0.289 & 0.600 & 0.610 \\
             VICReg+Non & 0.221 & 0.279 & 0.554 & 0.617 \\ 
             Barlow+Non & -0.009 & -0.011 & -0.171 & 1.171 \\ 
             ZeroJump & 0.270 & 0.262 & 0.528 & 0.636
        \end{tabular}
        \caption{Human-normalized aggregate metrics in Atari 100k. Scores were collected from 10 random runs.}
        \label{tab:nonterm-noncontr}
    \end{minipage}%
    \hspace{0.01\textwidth}  
    \begin{minipage}{0.35\textwidth}
        \centering
        \footnotesize
        \setlength{\tabcolsep}{4pt}
        \begin{tabular}{@{}l|l|l|l|l@{}}
              &Median & IQM & Mean & Opt.Gap \\
             \hline
             Stop-Grad & \textbf{0.271} & \textbf{0.303} & \textbf{0.615} & \textbf{0.577}\\
             No Stop-Grad & 0.266 & 0.282 & 0.595 & 0.611 \\ 
        \end{tabular}
        \caption{Human-normalized aggregate metrics in Atari 100k by VICReg-High. Scores, collected from 10 random runs to assess the efficacy of including stop-gradient.}
        \label{tab:stop_grad}
    \end{minipage}
\end{table}

\textbf{Why use stop-gradient in Barlow/VICReg?} Barlow Twins and VICReg effectively prevent collapse without resorting to symmetry-breaking architectural techniques such as predictor layers or stop-gradient mechanisms. While not strictly necessary in this scenario, we choose to include a stop-gradient 
due to its empirically observed performance improvement, as depicted in Table \ref{tab:stop_grad}. A more grounded reason stems from the architectural asymmetry introduced by the transition model. In the absence of a stop-gradient, gradients from the encoder's upper branch flow through the transition model, whereas gradients from the lower branch directly influence the encoder. This asymmetry can potentially lead to suboptimal encoder updates. Despite collapse avoidance in both cases, the inclusion of a stop-gradient is maintained for its superior performance outcomes.

\textbf{Why not other objectives?}
Even though there are newly proposed SSL objectives~\citep{pmlr-v235-silva24c,pmlr-v235-zhang24bi,weng2024modulatespectrumselfsupervisedlearning}, it is impractical to include all objectives in experiments due to limited computational resources and the need to prioritize rigorous evaluation to draw precise conclusions. However, we attempt to cover the two main families of SSL methods within SPR. The first is self-distillation, represented by BYOL~\citep{Grill2020BootstrapYO} or SimSiam~\citep{Chen2020ExploringSS}, which are already incorporated into SPR. The second family includes canonical correlation methods, such as VICReg and Barlow. Another category is Deep Metric Learning, which includes contrastive learning variants~\citep{balestriero2023cookbook}. However, we do not separately test contrastive objectives, as they have already been shown to be ineffective in SPR~\citep{Schwarzer2020DataEfficientRL}. 

\textbf{Removing Features with Masking} We discussed why post-loss-calculation modifications cannot be applied to objectives that involve components in the feature dimension rather than the batch dimension. However, non-terminal masking can be employed to exclude samples from the batch before calculating the SSL loss. Thus, we masked features during the training of the SPR-VICReg and SPR-Barlow agents, leading to unexpected results. As shown in Table \ref{tab:nonterm-noncontr}, the SPR-Barlow agent performed even worse than the random agent. A likely explanation is that the Barlow Twins' objective relies on batch normalization to compute the cross-covariance matrix. Since masking causes the batch size to vary dynamically, the batch statistics become inconsistent, adversely affecting the batch normalization process. However, this degradation is not observed to the same extent in the SPR-VICReg agent, as the VICReg objective does not rely on batch normalization.

\textbf{Continuous Control Formulation} Although SPR is created specifically for discrete control, delving into the impact of SSL objectives solely within discrete control domains doesn't provide a comprehensive understanding. This is why we adopt a parallel setup to that of PlayVirtual~\citep{yu2021playvirtual}, where they establish an SPR-like scheme referred to as SPR\textsuperscript{\textdagger} as a baseline for continuous control. They utilize the soft actor-critic algorithm~\citep{Haarnoja2018SoftAO}, instead of Q-learning due to the continuous nature of the actions. They do not use terminal state masking (since terminal states for control problems are target states) and prioritized replay weighting (since they use a uniform buffer). This shows the importance of generally applicable auxiliary tasks for data-efficient RL.


We evaluate PlayVirtual and SPR\textsuperscript{\textdagger} from scratch since we were not able to replicate \citet{yu2021playvirtual}'s results, potentially due to different benchmark versions. Furthermore, we assess the performance of VICReg-High and Barlow Twins within the SPR\textsuperscript{\textdagger} configuration. We exclude VICReg-Low in this setting due to the minimal performance difference observed in Atari.

Finally, we explore the potential impact of incorporating the predictor network into Barlow Twins and VICReg, even though they inherently do not need it to prevent dimension collapse. Although the addition of a predictor network is novel in Barlow Twins, VICReg becomes similar to the SPR with this addition like SPR with variance-covariance regularization. The decision to refrain from conducting similar experiments in Atari stems from the substantially higher experimental costs, which are at least 10 times greater than those in the control setting.

\section{Evaluation Setup}\label{app:eval}

\subsection{Benchmarking: Rliable Framework}
\citet{Agarwal2021DeepRL} discusses the limitations of using mean and median scores as singular estimates in RL benchmarks and highlights the disparities between conventional single-point estimates and the broader interval estimates, emphasizing the potential ramifications for benchmark dependability and interpretation. 
In alignment with their suggestions, we provide a succinct overview of human-normalized scores, furnished with stratified bootstrap confidence intervals, in Figures \ref{fig:disc} and \ref{fig:cont}.

\subsection{Atari 100k}
We assess the SPR framework in a reduced-sample Atari setting, called the Atari 100k benchmark~\citep{Kaiser2019ModelBasedRL}. In this setting, the training dataset comprises 100,000 environment steps, which is equivalent to about 400,000 frames or slightly under two hours of equivalent human experience. This contrasts with the conventional benchmark of 50,000,000 environment steps, corresponding to approximately 39 days of accumulated experience.

The main metric for this setting, widely acknowledged for assessing performance in the Atari 100k context, is the human-normalized score. This measure is mathematically defined as in equation \ref{eq:calc}, where the random score represents outcomes from a random policy, while the human score comes from human player performance~\citep{Wang2015DuelingNA}.
 \begin{equation} \label{eq:calc}
\frac{score_{\text{agent}} - score_{\text{random}}}{score_{\text{human}} - score_{\text{random}}}     
 \end{equation}

\subsection{Deep Mind Control Suite}
In the Deep Mind Control Suite~\citep{tassa2018deepmind}, the agent is configured to function solely based on pixel inputs. This choice is justified by several reasons: the environments involved offer a reasonably challenging and diverse array of tasks, the sample efficiency of model-free reinforcement learning algorithms is notably low when operating directly from pixels in these benchmarks and the performance on the DM control suite is comparable to the context of robot learning in real-world benchmarks.

We use the following six environments ~\citep{yarats2020improving} for benchmarking: ball-in-cup, finger-spin, reacher-easy, cheetah-run, walker-walk and cartpole-swingup, for 100k steps each.

\section{Results and Discussion}

\subsection{Atari 100k}
We mainly investigate the following new SPR models, along with the original SPR: (i) SPR-Naked, featuring no modifications, (ii) SPR-Naked+Non, incorporating terminal masking, (iii) SPR-Naked+Prio, integrating prioritized replay weighting, (iv) SPR-Barlow, (v) SPR-VICReg-High, characterized by a high covariance weight, and (vi) SPR-VICReg-Low, characterized by a low covariance weight. Moreover, we discuss SR-SPR and BBF with their no modification versions.

\begin{figure*}[t]
  \centering
  \includegraphics[width=1\textwidth]{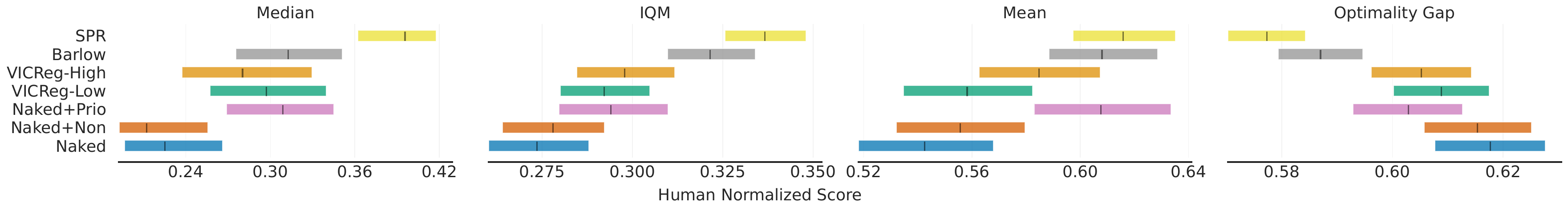}
  \caption{Mean, median, interquartile mean human normalized scores and optimality gap (lower is better) computed with stratified bootstrap confidence intervals in Atari 100k. 50 runs for SPR-Barlow, SPR-VICReg-High, SPR-VICReg-Low, SPR-Naked+Prio, SPR-Naked+Non,SPR-Naked, 100 runs for SPR from ~\citep{Agarwal2021DeepRL}.}
  \label{fig:disc}
\end{figure*}

Figure~\ref{fig:disc} shows the performance of the seven agents
in the Atari 100k benchmark, calculated using the rliable framework~\citep{Agarwal2021DeepRL}. The individual game performances are given in Appx.~\ref{app:full-results} and we describe the evaluation setup in Section ~\ref{app:eval}.

\textbf{SPR and SSL Modifications}. The original SPR agent performs the best (top row of Fig.~\ref{fig:disc}). The modifications to the SPR's SSL objective (see Section ~\ref{app:spr}) have significant impact on the performance but they are not mentioned in the relevant papers (SPR~\citep{Schwarzer2020DataEfficientRL}, SR-SPR~\citep{DOro2023SampleEfficientRL,Nikishin2022ThePB}, or BBF~\citep{Schwarzer2023BiggerBF}). The no modifications version, SPR-Naked, performs the worst with a nearly 20\% performance drop based on the IQM score (last row of Fig.~\ref{fig:disc}). 
This is crucial because such modifications may not be suitable
for all problem domains, which limits their transferability and generalizability. On the other hand, the role of terminal masking and prioritized replay weighting in SPR is especially interesting, as they help boost performance in situations where pure representation learning struggles.

Incorporating prioritized replay weights has a positive effect on SPR (5$^{th}$ row of Fig.~\ref{fig:disc}). These weights act as markers for Bellman errors that mirror the agent's Q-value approximation performance on particular transitions. Introducing these weights into the representation loss intensifies the emphasis on refining representations that the agent struggles with. 

Empirically, terminal state masking shows negligible positive effects, unlike replay weighting, 
(6$^{th}$ row of Fig.~\ref{fig:disc}). The limited impact of masking might be attributed to the episode lengths, where the agent encounters many regular states but only a single terminal state. The SSL loss may be primarily influenced by intermediate states, which could reduce the effectiveness of masking in these scenarios.

On the other hand, there is a clear synergy between these modifications within SPR. Masking terminal states might be advantageous when agents encounter frequent failures during the initial stages of training or due to the nature of the games.
In such cases, terminal states may dominate the replay buffer, which could introduce biased representations that become challenging to correct later on and make it harder for the agent to adapt and improve as it progresses

\begin{figure*}[t]
  \centering
  \includegraphics[width=1\textwidth]{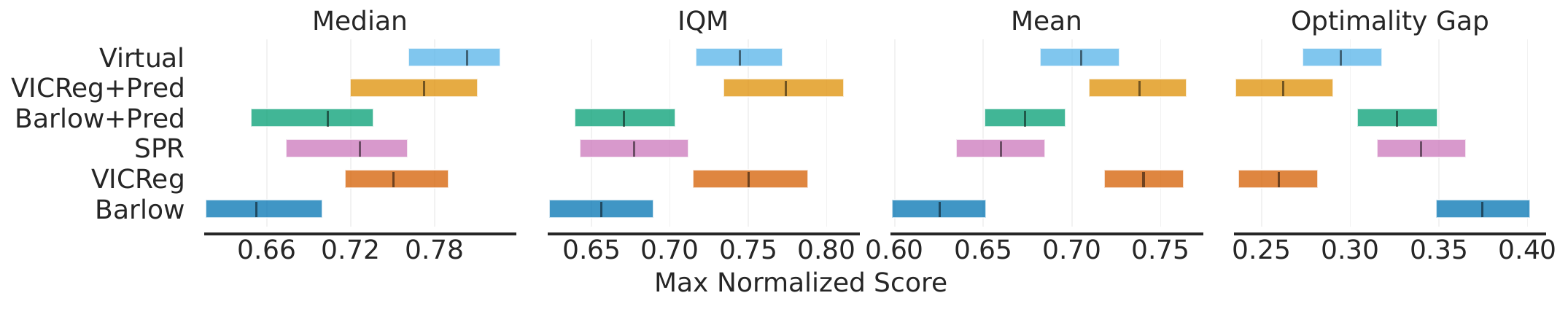}
  \caption{Mean, median, interquartile mean max normalized scores and optimality gap (lower is better) computed with stratified bootstrap confidence intervals in Deep Mind Control Suite 100k, 10 runs for all agents.}
  \label{fig:cont}
\end{figure*}

\textbf{SPR-Barlow}. The performance of the Barlow Twins agent is close to the SPR's (2$^{nd}$ row of Fig.~\ref{fig:disc}), with only a 5\% difference, where as SPR-Naked has a 20\% gap.
As described in Section ~\ref{app:sprs}, modifications related to SSL do not directly apply to Barlow Twins, VICReg, or any other method of regularization in the feature dimension. As such, performing similarly to a method with RL-specific modifications suggests that Barlow Twins has the potential to serve as a substitute, indicating its promise as a versatile SSL objective for data-efficient RL. 


The performance gap between SPR-naked and the feature decorrelation methods (Barlow and VICReg) in this context is somewhat surprising
since BYOL or Simsiam outperforms them in image classification. In vision pretraining, the goal is to obtain embeddings with well-defined clusters based on the training corpora, enhancing classification performance, where feature decorrelation may be of hindrance. In RL, it is important to differentiate between states (good, bad, or promising if they have not been explored yet) that may not be too different in the image space. As such, methods that emphasize the use of the entire embedding space potentially have a better chance of state separation.


To test this, we evaluate the rank~\citep{kumar2021implicitunderparameterizationinhibitsdataefficient} of the advantage and value heads, as well as the output of the convolution head, which is shared by both the RL and SSL objectives. We evaluated multiple methods like Barlow Twins and VICReg, in addition to a variant without SSL loss. We found that the rank converges similarly across different games and even if they don't, this does not correlate with performance. We also measured dormant neurons~\citep{sokar2023dormantneuronphenomenondeep} and observed that the results were consistent with the rank findings. These evaluations are detailed in Appx.~\ref{app:rank}.

\textbf{SPR-VICRegs}. Initially, we used the default VICReg hyperparameters given in the original paper~\citep{Bardes2021VICRegVR}. Surprisingly, VICReg exhibits a 13\% lower performance (4$^{th}$ row of Fig.~\ref{fig:disc}) compared to SPR although it surpasses SPR-Naked. It also falls short of Barlow Twins. This outcome is not immediately evident given that it has a high similarity to the Barlow Twins' objective. One plausible explanation could be the presence of multiple loss components, possibly undermining covariance. To address this, we explore alternative hyperparameters, selecting the set with the highest covariance hyperparameter that avoids collapse and denoting it as SPR-VICReg-High, while the previous one is referred to as SPR-VICReg-Low. However, the performance only marginally increases by 2\% (3$^{rd}$ row of Fig.~\ref{fig:disc}), lacking behind Barlow Twins once again. The underlying reasons for this performance gap remain subject to further exploration. Nonetheless, it still showcases the effectiveness of feature decorrelation-based objectives since both types outperform SPR-Naked.

\textbf{BBF and SR-SPR}. It could be argued that modifications to SPR significantly influence performance, particularly due to its relatively low score on Atari 100k, where such changes may have an amplified effect, whereas they might have a more limited impact on stronger models. BBF, the leading value-based agent achieving human-level results on Atari 100k, is built upon SR-SPR, a variant of SPR. Notably, both SR-SPR and BBF exhibit IQM values nearly 3x and 2x higher than SPR, respectively. Thus, their unmodified results will provide insight into whether modifications still play a significant role, even when the model is highly efficient and performing at a human level.

\begin{figure}[t]
  \centering
  \small
  \includegraphics[width=0.48\textwidth]{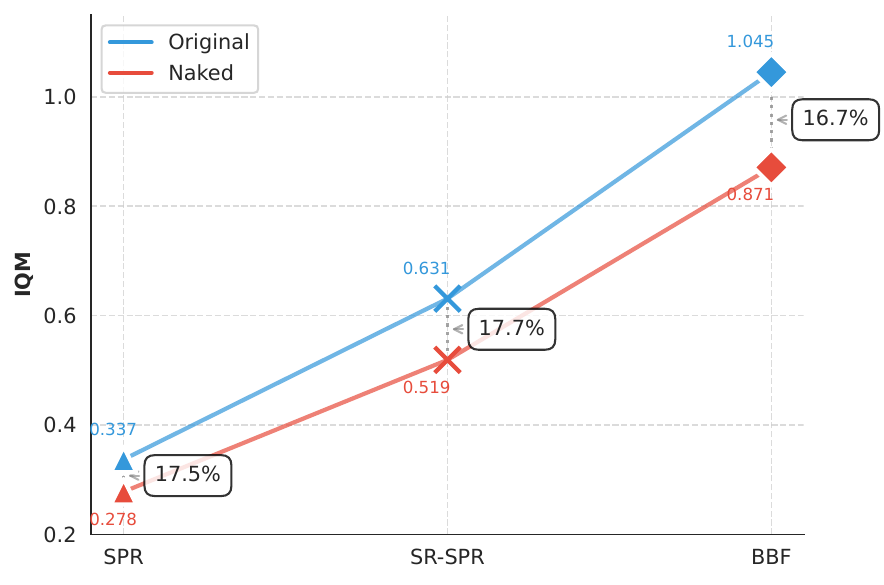} 
  \caption{Comparison of IQM performance for the SPR, SR-SPR, and BBF agents alongside their corresponding naked versions. Naked results of SR-SPR and BBF are averaged out across 10 different runs}
  \label{fig:scale}
\end{figure}

As shown in Figure ~\ref{fig:scale}, we observe that modifications result in a fairly consistent performance decline across all models. Due to computational constraints, we did not conduct experiments to determine which modifications have the greatest impact or whether certain SSL objectives could reduce the need for modifications. However, our findings further support and strengthen our earlier conclusions regarding the impact of modifications on SPR.

\subsection{DMControl}
We further evaluate the SSL objectives with the DMControl suite, described in Section~\ref{app:eval}) since this domain can provide additional insights into the efficacy of SSL objectives in RL. However, since there is no terminal state in this environment and a uniform replay buffer is used, modifications to the SPR loss are not feasible. As such, this evaluation will focus on the generalization of used objectives across domains without targeted optimization for specific problems. 

Moreover, SPR is not explicitly designed for continuous control. As such, we use a different set of agents modified for continuous control as described in Section~\ref{app:sprs} but keep the same SSL hyperparameters from the Atari benchmark. We pick SPR-VICReg-High due to its better performance over the lower covariance version. We additionally evaluate SPR-Barlow and SPR-Vicreg with an MLP layer as an additional predictor, reflecting \citet{Bardes2021VICRegVR}'s findings on the enhanced performance of BYOL with variance regularization. We build upon the PlayVirtual~\citep{yu2021playvirtual} methodology,  which is an SPR equipped with an improved transition model, and use it as our baseline.

We observe from Fig.~\ref{fig:cont} that the Barlow Twins objective exhibits the lowest performance, although it closely aligns with SPR, with IQM scores of 0.656, and 0.677 respectively. An interesting observation is that VICReg with an IQM of 0.75 is as good as PlayVirtual~\citep{yu2021playvirtual} with 0.744. This underscores the potential of SSL objectives in continuous control. While their impact is vital in discrete control as well, the overall effect, especially when considering the maximum score (representing human performance), is relatively modest. Nevertheless, a substantial improvement is evident in continuous control, even when compared to the highest achievable score. We also see that adding a predictor network has a minimal but positive impact on the IQM performances of both Barlow and VICReg.

\section{Conclusion}
Our study demonstrates the significant impact of SSL objective modifications within the SPR framework for reinforcement learning, particularly in data-efficient scenarios. We show that specific adjustments like terminal state masking and prioritized replay weighting substantially improve performance on the Atari 100k benchmark, with benefits extending to derivative frameworks such as SR-SPR and BBF. However, our experiments on the DeepMind Control Suite reveal that these enhancements are not universally applicable across all RL environments. Investigation of alternative SSL objectives (e.g., Barlow Twins, VICReg) further elucidates the nuanced relationship between objective choice and RL task characteristics. These findings emphasize the critical role of carefully tailored SSL objectives in achieving data efficiency in self-predictive reinforcement learning, highlighting the need for a context-sensitive approach to SSL modification in RL algorithm development. Our work provides valuable insights for researchers and practitioners seeking to optimize RL algorithms across diverse applications, potentially leading to more efficient and effective reinforcement learning systems.

\section*{Acknowledgements}
This work was funded by the KUIS AI Center at Koç University, Turkey.



\bibliography{main}
\bibliographystyle{rlj}

\beginSupplementaryMaterials

\section{Background}

\subsection{Barlow Twins}

The Barlow Twins~\citep{Zbontar2021BarlowTS} employs a symmetric network with twin branches, each processing a different augmented perspective of input data. It aims to minimize off-diagonal components and align diagonal elements of a cross-covariance matrix derived from the representations of these branches. The process involves generating two altered views ($Y^A$ and $Y^B$) using data augmentations, inputting them into a function $f_{\theta}$ to produce embeddings ($Z^{A}$ and $Z^{B}$).

The Barlow Twins loss is defined as:
\begin{equation}
\mathcal{L_{BT}} \triangleq  \underbrace{\sum_i  (1-\mathcal{C}_{ii})^2}_\text{invariance term}  + ~~\lambda \underbrace{\sum_{i}\sum_{j \neq i} {\mathcal{C}_{ij}}^2}_\text{redundancy reduction term}
\label{eq:lossBarlow}
\end{equation}
where $\lambda > 0$ balances the invariance (diagonal elements) and redundancy reduction (off-diagonal) in the loss function. $\mathcal{C}$ is the cross-correlation matrix from embedding outputs of identical networks in the batch. A matrix element is defined as:
\begin{equation}
\mathcal{C}_{ij} \triangleq \frac{
\sum_b z^A_{b,i} z^B_{b,j}}
{\sqrt{\sum_b {(z^A_{b,i})}^2} \sqrt{\sum_b {(z^B_{b,j})}^2}}
\label{eq:crosscorr}
\end{equation}

where $b$ represents the samples in the batch, and $i$ and $j$ represent dimension indices of the networks' output. Each dimension of the square covariance matrix, $\mathcal{C}$, is the same as the embedding dimension (output dimensionality of the networks). Its values range between -1 (indicating complete anti-correlation) and 1 (representing perfect correlation).

\subsection{VICReg}

VICReg~\citep{Bardes2021VICRegVR} is a method designed to tackle the challenge of collapse directly. It achieves this by introducing a straightforward regularization term that specifically targets the variance of the embeddings along each dimension independently. In addition to addressing the variance, VICReg includes a mechanism to diminish redundancy and ensure decorrelation among the embeddings, accomplished through covariance regularization.

The variance regularization term is a hinge function on the standard deviation of the embeddings along the batch dimension:
\begin{equation}
v(Z) = \frac{1}{d} \sum_{j=1}^d \max(0, \gamma - S(z^j, \epsilon))
\end{equation}
where $S$ is the regularized standard deviation defined by:
\begin{equation}
S(x; \epsilon) = \sqrt{\text{Var}(x) + \epsilon}
\end{equation}
Covariance matrix of $Z$ is defined as:
\begin{equation}
    C(Z) = \frac{1}{n-1} \sum_{i=1}^n (z_i - \bar{z})(z_i - \bar{z})^T
\end{equation}
where $\bar{z} = \frac{1}{n} \sum_{i=1}^n z_i$. Covariance regularization is defined as:
\begin{equation}
c(Z) = \frac{1}{d}\sum_{i}\sum_{j \neq i} {\mathcal{C}_{ij}}^2
\end{equation}
where $d$ is the feature dimension. The invariance criterion between $Z$ and $Z'$ is the mean-squared Euclidean distance between each pair of vectors, without any normalization.
\begin{equation}
s(Z, Z') = \frac{1}{n} \sum_{i=1}^n ||z_i - z'_i||^2
\end{equation}
The overall loss function is a weighted average of the invariance, variance, and covariance terms:
\begin{equation} \label{eq:vicreg}
l(Z, Z') = \alpha v(Z) + \beta c(Z) + \gamma s(Z, Z')
\end{equation} 
where $\alpha$, $\lambda$, and $\gamma$ hyper-parameters control the importance of each term in the loss.

VICReg is quite similar to Barlow Twins in terms of its loss formulation. However, instead of decorrelating the cross-correlation matrix directly, it regularizes the variance along each dimension of the representation, reduces correlation and minimizes the difference of embeddings. This prevents dimension collapse and also forces the two views to be encoded similarly. Additionally, reducing covariance encourages different dimensions of the representation to capture distinct features.

\section{Rank and Dormant Neuron}\label{app:rank}
\citet{kumar2021implicitunderparameterizationinhibitsdataefficient} introduced the concept of *effective rank* for representations, represented as \(srank_{\delta}(\phi)\), with \(\delta\) being a threshold parameter, set to 0.01 as per their study. They proposed that effective rank is linked to the expressivity of a network, where a decrease in effective rank implies an implicit under-parameterization. The study provides evidence indicating that bootstrapping is the primary factor contributing to the collapse of effective rank, which in turn degrades performance.

\begin{figure}[htbp]
    \centering
    \caption{Rank of the output from the penultimate layer of the value head, measured every 10,000 steps and averaged across 10 different runs for every game.}
    \begin{subfigure}[b]{0.2\textwidth}
        \centering
        \includegraphics[width=\textwidth]{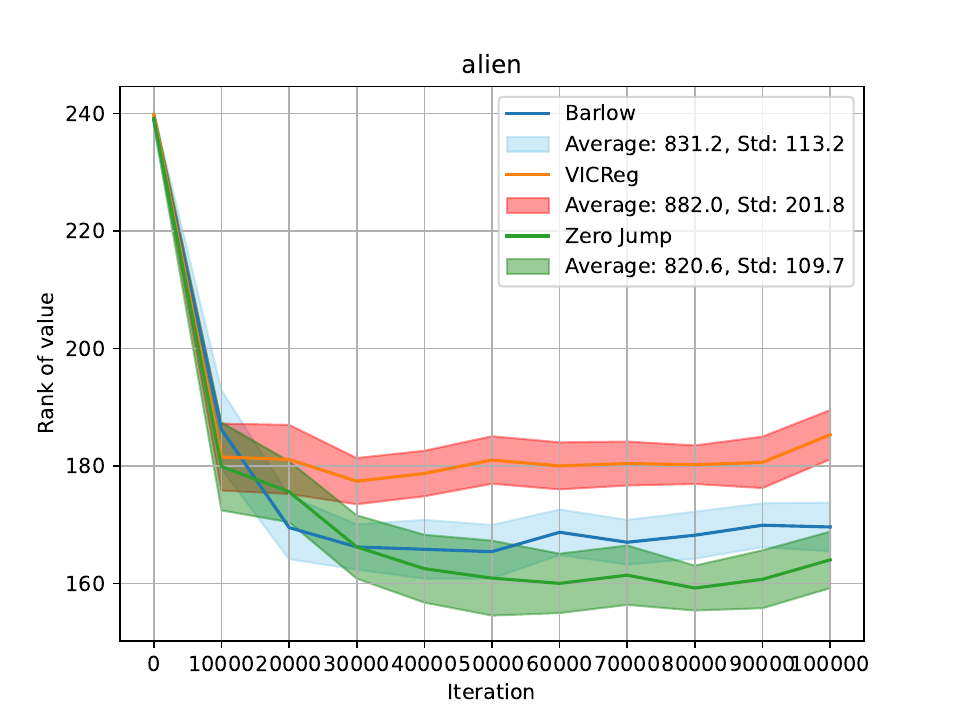}
        \label{fig:value-sub1}
    \end{subfigure}
    \begin{subfigure}[b]{0.2\textwidth}
        \centering
        \includegraphics[width=\textwidth]{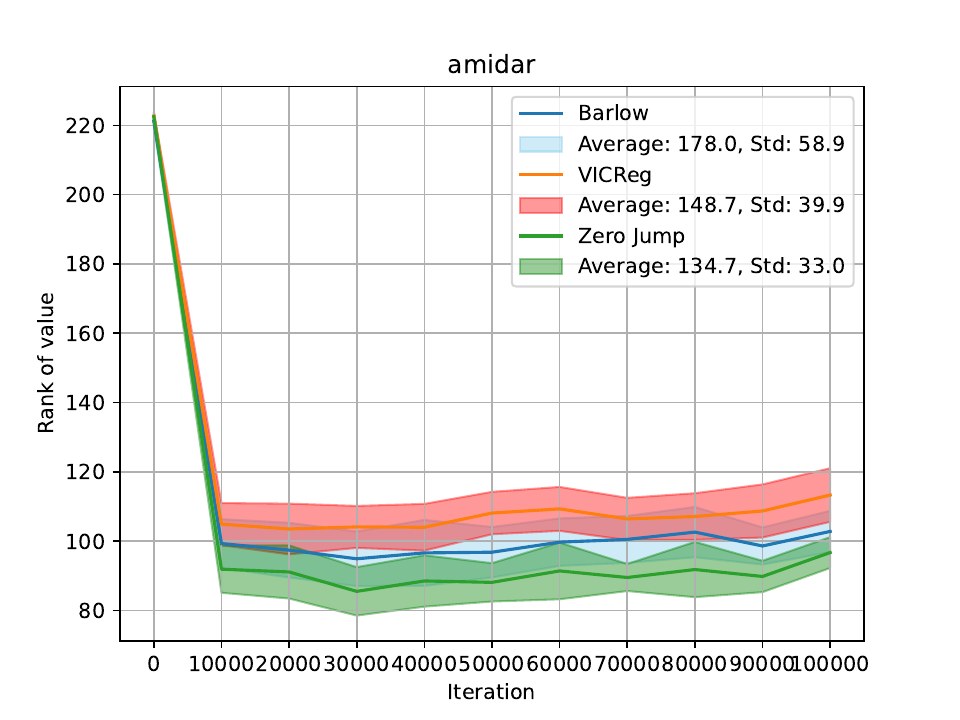}
        \label{fig:value-sub2}
    \end{subfigure}
    \begin{subfigure}[b]{0.2\textwidth}
        \centering
        \includegraphics[width=\textwidth]{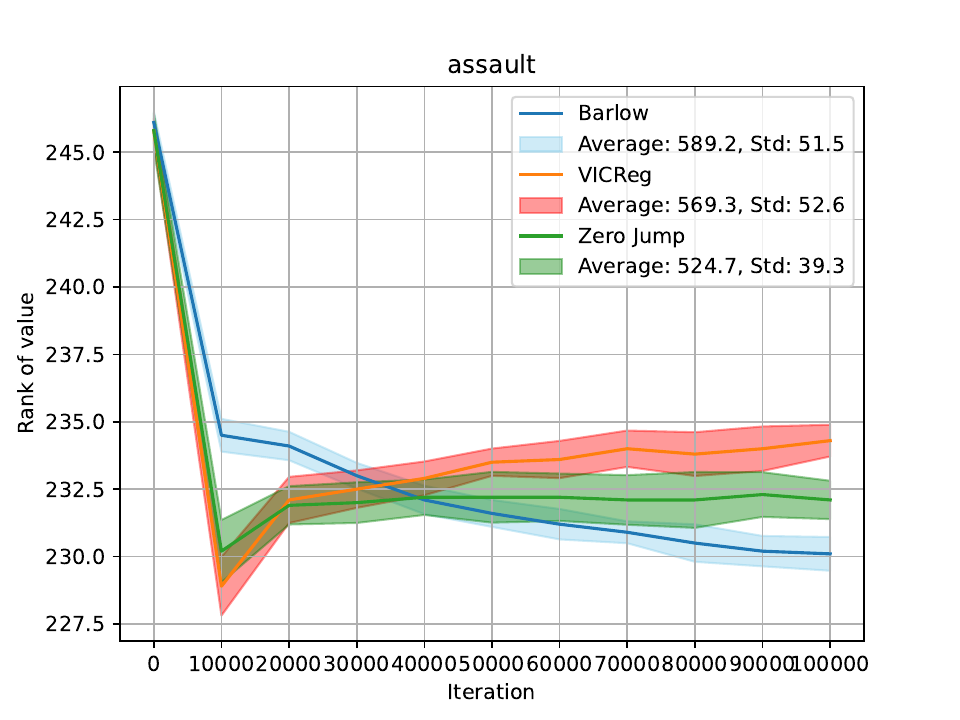}
        \label{fig:value-sub3}
    \end{subfigure}
    \begin{subfigure}[b]{0.2\textwidth}
        \centering
        \includegraphics[width=\textwidth]{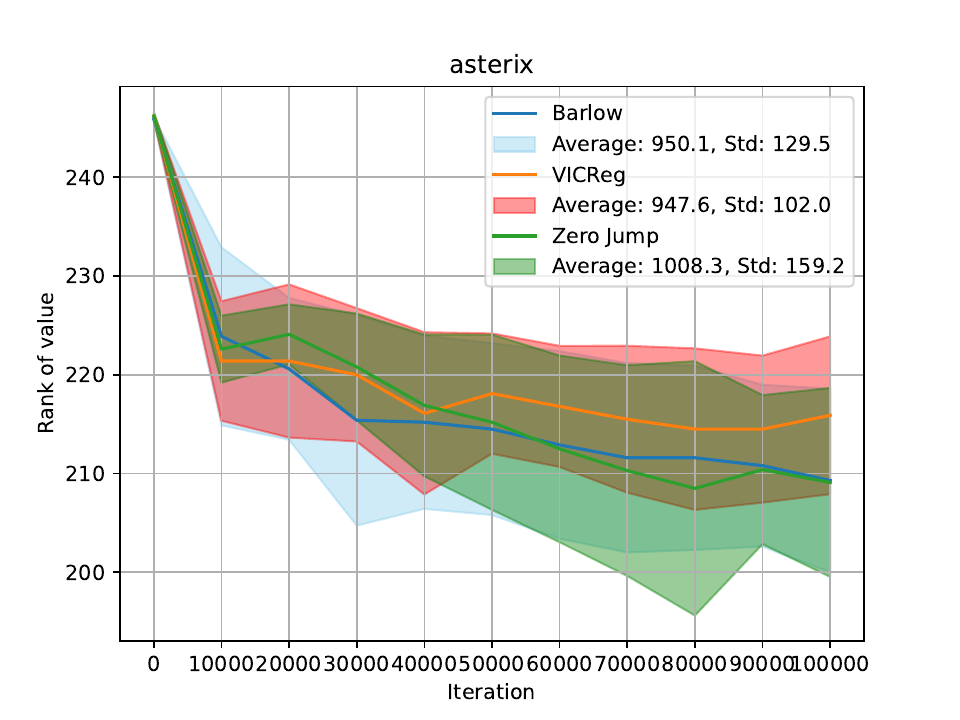}
        \label{fig:value-sub4}
    \end{subfigure}

    \begin{subfigure}[b]{0.2\textwidth}
        \centering
        \includegraphics[width=\textwidth]{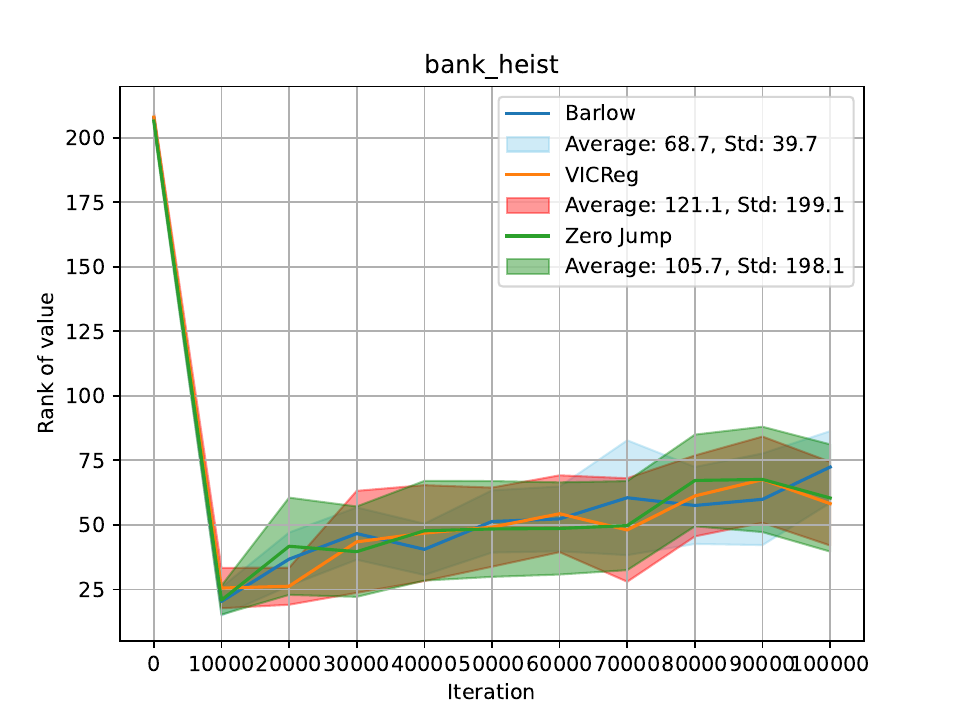}
        \label{fig:value-sub5}
    \end{subfigure}
    \begin{subfigure}[b]{0.2\textwidth}
        \centering
        \includegraphics[width=\textwidth]{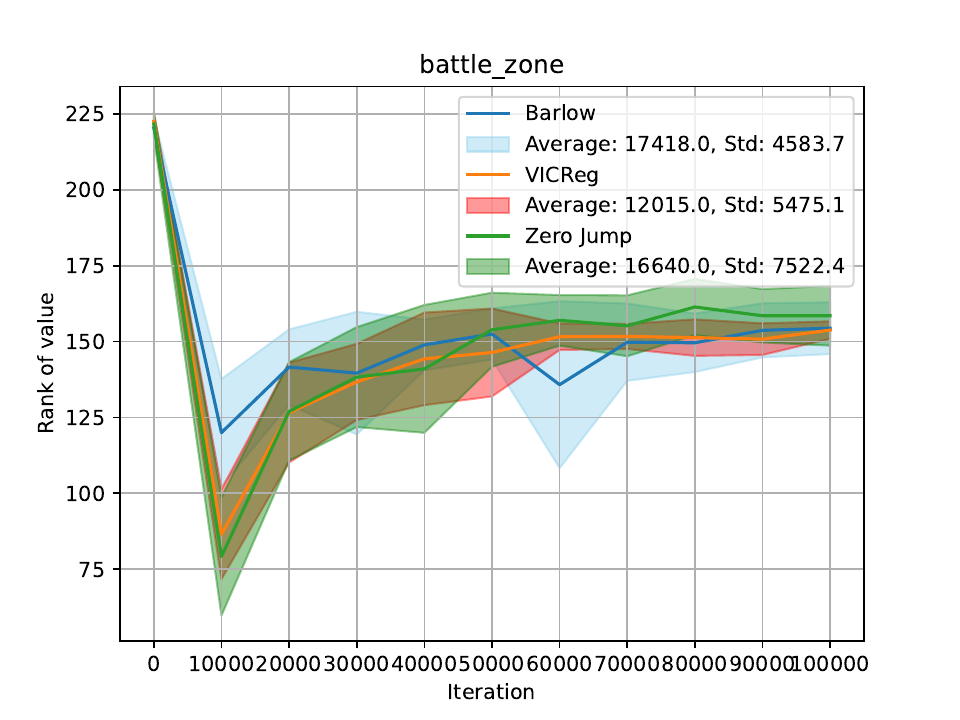}
        \label{fig:value-sub6}
    \end{subfigure}
    \begin{subfigure}[b]{0.2\textwidth}
        \centering
        \includegraphics[width=\textwidth]{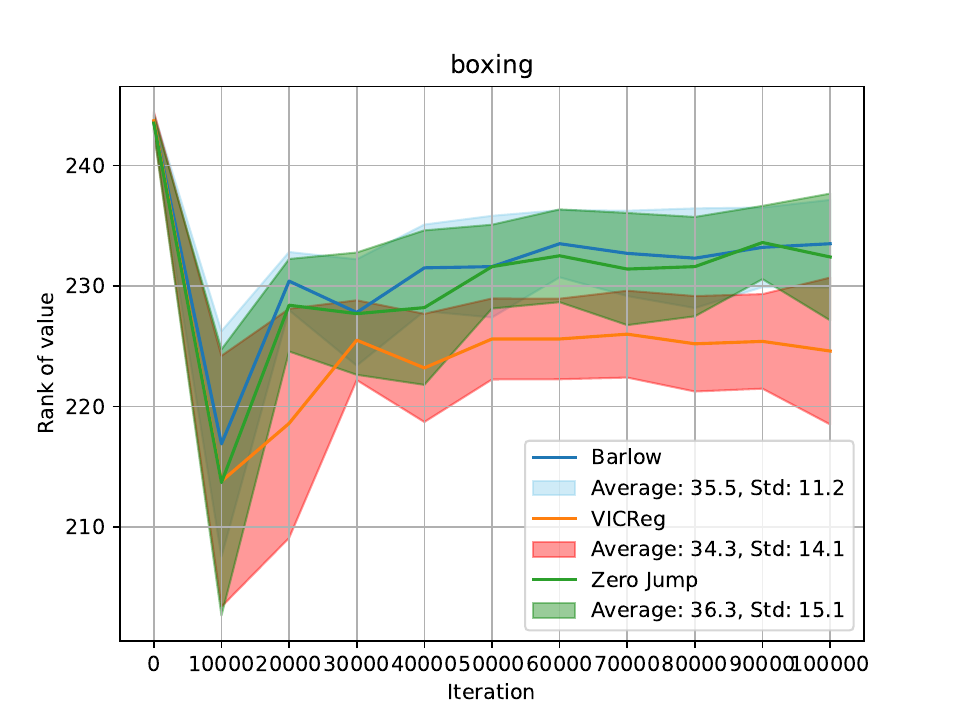}
        \label{fig:value-sub7}
    \end{subfigure}
    \begin{subfigure}[b]{0.2\textwidth}
        \centering
        \includegraphics[width=\textwidth]{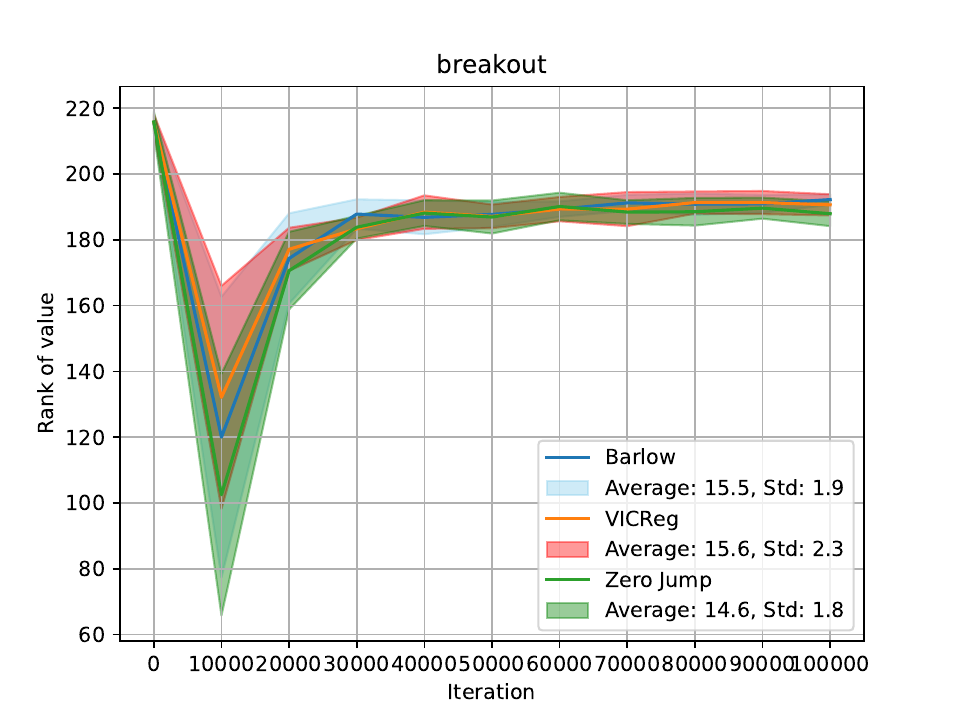}
        \label{fig:value-sub8}
    \end{subfigure}

    \begin{subfigure}[b]{0.2\textwidth}
        \centering
        \includegraphics[width=\textwidth]{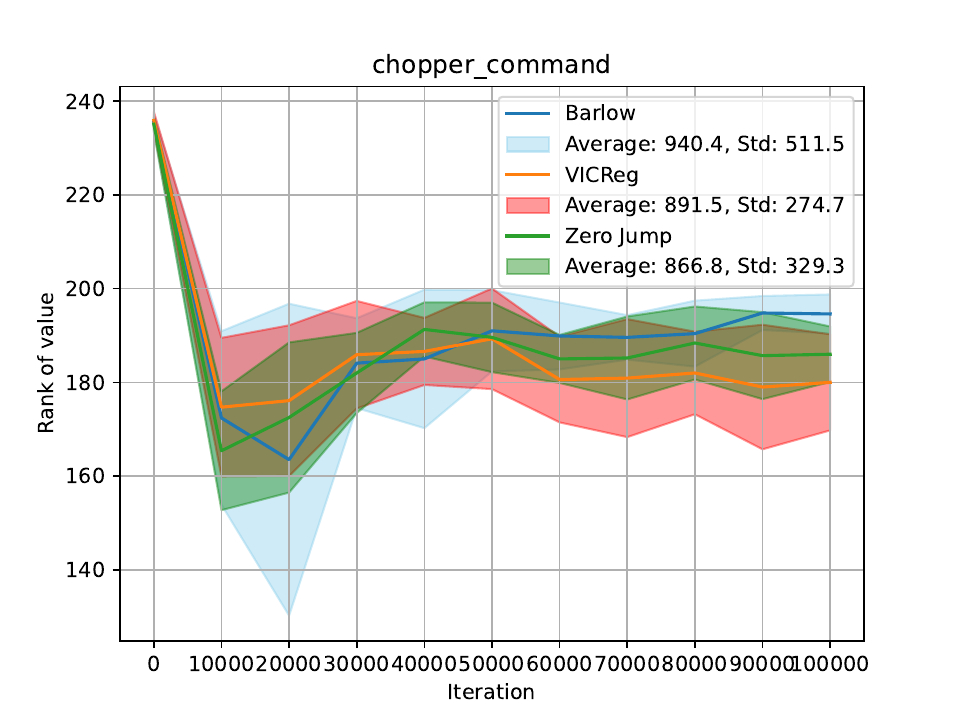}
        \label{fig:value-sub9}
    \end{subfigure}
    \begin{subfigure}[b]{0.2\textwidth}
        \centering
        \includegraphics[width=\textwidth]{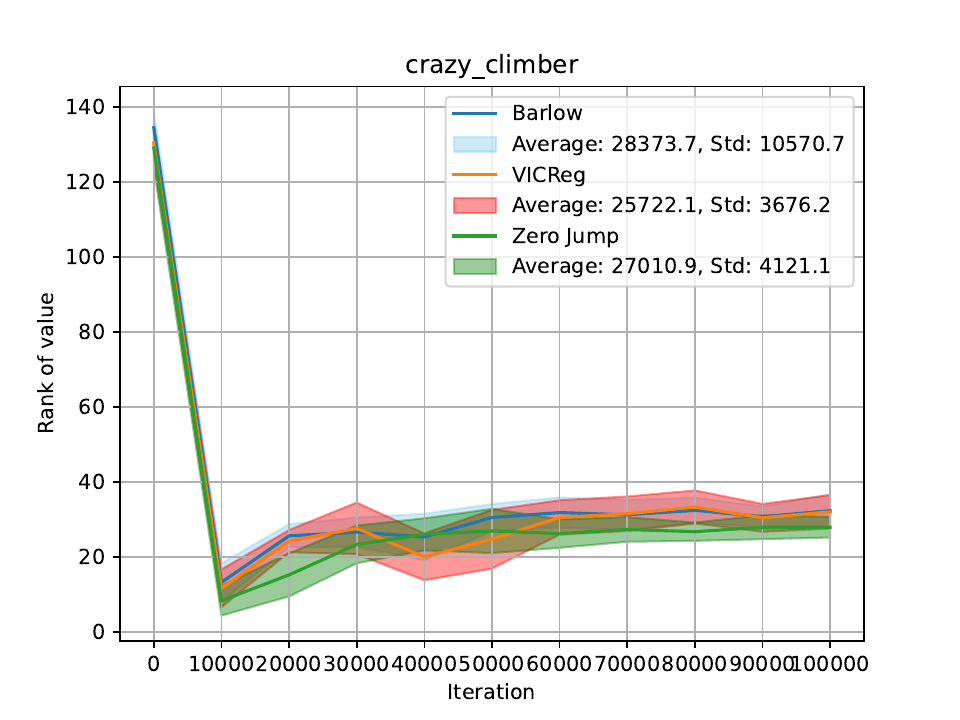}
        \label{fig:value-sub10}
    \end{subfigure}
    \begin{subfigure}[b]{0.2\textwidth}
        \centering
        \includegraphics[width=\textwidth]{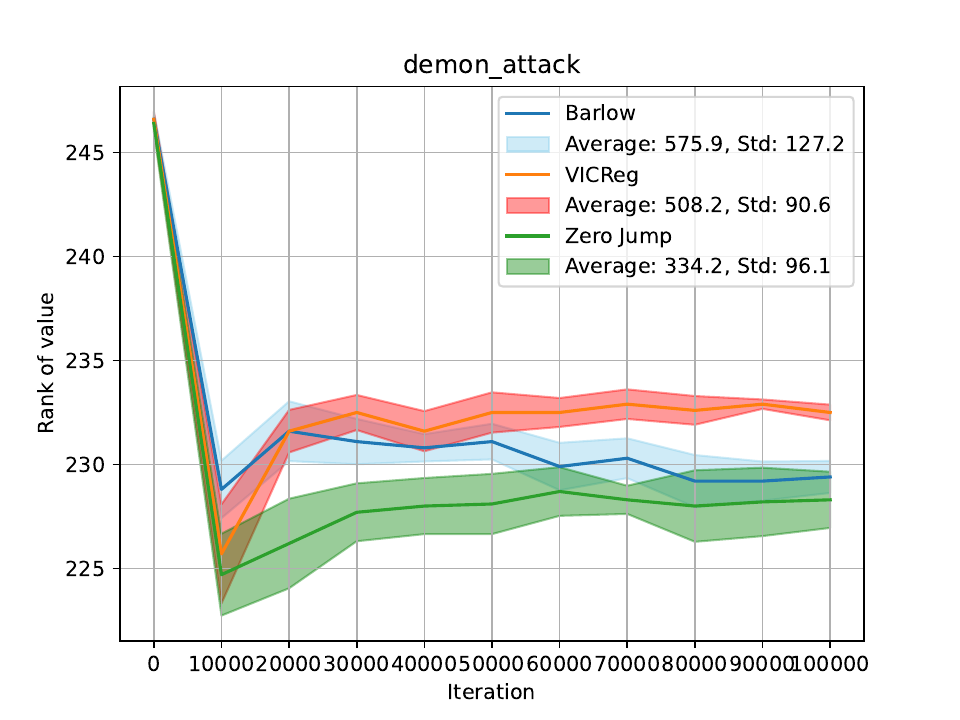}
        \label{fig:value-sub11}
    \end{subfigure}
    \begin{subfigure}[b]{0.2\textwidth}
        \centering
        \includegraphics[width=\textwidth]{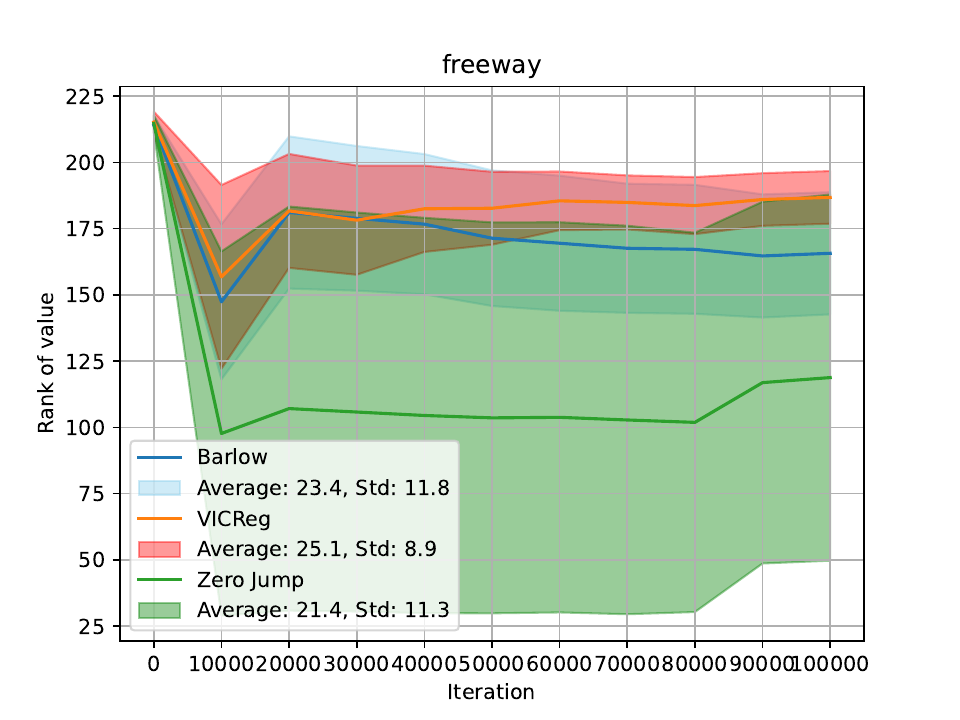}
        \label{fig:value-sub12}
    \end{subfigure}

    \begin{subfigure}[b]{0.2\textwidth}
        \centering
        \includegraphics[width=\textwidth]{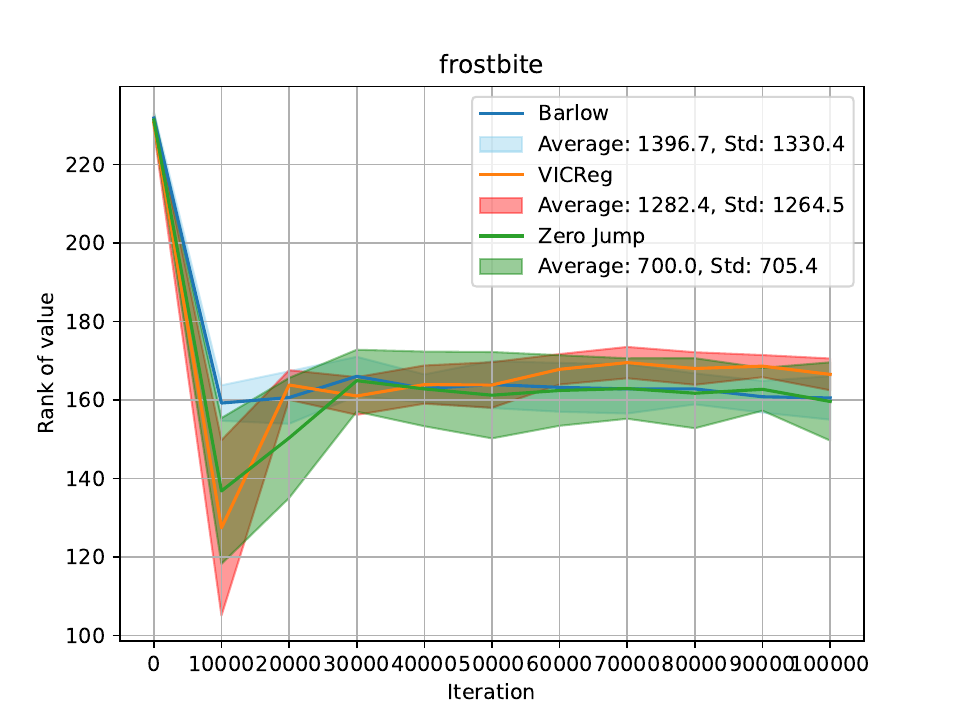}
        \label{fig:value-sub13}
    \end{subfigure}
    \begin{subfigure}[b]{0.2\textwidth}
        \centering
        \includegraphics[width=\textwidth]{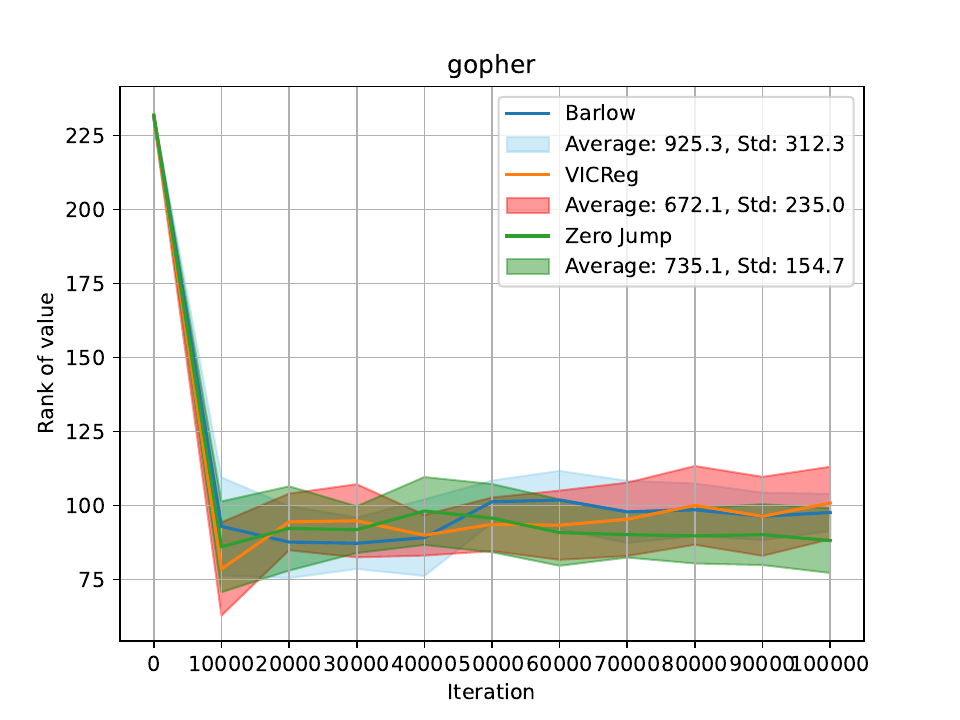}
        \label{fig:value-sub14}
    \end{subfigure}
    \begin{subfigure}[b]{0.2\textwidth}
        \centering
        \includegraphics[width=\textwidth]{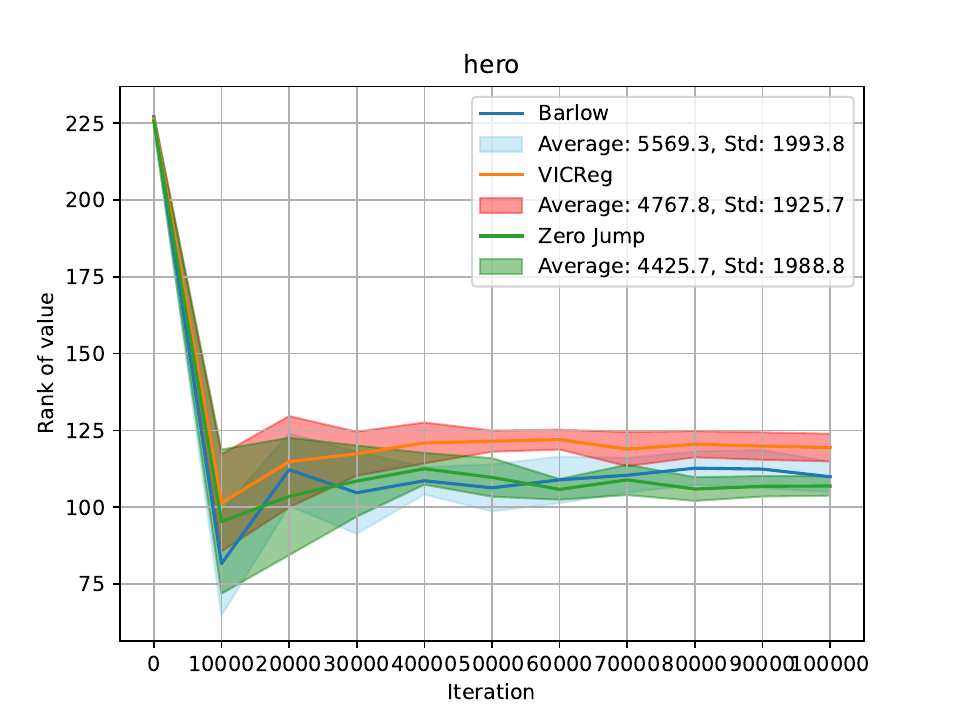}
        \label{fig:value-sub15}
    \end{subfigure}
    \begin{subfigure}[b]{0.2\textwidth}
        \centering
        \includegraphics[width=\textwidth]{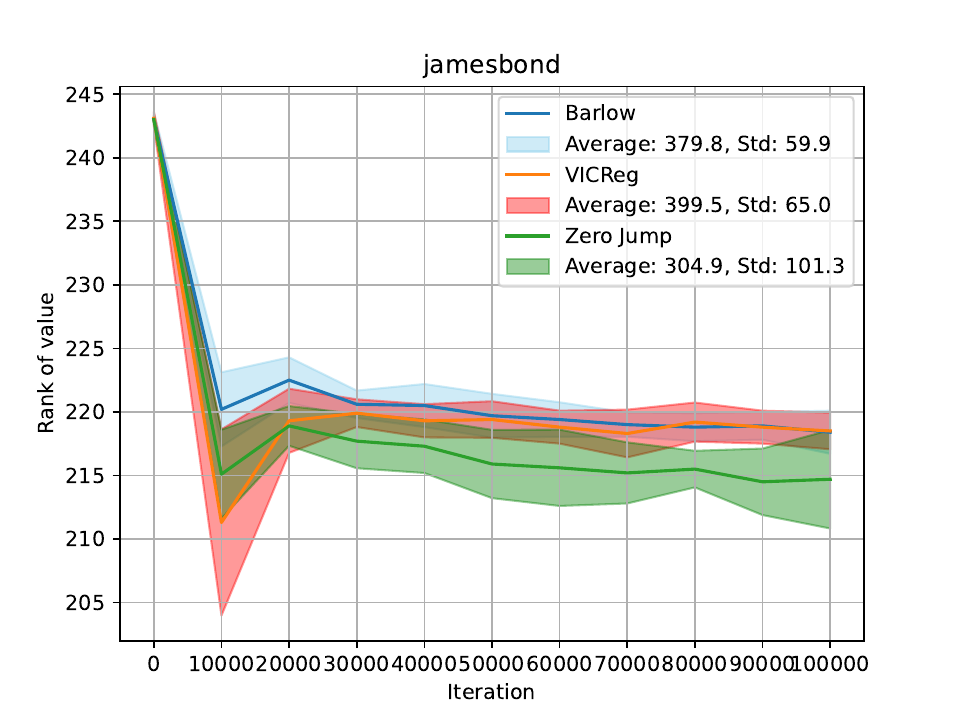}
        \label{fig:value-sub16}
    \end{subfigure}

    \begin{subfigure}[b]{0.2\textwidth}
        \centering
        \includegraphics[width=\textwidth]{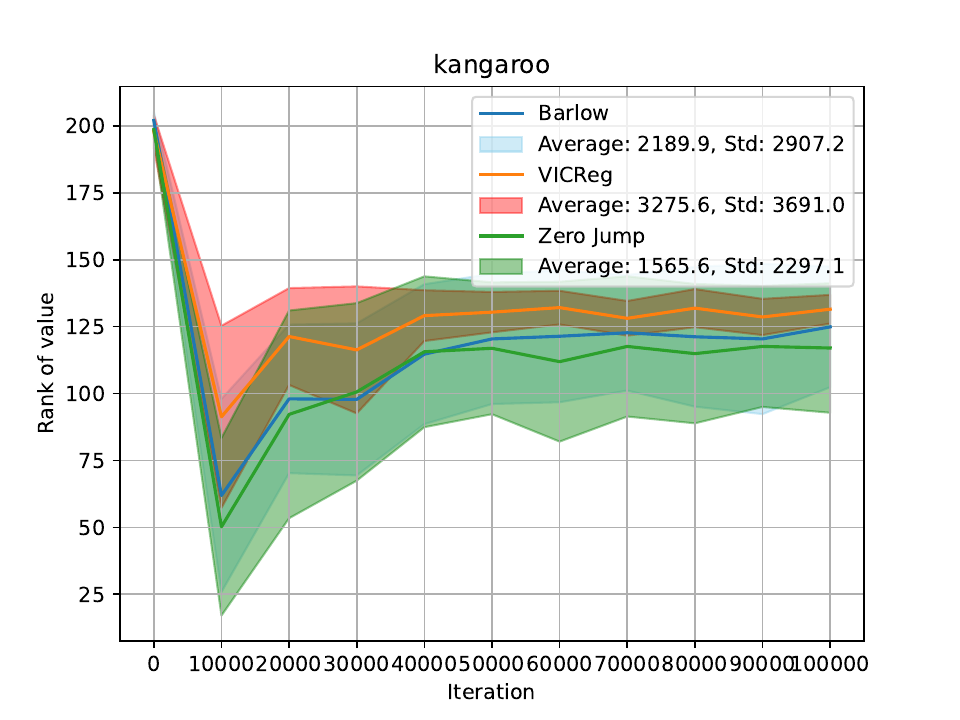}
        \label{fig:value-sub17}
    \end{subfigure}
    \begin{subfigure}[b]{0.2\textwidth}
        \centering
        \includegraphics[width=\textwidth]{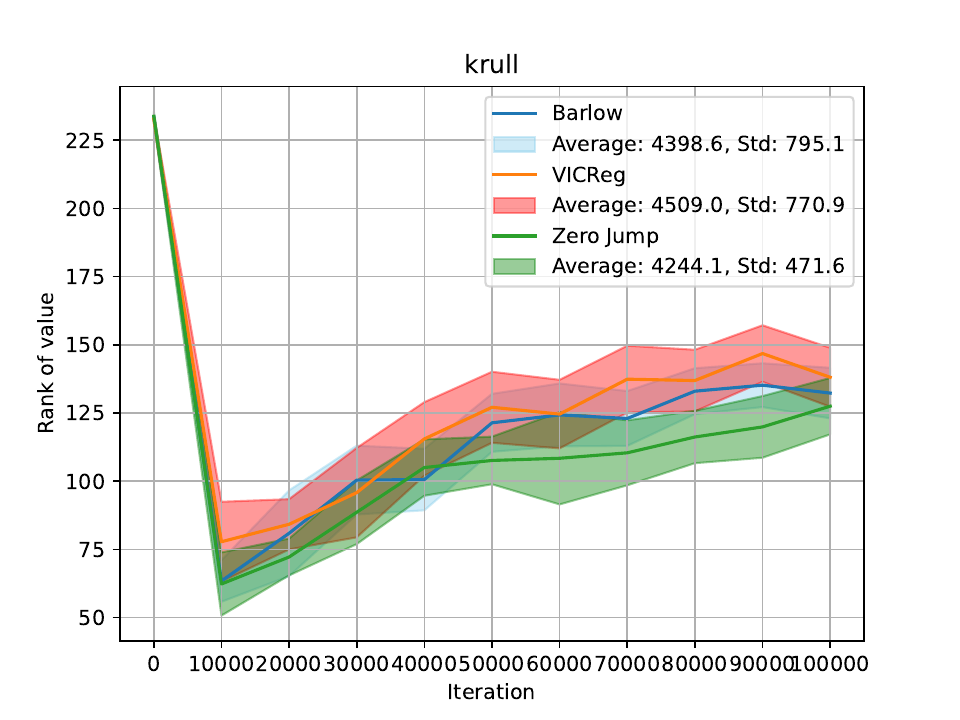}
        \label{fig:value-sub18}
    \end{subfigure}
    \begin{subfigure}[b]{0.2\textwidth}
        \centering
        \includegraphics[width=\textwidth]{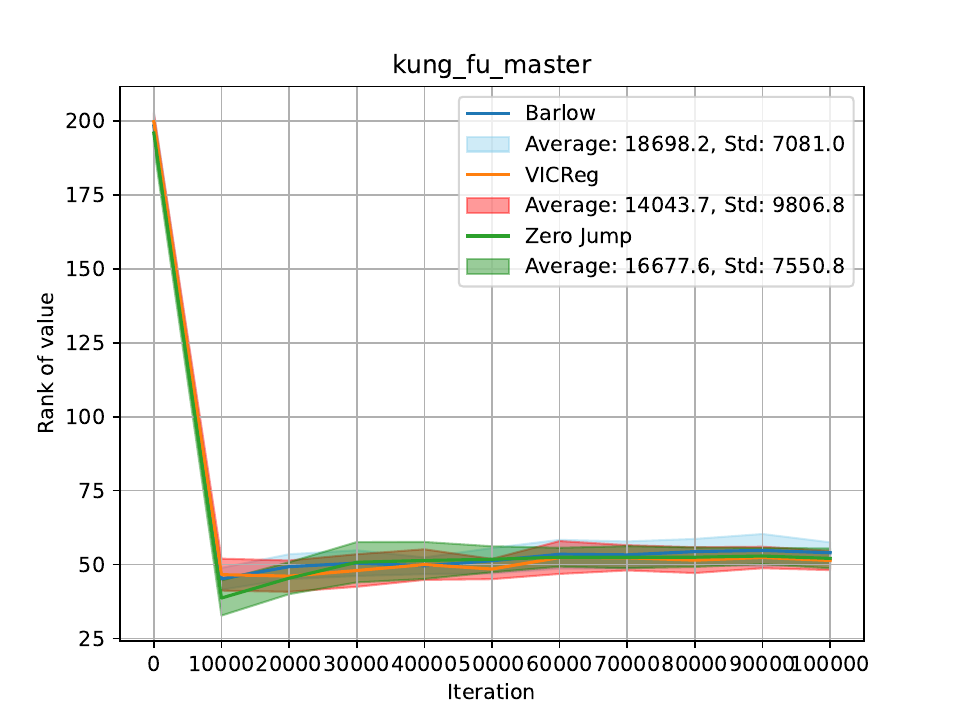}
        \label{fig:value-sub19}
    \end{subfigure}
    \begin{subfigure}[b]{0.2\textwidth}
        \centering
        \includegraphics[width=\textwidth]{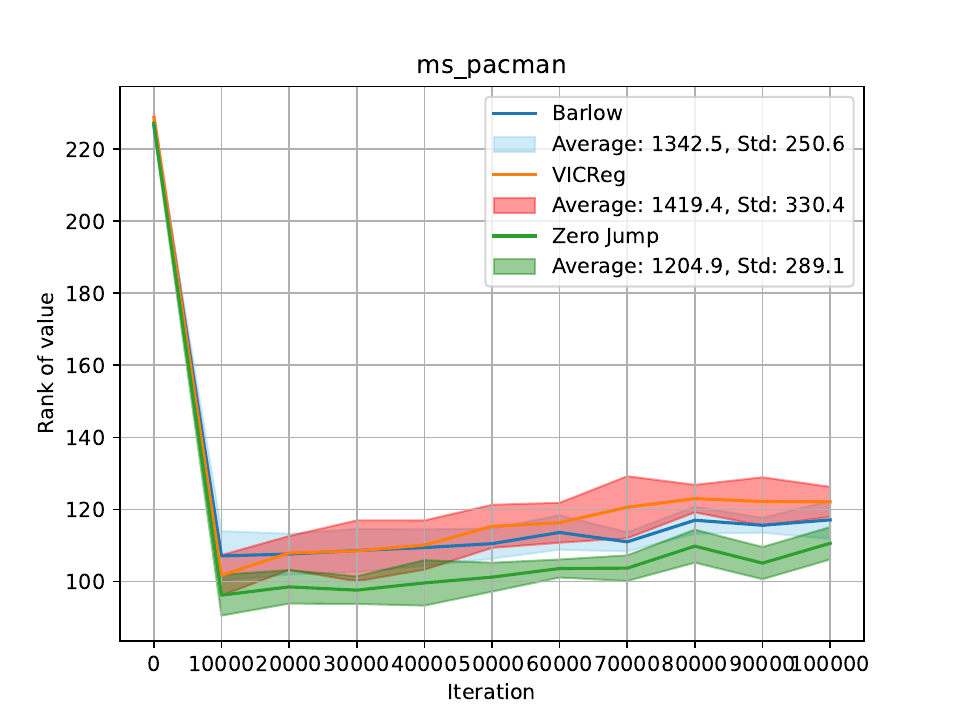}
        \label{fig:value-sub20}
    \end{subfigure}

    \begin{subfigure}[b]{0.2\textwidth}
        \centering
        \includegraphics[width=\textwidth]{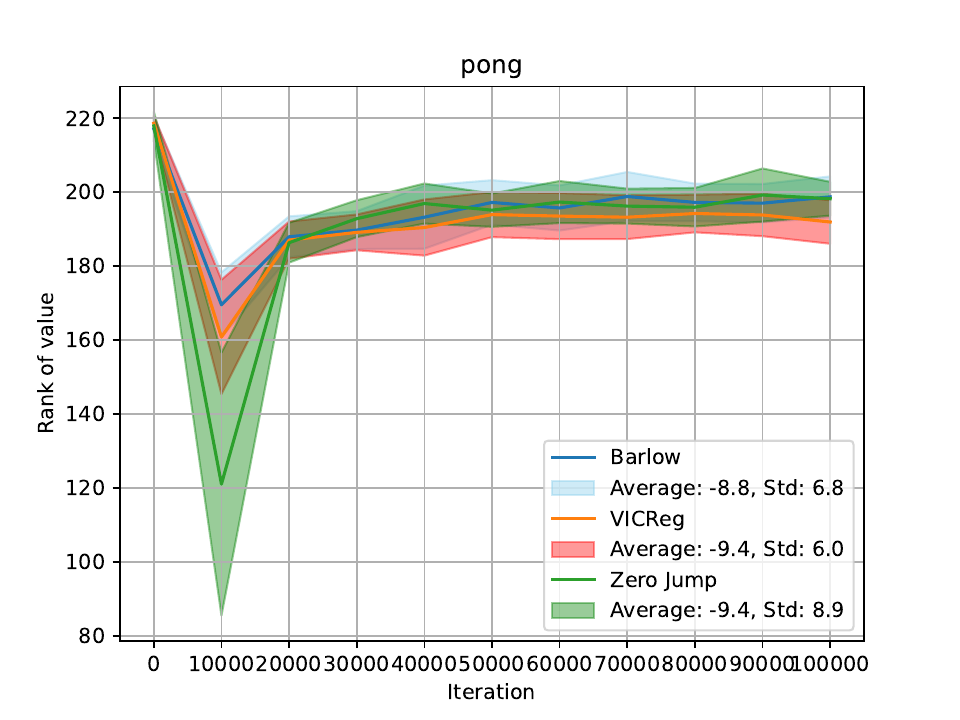}
        \label{fig:value-sub21}
    \end{subfigure}
    \begin{subfigure}[b]{0.2\textwidth}
        \centering
        \includegraphics[width=\textwidth]{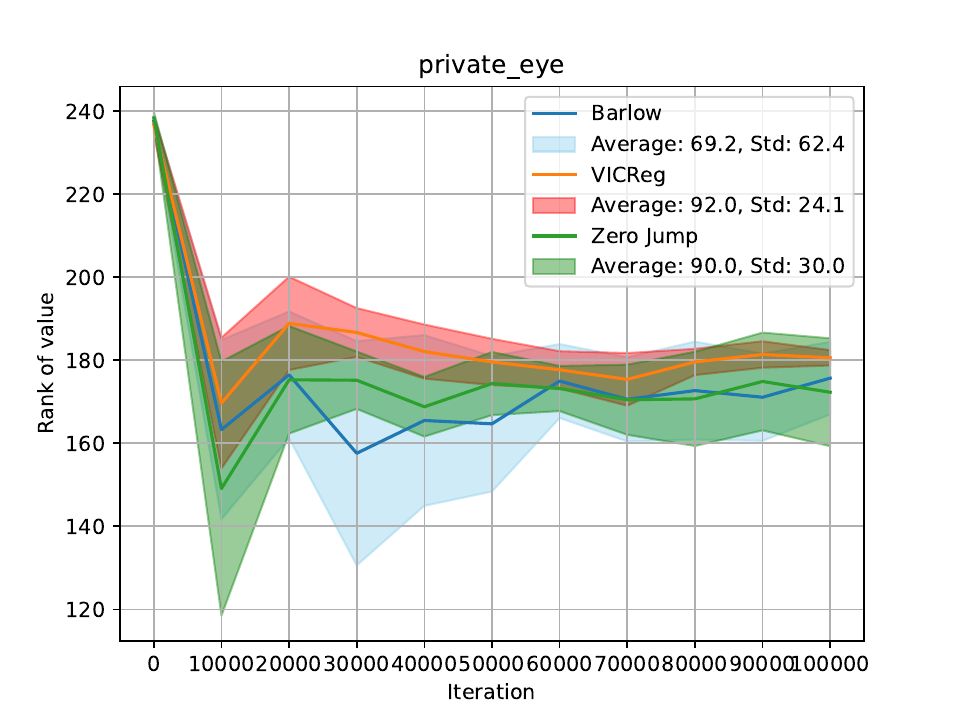}
        \label{fig:value-sub22}
    \end{subfigure}
    \begin{subfigure}[b]{0.2\textwidth}
        \centering
        \includegraphics[width=\textwidth]{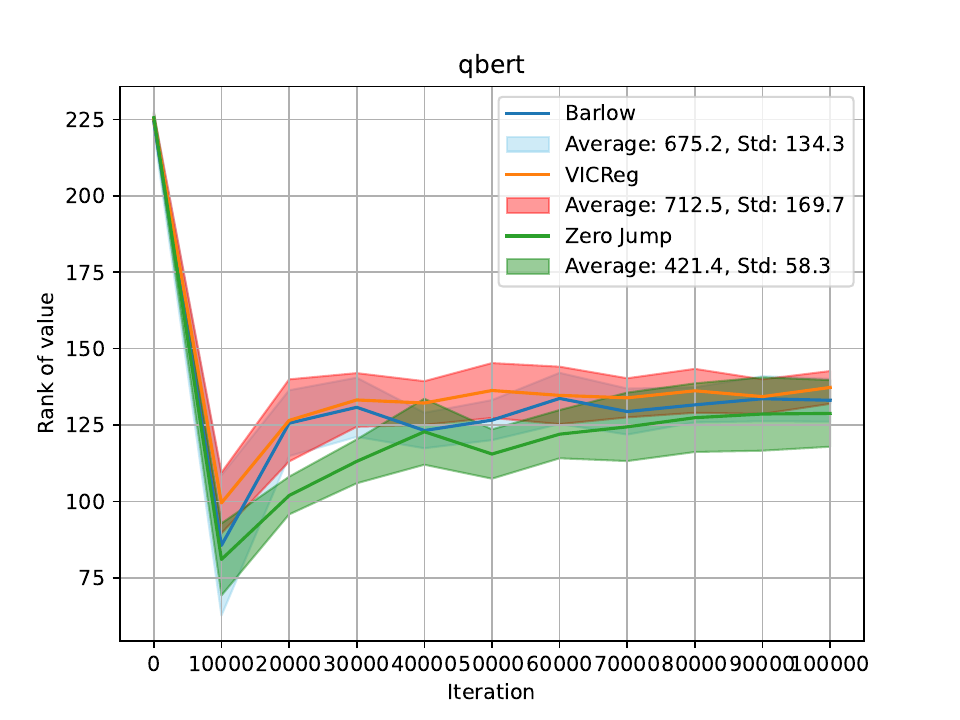}
        \label{fig:value-sub23}
    \end{subfigure}
    \begin{subfigure}[b]{0.2\textwidth}
        \centering
        \includegraphics[width=\textwidth]{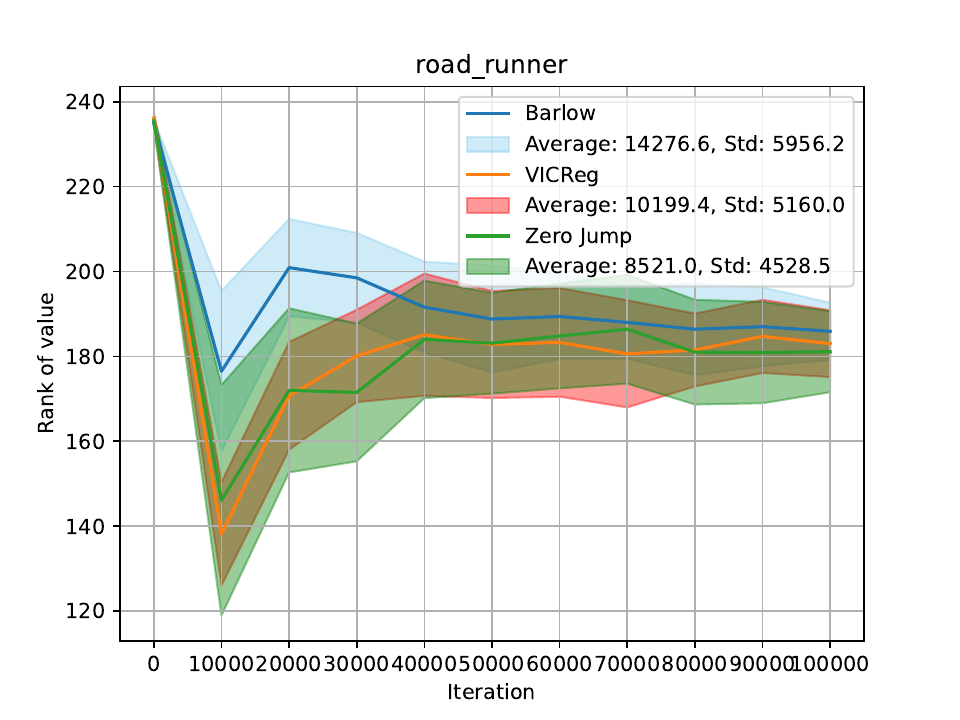}
        \label{fig:value-sub24}
    \end{subfigure}

    \begin{subfigure}[b]{0.2\textwidth}
        \centering
        \includegraphics[width=\textwidth]{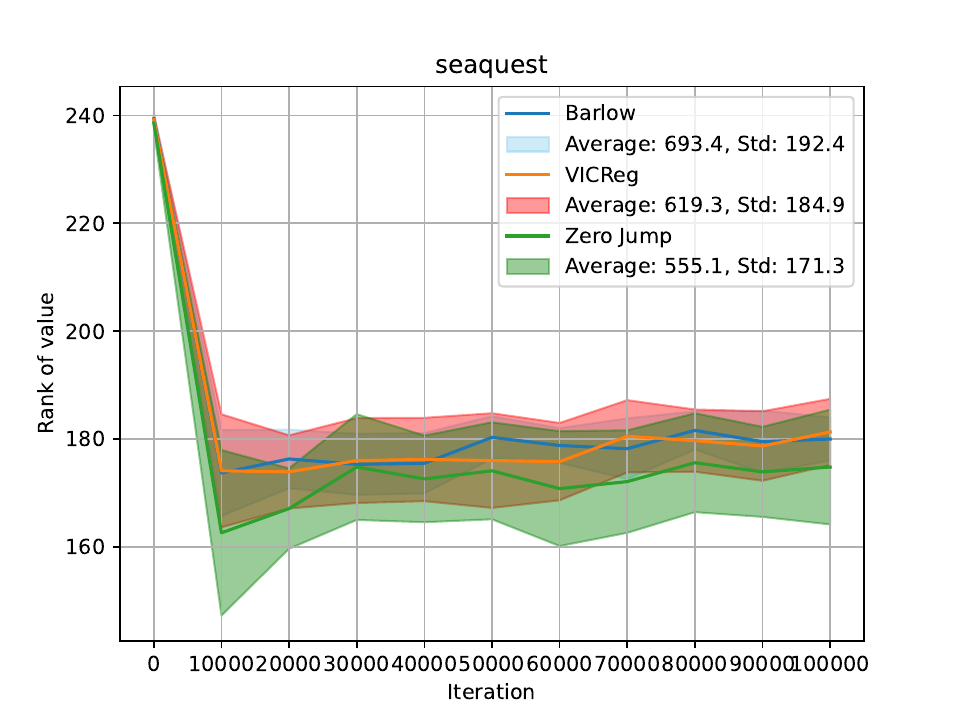}
        \label{fig:value-sub25}
    \end{subfigure}
    \begin{subfigure}[b]{0.2\textwidth}
        \centering
        \includegraphics[width=\textwidth]{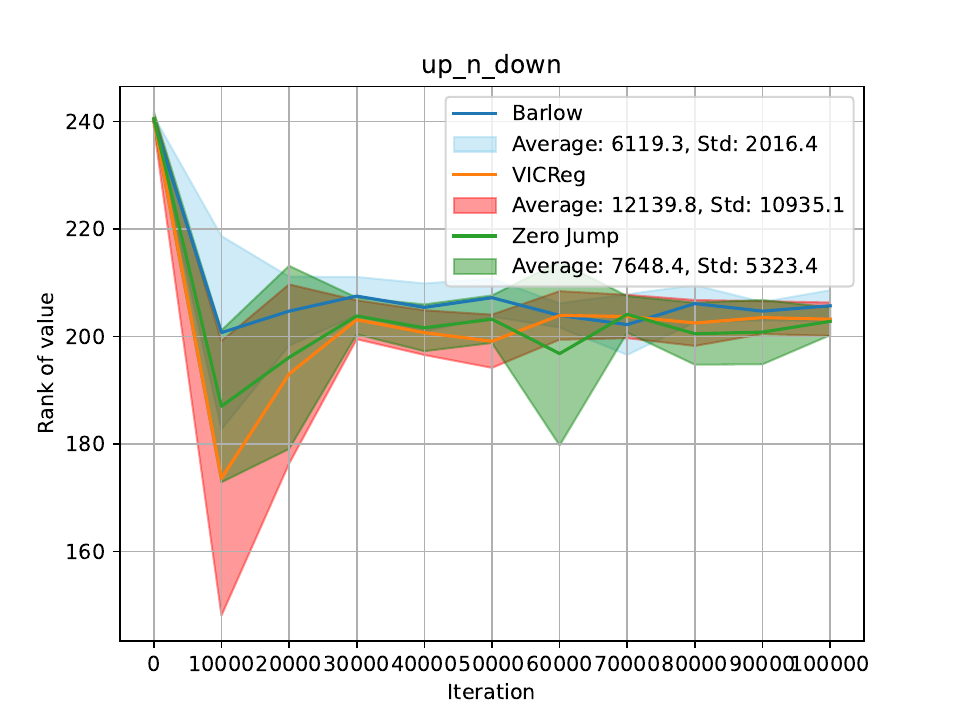}
        \label{fig:value-sub26}
    \end{subfigure}

    \begin{subfigure}[b]{0.2\textwidth}
    \end{subfigure}
    \begin{subfigure}[b]{0.2\textwidth}
    \end{subfigure}

    \label{fig:val-rank}

\end{figure}

\begin{figure*}[htbp]
    \centering
    \caption{Rank of the output from the  convolution encoder, measured every 10,000 steps and averaged across 10 different runs for every game.}

    \begin{subfigure}[b]{0.2\textwidth}
        \centering
        \includegraphics[width=\textwidth]{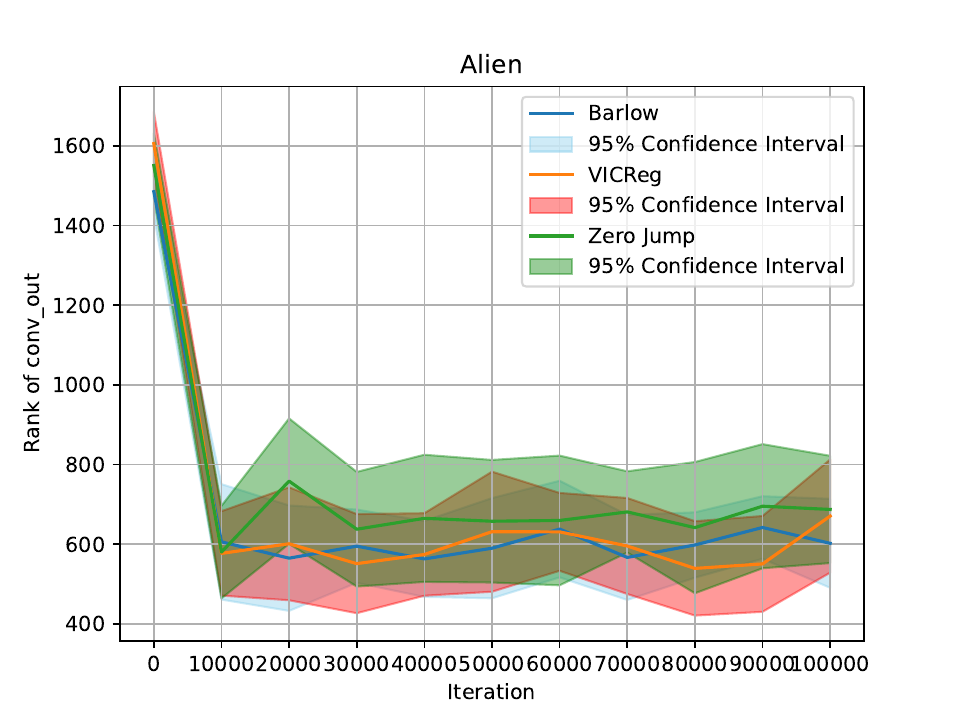}
        \label{fig:conv_out-sub1}
    \end{subfigure}
    \begin{subfigure}[b]{0.2\textwidth}
        \centering
        \includegraphics[width=\textwidth]{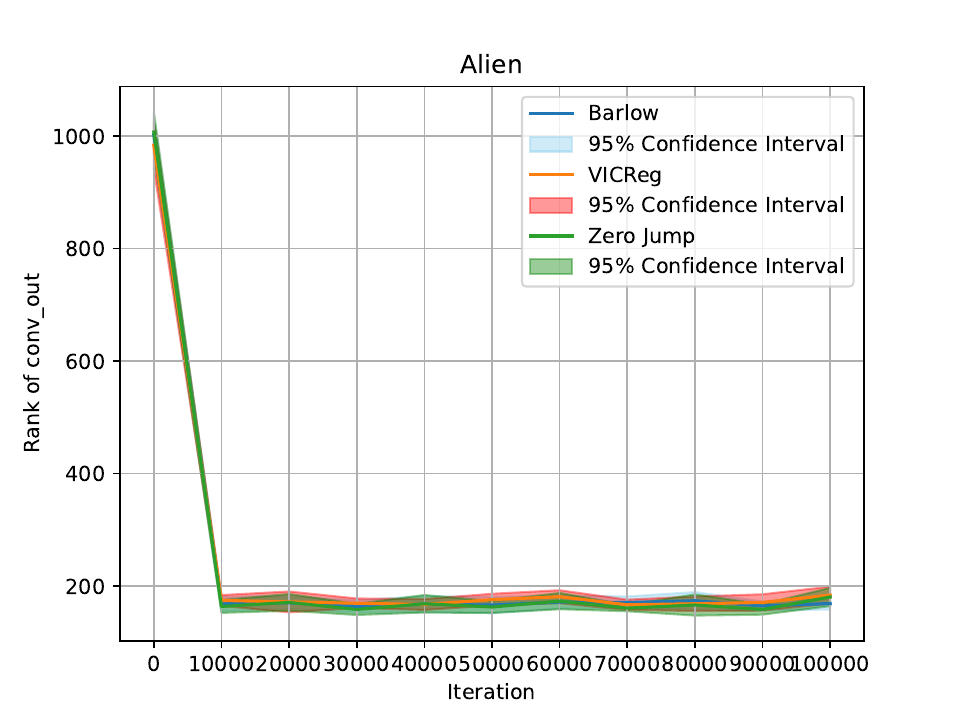}
        \label{fig:conv_out-sub2}
    \end{subfigure}
    \begin{subfigure}[b]{0.2\textwidth}
        \centering
        \includegraphics[width=\textwidth]{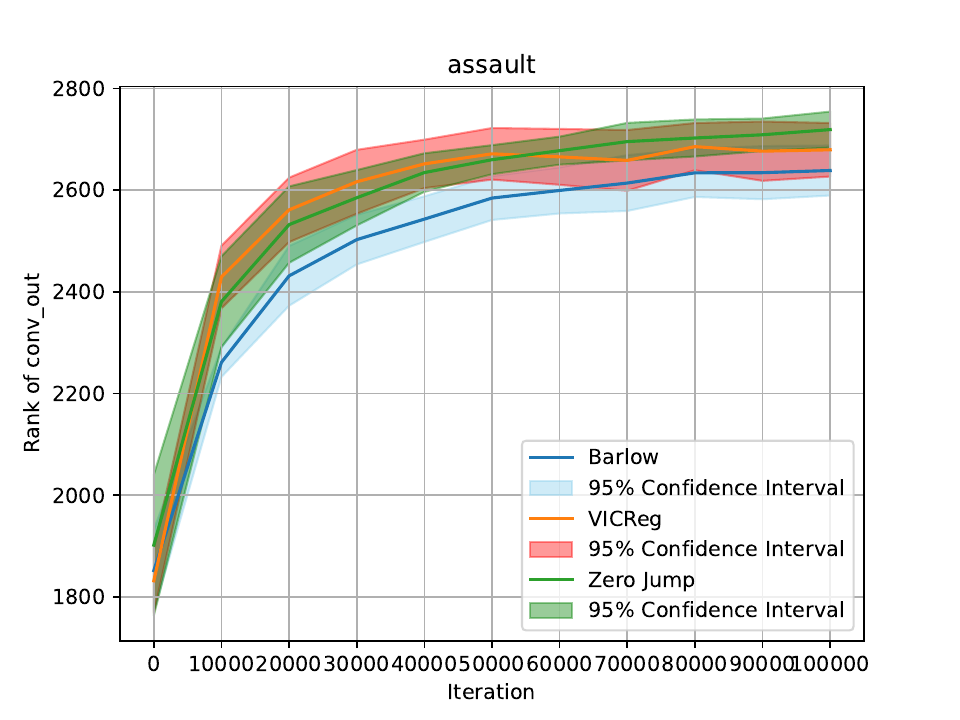}
        \label{fig:conv_out-sub3}
    \end{subfigure}
    \begin{subfigure}[b]{0.2\textwidth}
        \centering
        \includegraphics[width=\textwidth]{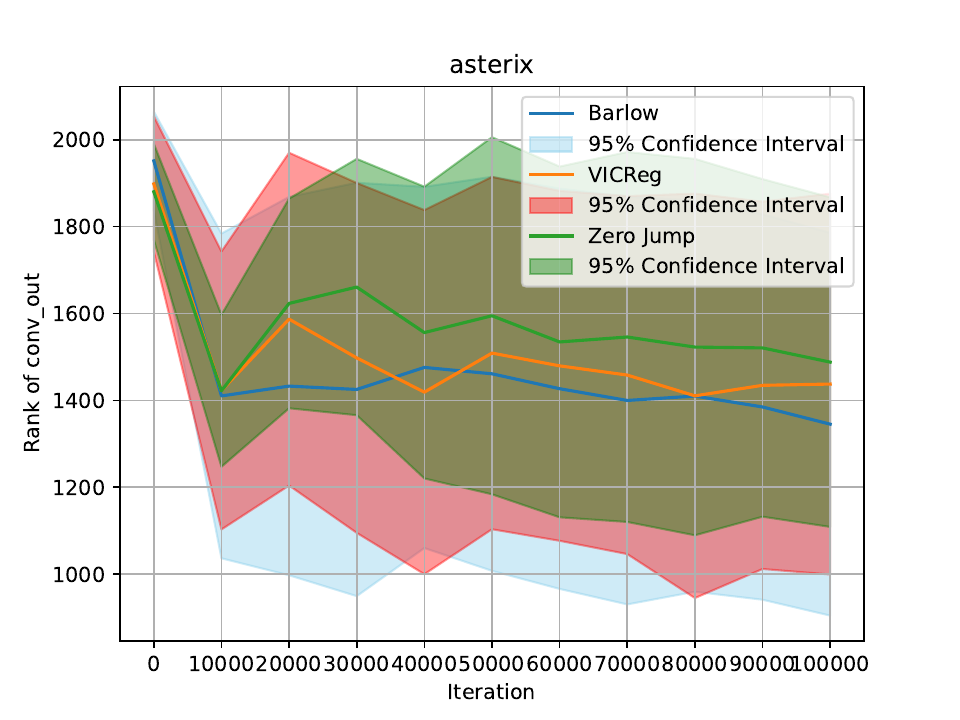}
        \label{fig:conv_out-sub4}
    \end{subfigure}

    \begin{subfigure}[b]{0.2\textwidth}
        \centering
        \includegraphics[width=\textwidth]{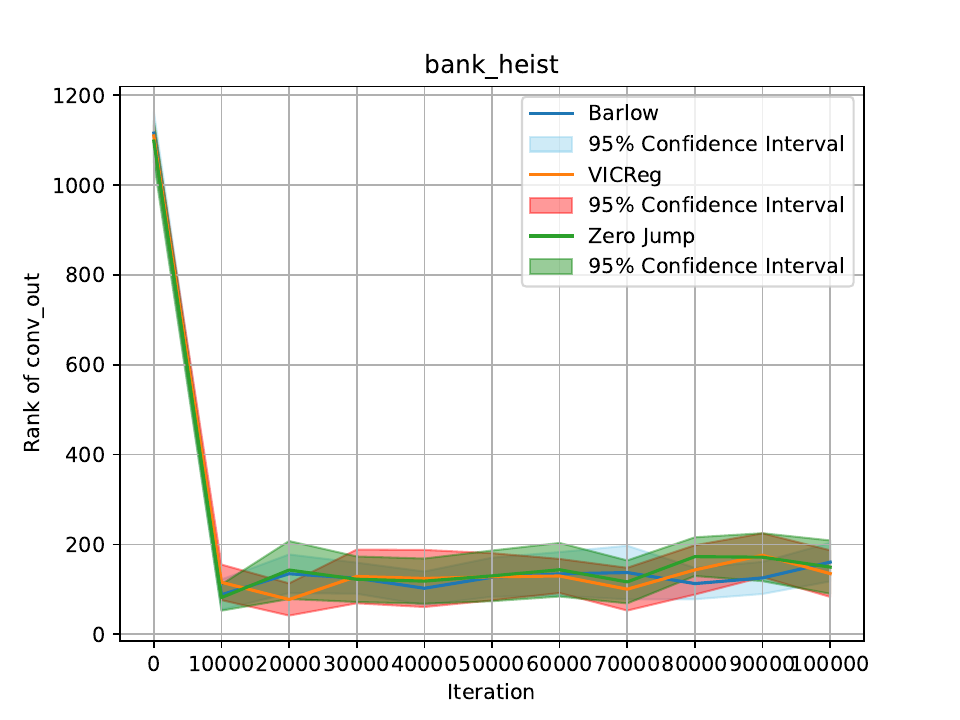}
        \label{fig:conv_out-sub5}
    \end{subfigure}
    \begin{subfigure}[b]{0.2\textwidth}
        \centering
        \includegraphics[width=\textwidth]{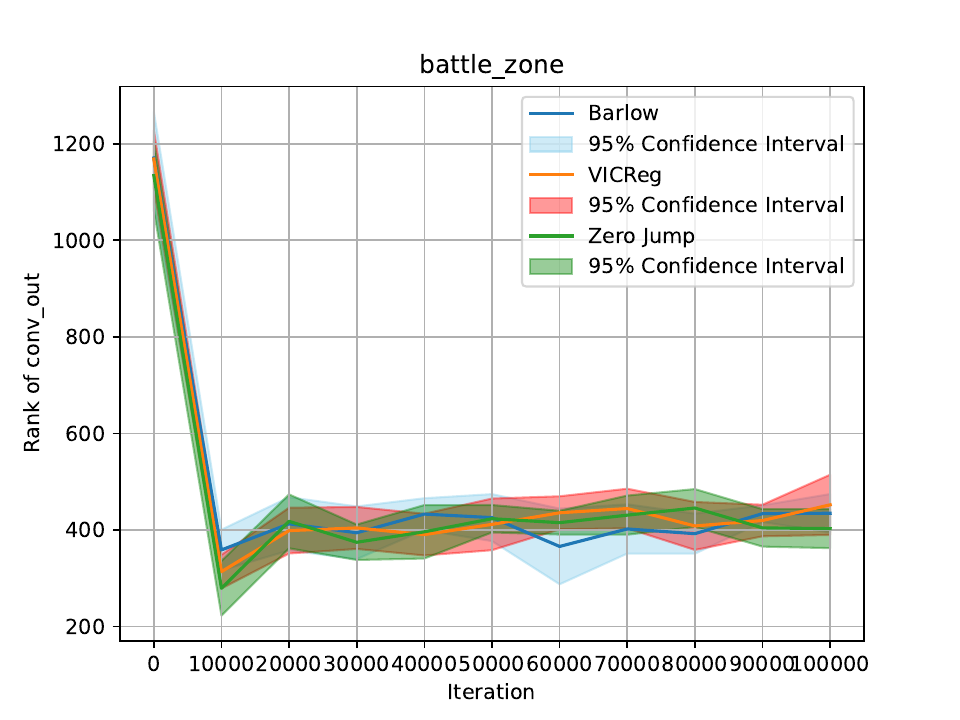}
        \label{fig:conv_out-sub6}
    \end{subfigure}
    \begin{subfigure}[b]{0.2\textwidth}
        \centering
        \includegraphics[width=\textwidth]{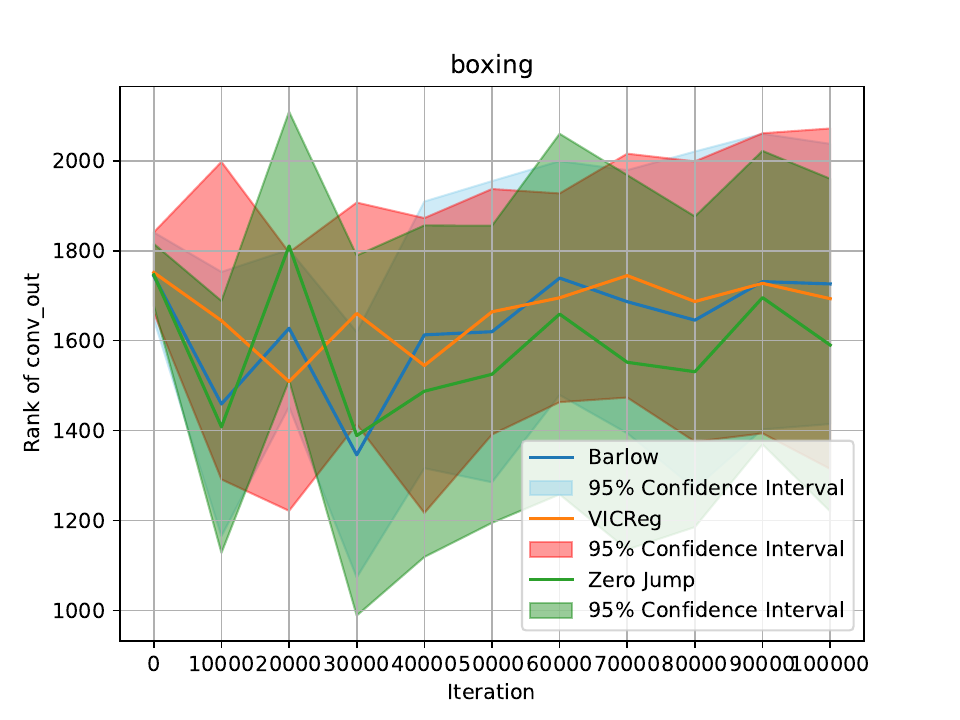}
        \label{fig:conv_out-sub7}
    \end{subfigure}
    \begin{subfigure}[b]{0.2\textwidth}
        \centering
        \includegraphics[width=\textwidth]{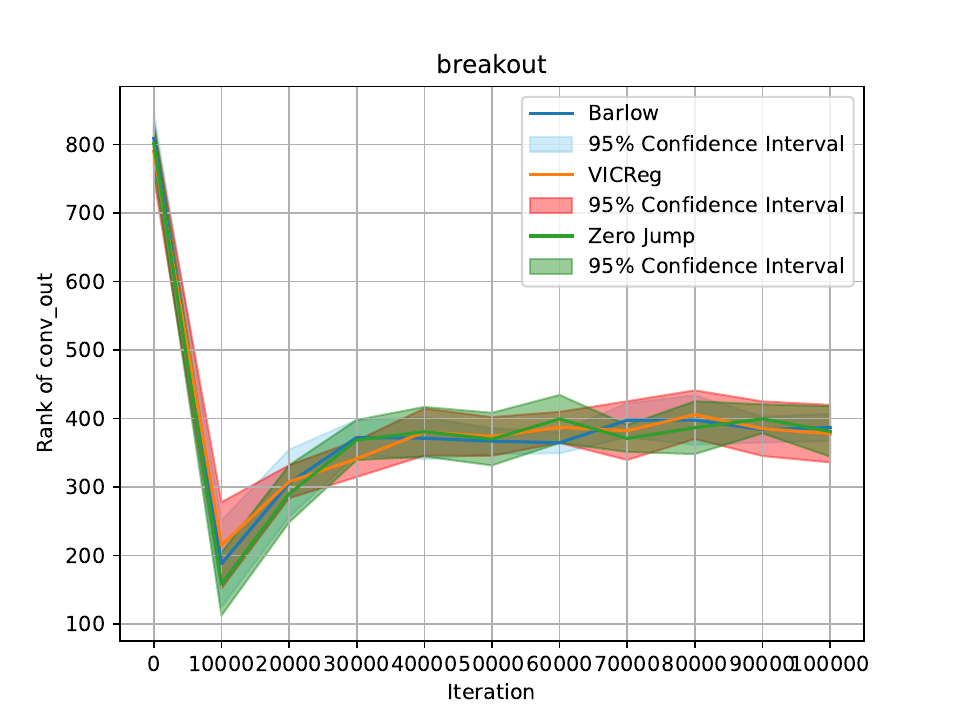}
        \label{fig:conv_out-sub8}
    \end{subfigure}

    \begin{subfigure}[b]{0.2\textwidth}
        \centering
        \includegraphics[width=\textwidth]{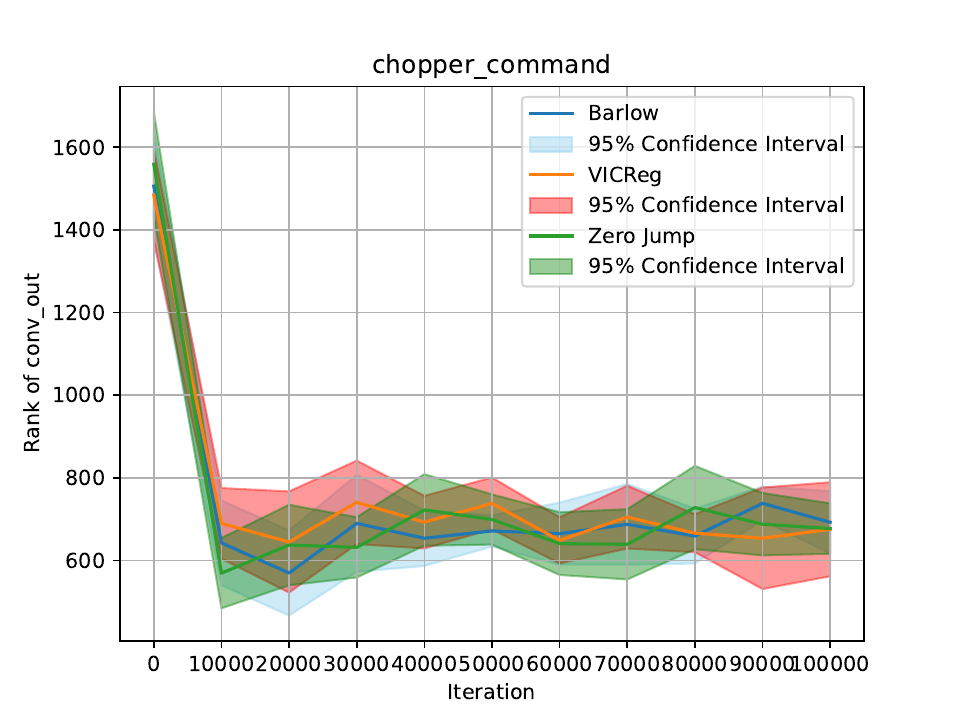}
        \label{fig:conv_out-sub9}
    \end{subfigure}
    \begin{subfigure}[b]{0.2\textwidth}
        \centering
        \includegraphics[width=\textwidth]{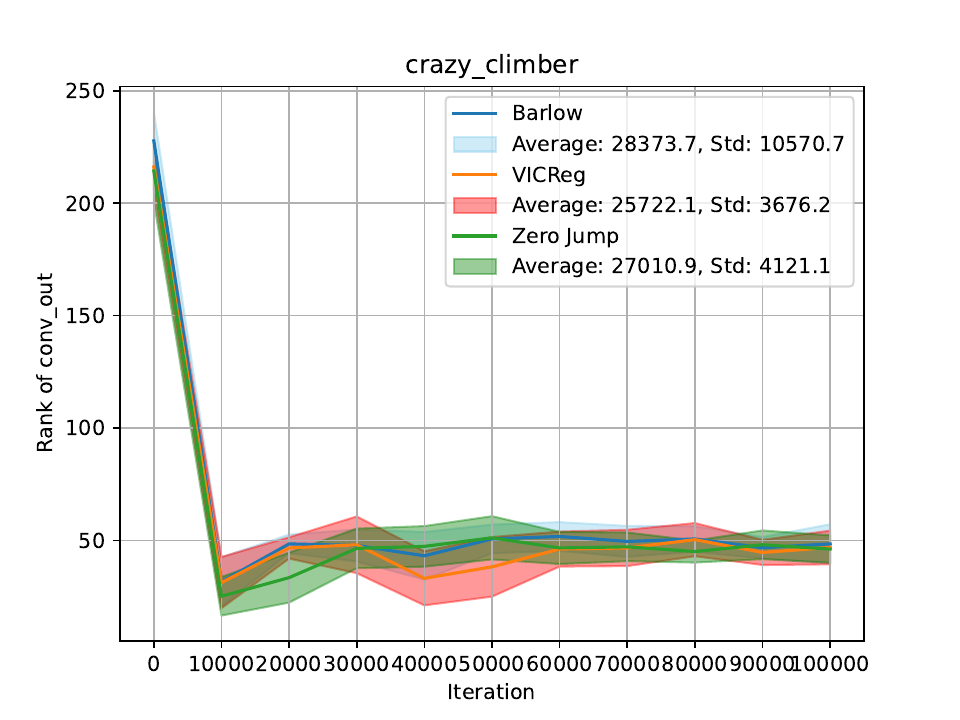}
        \label{fig:conv_out-sub10}
    \end{subfigure}
    \begin{subfigure}[b]{0.2\textwidth}
        \centering
        \includegraphics[width=\textwidth]{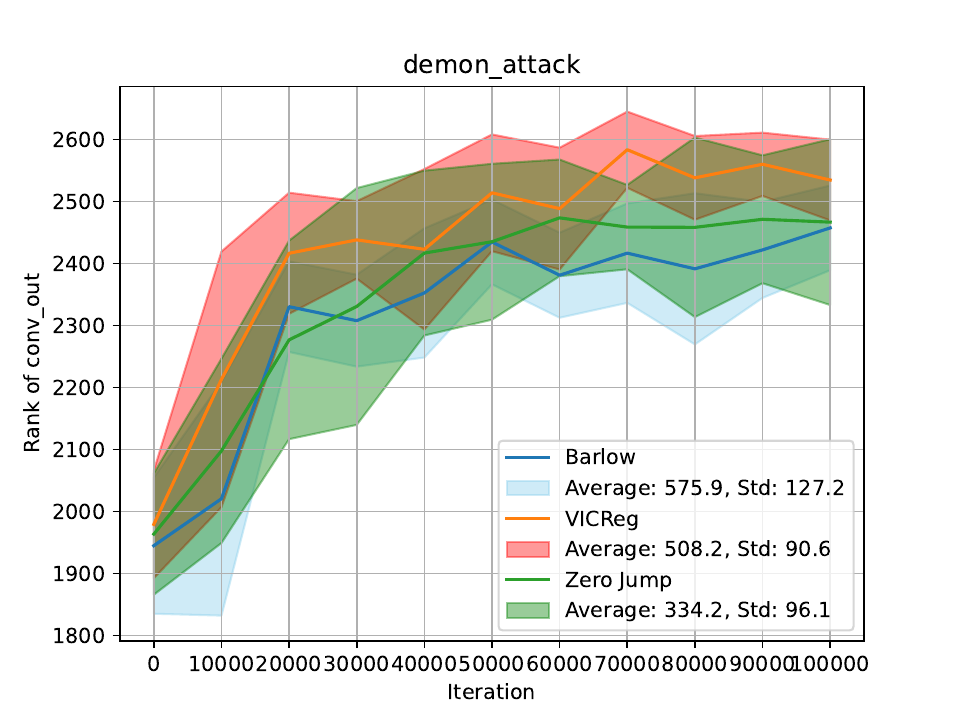}
        \label{fig:conv_out-sub11}
    \end{subfigure}
    \begin{subfigure}[b]{0.2\textwidth}
        \centering
        \includegraphics[width=\textwidth]{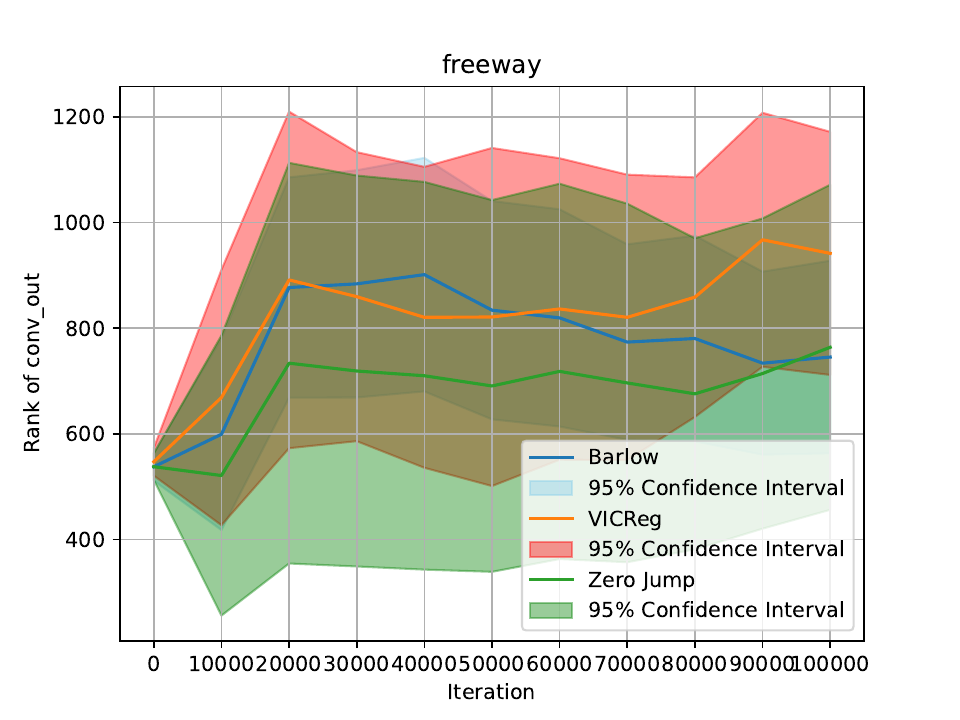}
        \label{fig:conv_out-sub12}
    \end{subfigure}

    \begin{subfigure}[b]{0.2\textwidth}
        \centering
        \includegraphics[width=\textwidth]{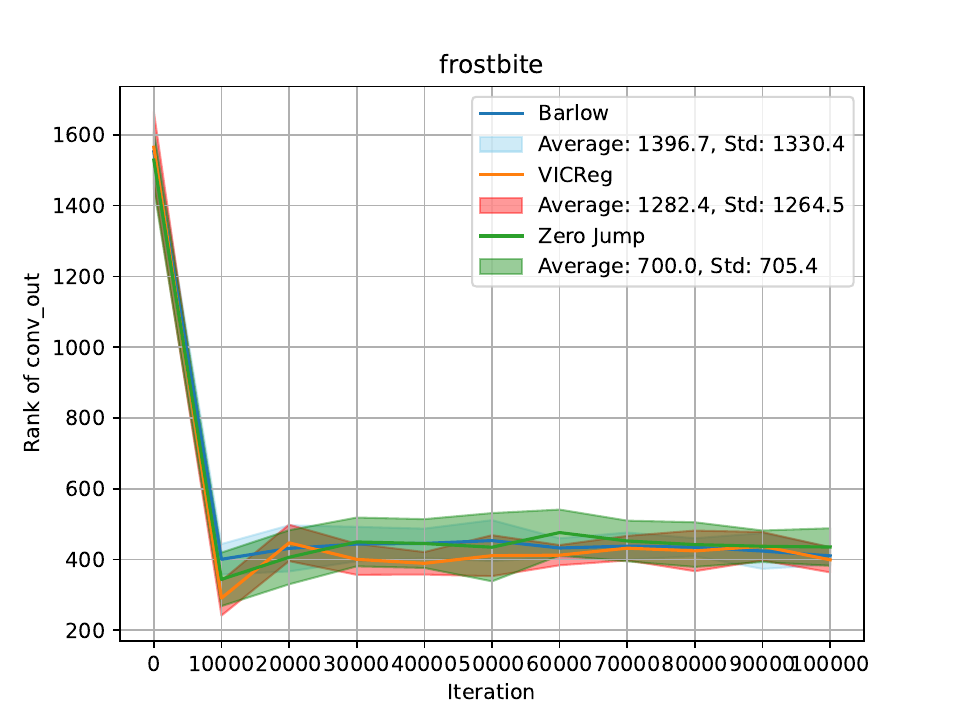}
        \label{fig:conv_out-sub13}
    \end{subfigure}
    \begin{subfigure}[b]{0.2\textwidth}
        \centering
        \includegraphics[width=\textwidth]{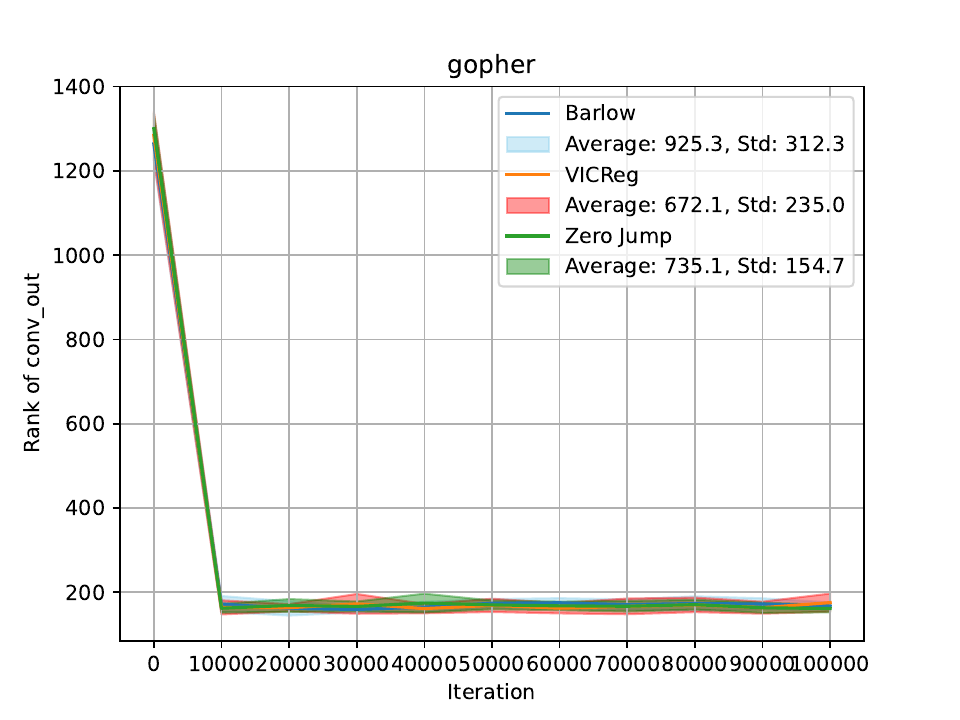}
        \label{fig:conv_out-sub14}
    \end{subfigure}
    \begin{subfigure}[b]{0.2\textwidth}
        \centering
        \includegraphics[width=\textwidth]{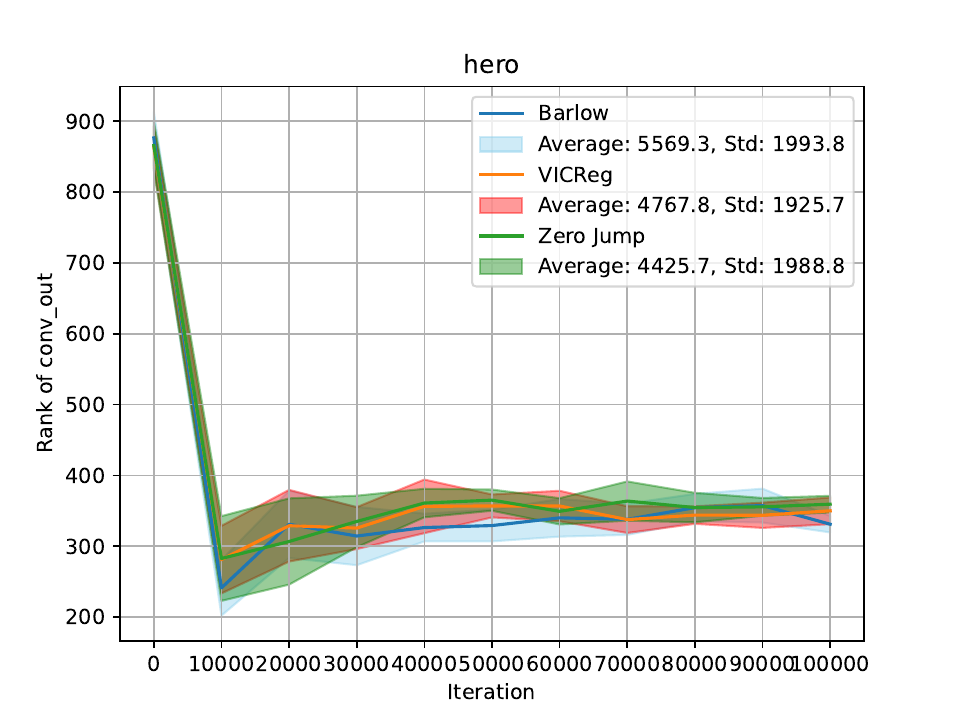}
        \label{fig:conv_out-sub15}
    \end{subfigure}
    \begin{subfigure}[b]{0.2\textwidth}
        \centering
        \includegraphics[width=\textwidth]{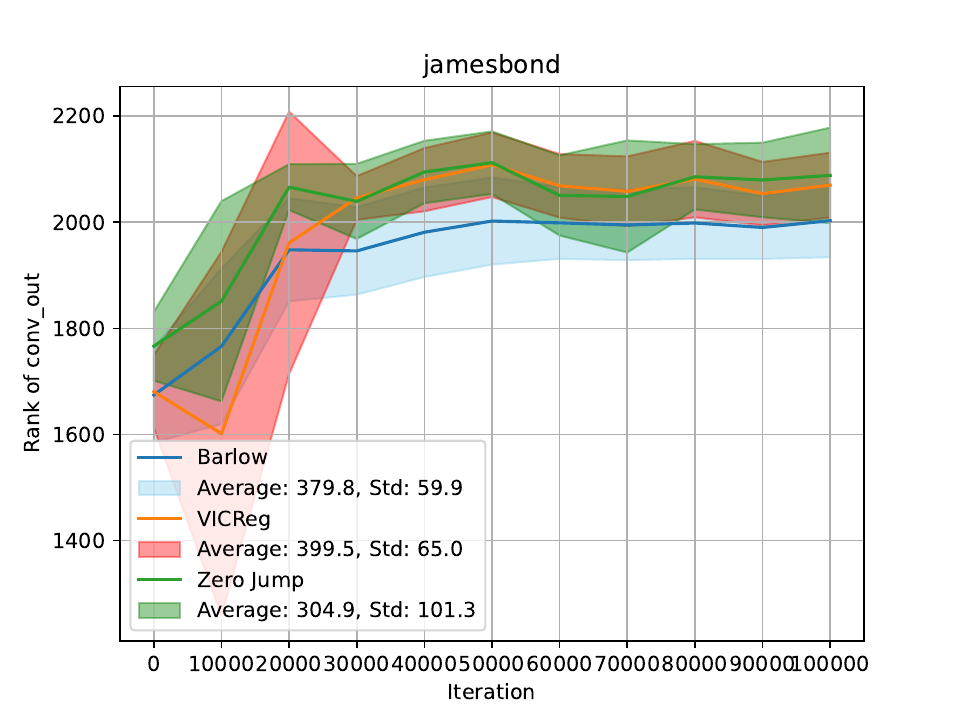}
        \label{fig:conv_out-sub16}
    \end{subfigure}

    \begin{subfigure}[b]{0.2\textwidth}
        \centering
        \includegraphics[width=\textwidth]{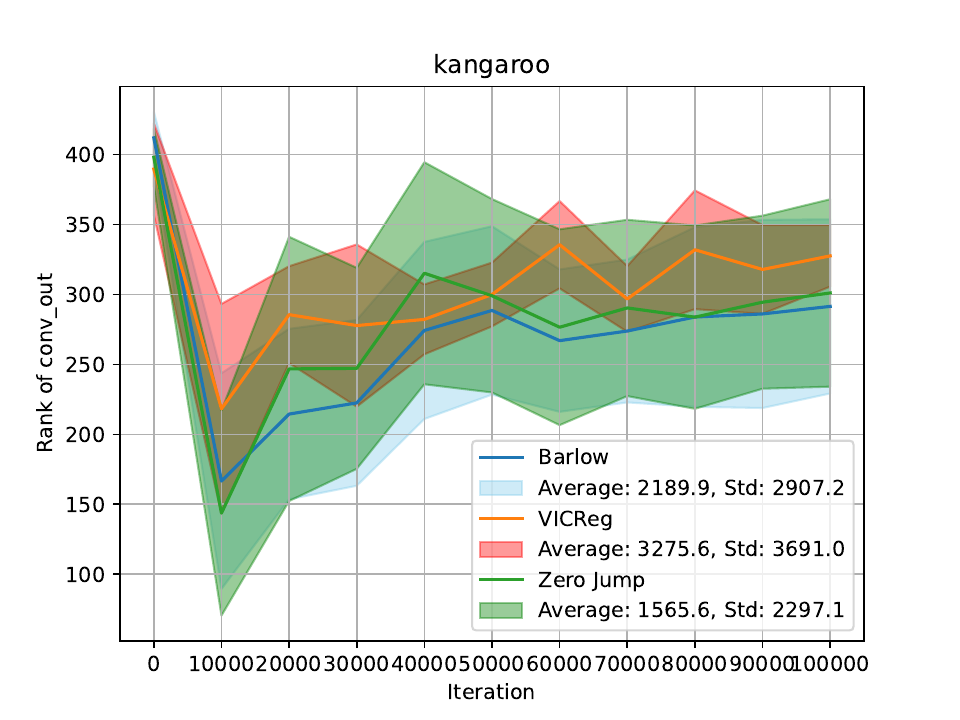}
        \label{fig:conv_out-sub17}
    \end{subfigure}
    \begin{subfigure}[b]{0.2\textwidth}
        \centering
        \includegraphics[width=\textwidth]{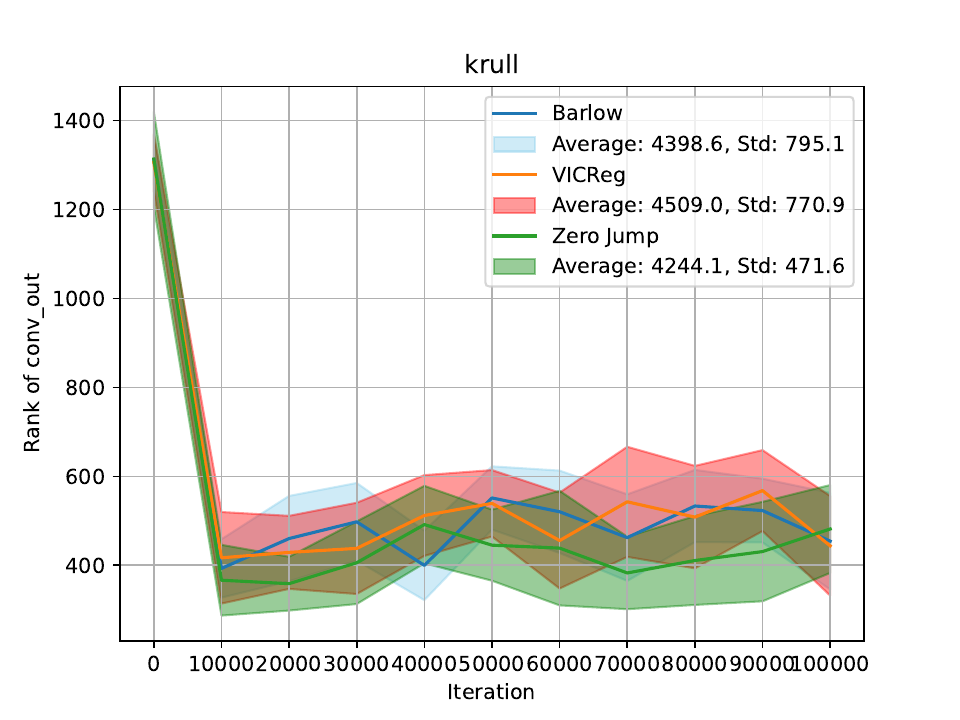}
        \label{fig:conv_out-sub18}
    \end{subfigure}
    \begin{subfigure}[b]{0.2\textwidth}
        \centering
        \includegraphics[width=\textwidth]{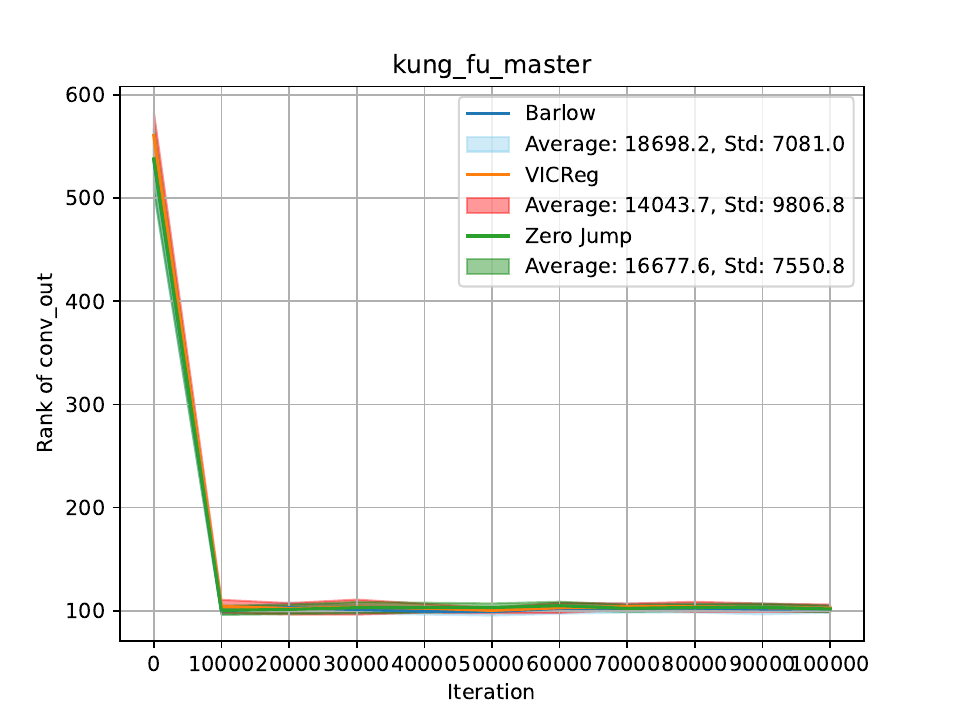}
        \label{fig:conv_out-sub19}
    \end{subfigure}
    \begin{subfigure}[b]{0.2\textwidth}
        \centering
        \includegraphics[width=\textwidth]{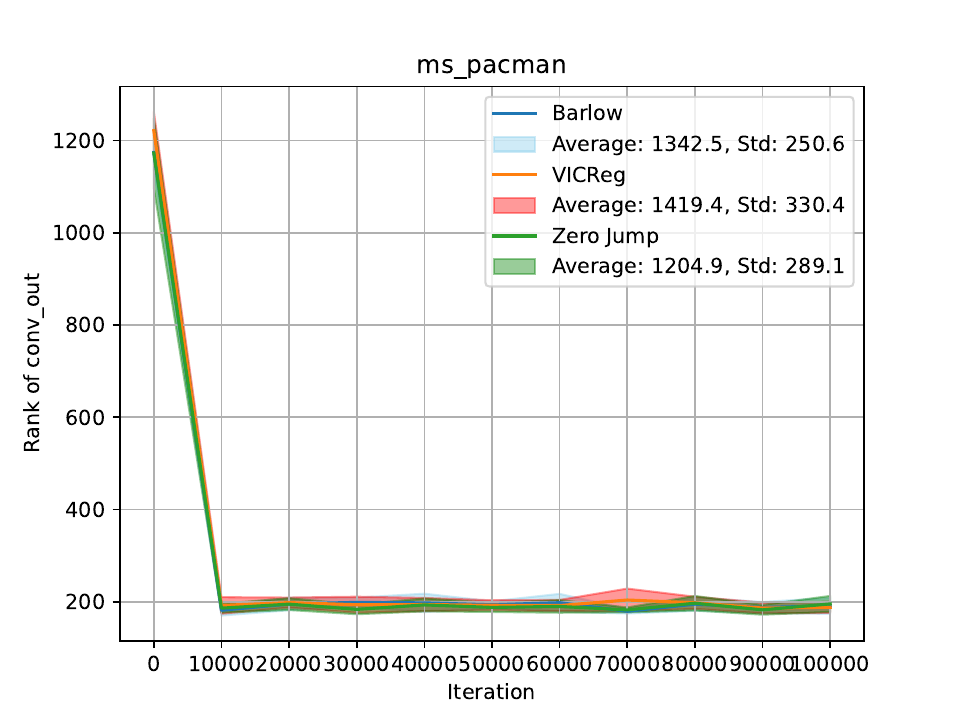}
        \label{fig:conv_out-sub20}
    \end{subfigure}

    \begin{subfigure}[b]{0.2\textwidth}
        \centering
        \includegraphics[width=\textwidth]{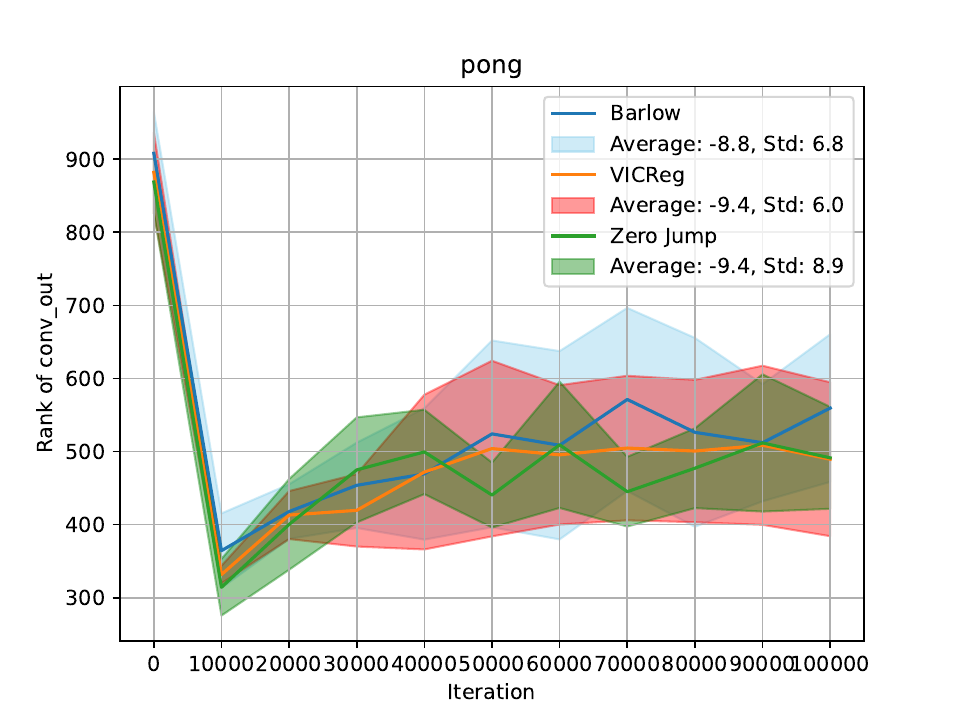}
        \label{fig:conv_out-sub21}
    \end{subfigure}
    \begin{subfigure}[b]{0.2\textwidth}
        \centering
        \includegraphics[width=\textwidth]{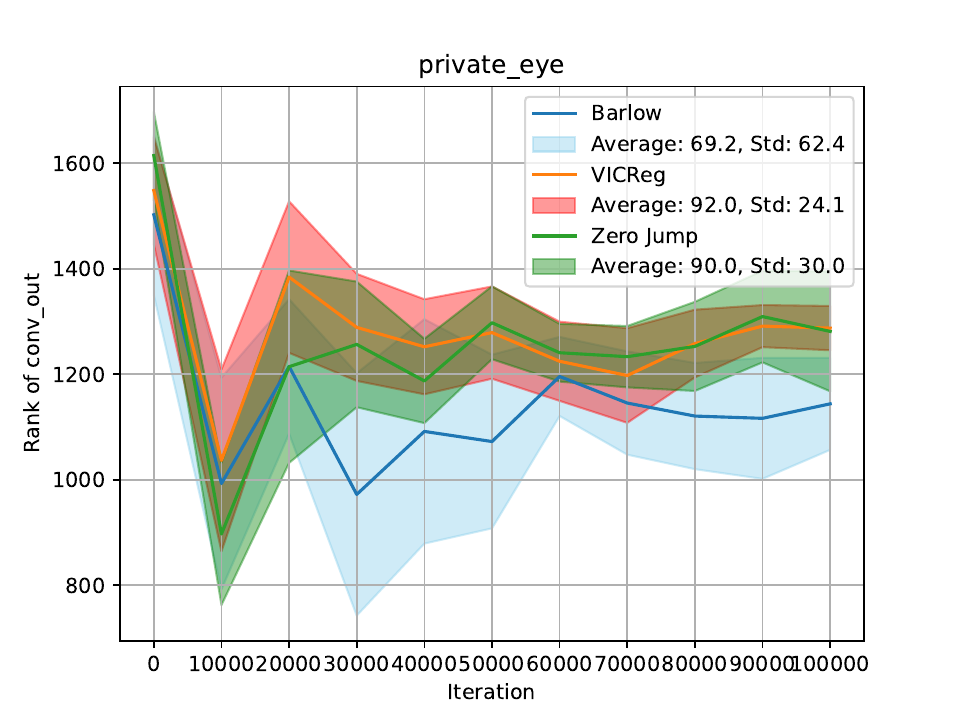}
        \label{fig:conv_out-sub22}
    \end{subfigure}
    \begin{subfigure}[b]{0.2\textwidth}
        \centering
        \includegraphics[width=\textwidth]{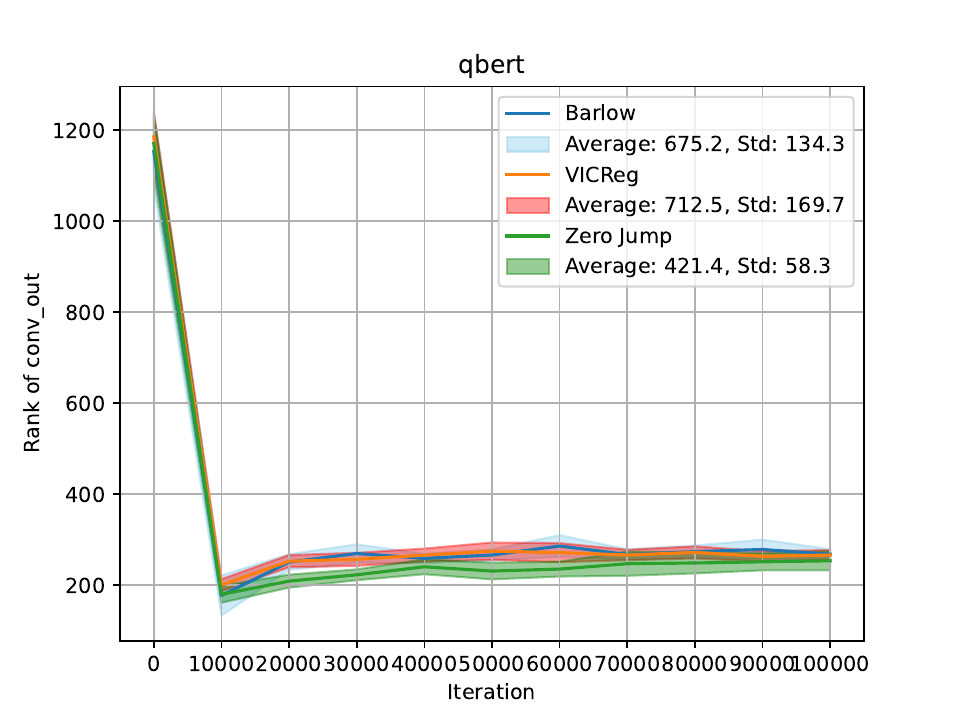}
        \label{fig:conv_out-sub23}
    \end{subfigure}
    \begin{subfigure}[b]{0.2\textwidth}
        \centering
        \includegraphics[width=\textwidth]{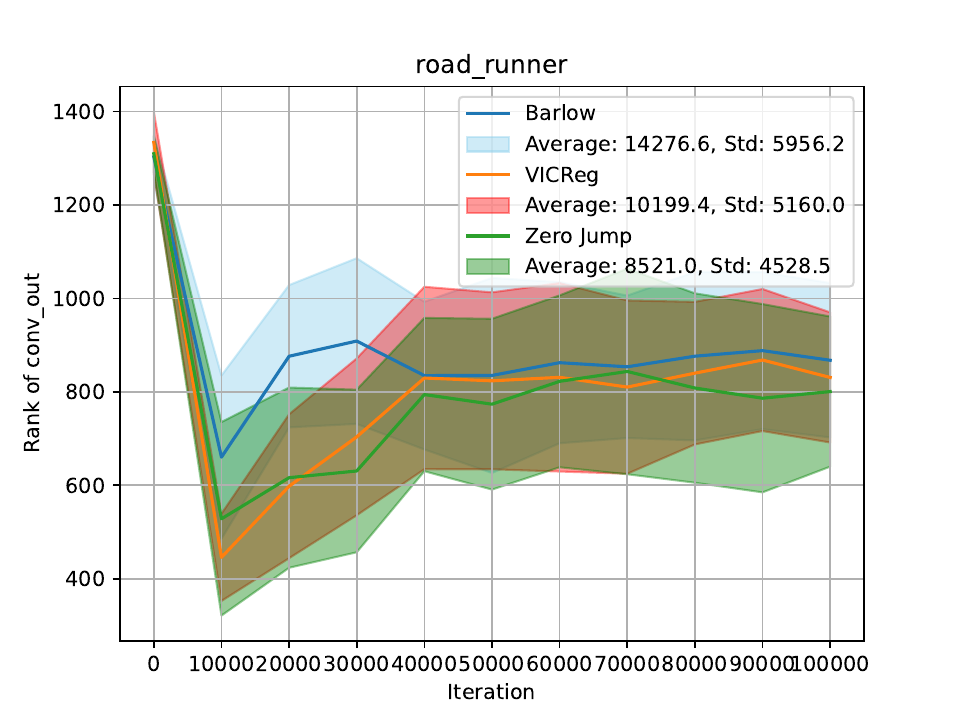}
        \label{fig:conv_out-sub24}
    \end{subfigure}

    \begin{subfigure}[b]{0.2\textwidth}
        \centering
        \includegraphics[width=\textwidth]{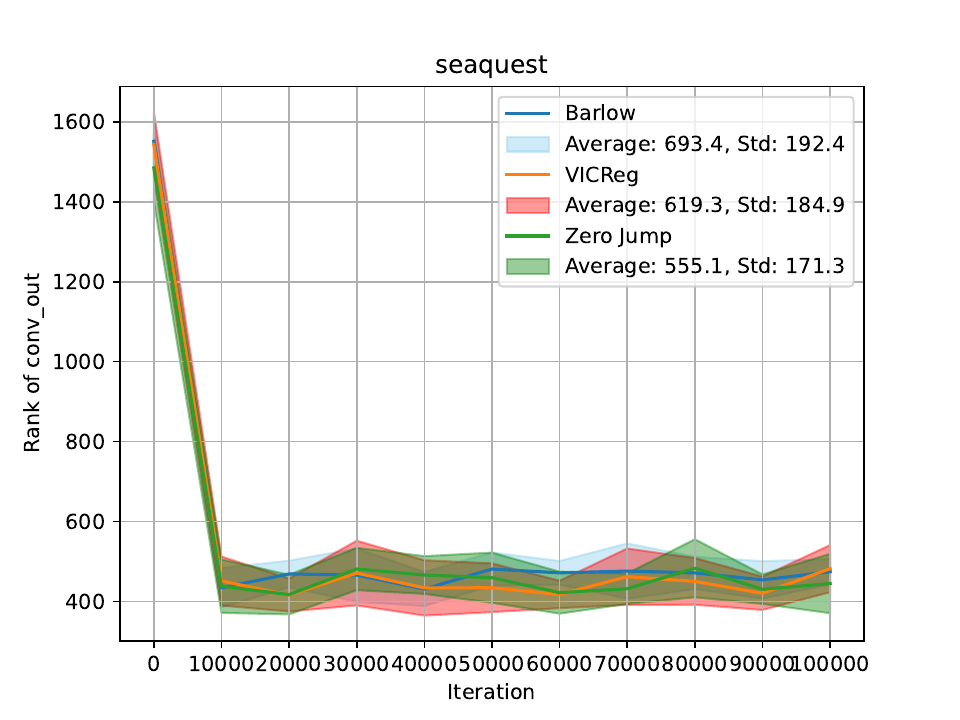}
        \label{fig:conv_out-sub25}
    \end{subfigure}
    \begin{subfigure}[b]{0.2\textwidth}
        \centering
        \includegraphics[width=\textwidth]{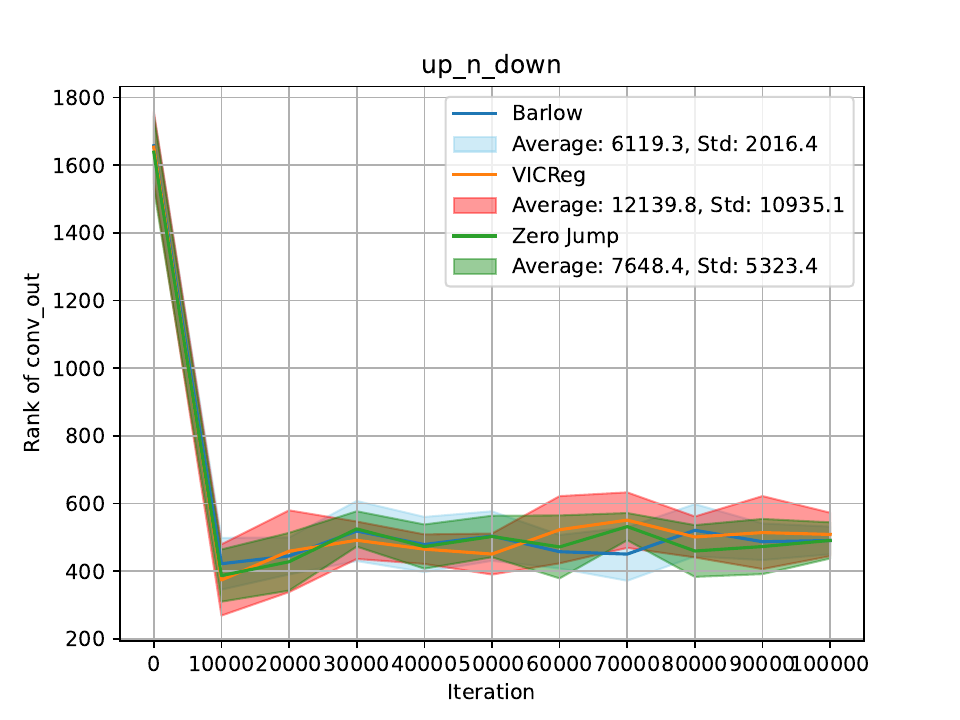}
        \label{fig:conv_out-sub26}
    \end{subfigure}

    \begin{subfigure}[b]{0.2\textwidth}
    \end{subfigure}
    \begin{subfigure}[b]{0.2\textwidth}
    \end{subfigure}

    \label{fig:conv_out-rank}
\end{figure*}

\begin{figure*}[htbp]
    \centering
    \caption{Rank of the output from the penultimate layer of the advantage head, measured every 10,000 steps and averaged across 10 different runs for every game.}
    \begin{subfigure}[b]{0.2\textwidth}
        \centering
        \includegraphics[width=\textwidth]{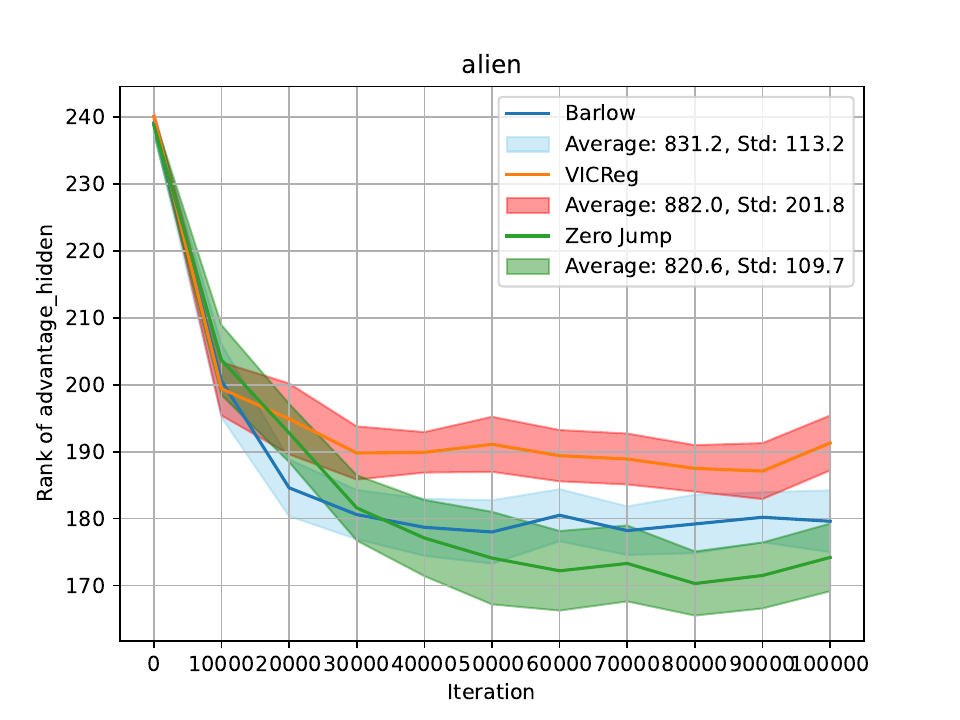}
        \label{fig:advantage_hidden-sub1}
    \end{subfigure}
    \begin{subfigure}[b]{0.2\textwidth}
        \centering
        \includegraphics[width=\textwidth]{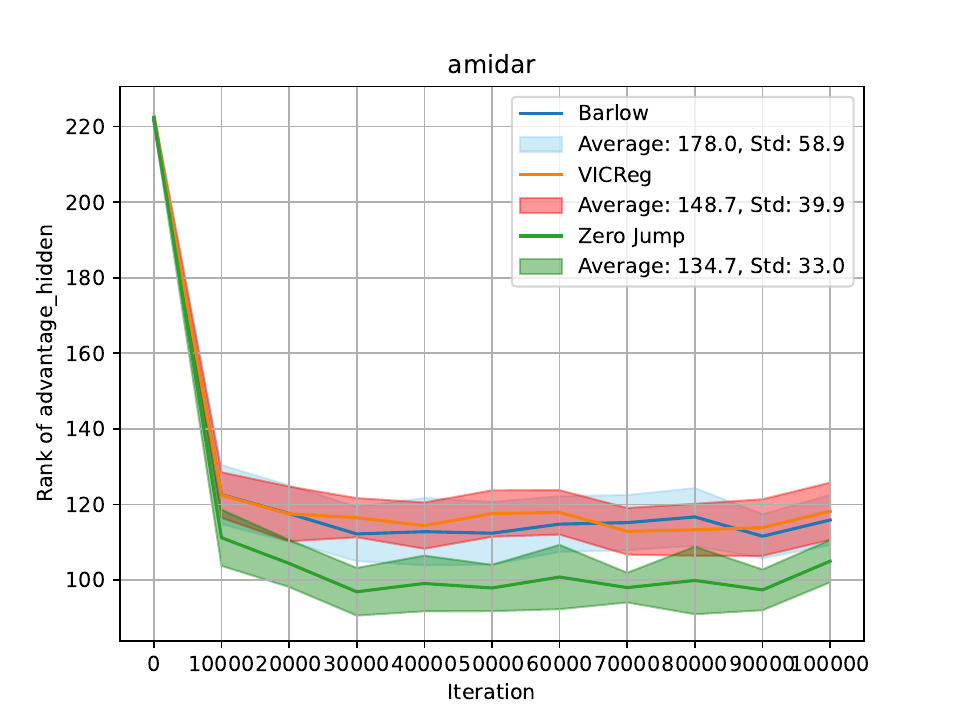}
        \label{fig:advantage_hidden-sub2}
    \end{subfigure}
    \begin{subfigure}[b]{0.2\textwidth}
        \centering
        \includegraphics[width=\textwidth]{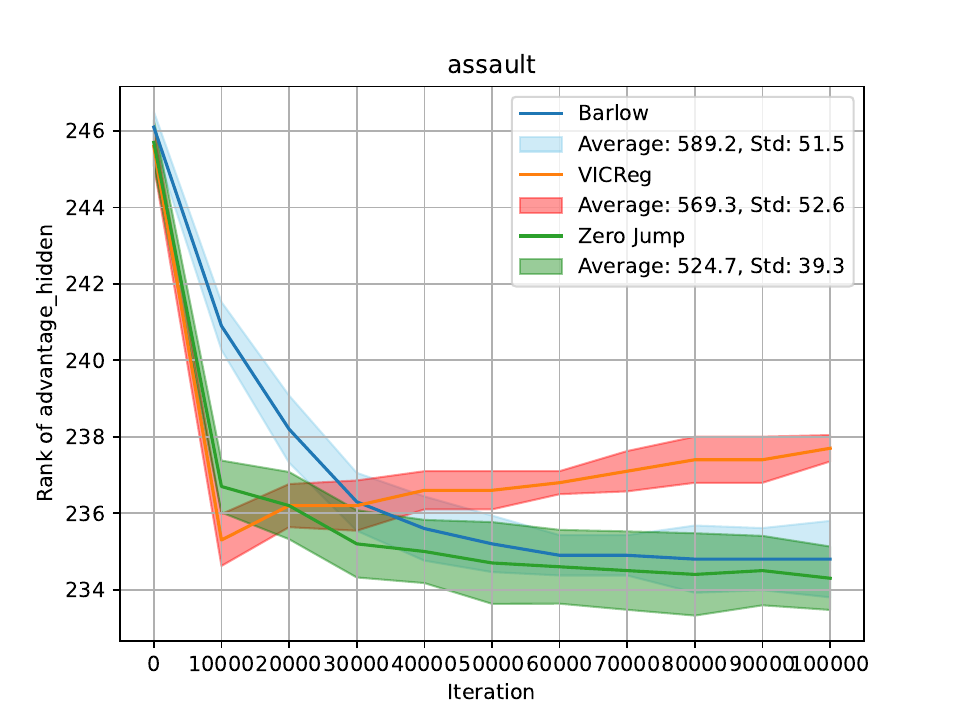}
        \label{fig:advantage_hidden-sub3}
    \end{subfigure}
    \begin{subfigure}[b]{0.2\textwidth}
        \centering
        \includegraphics[width=\textwidth]{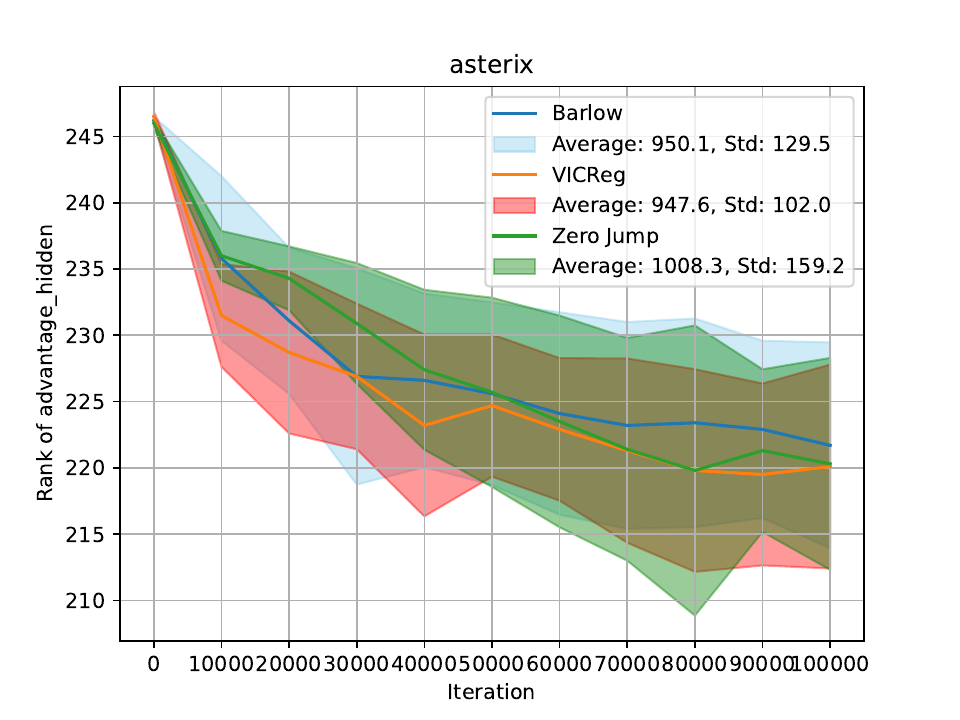}
        \label{fig:advantage_hidden-sub4}
    \end{subfigure}

    \begin{subfigure}[b]{0.2\textwidth}
        \centering
        \includegraphics[width=\textwidth]{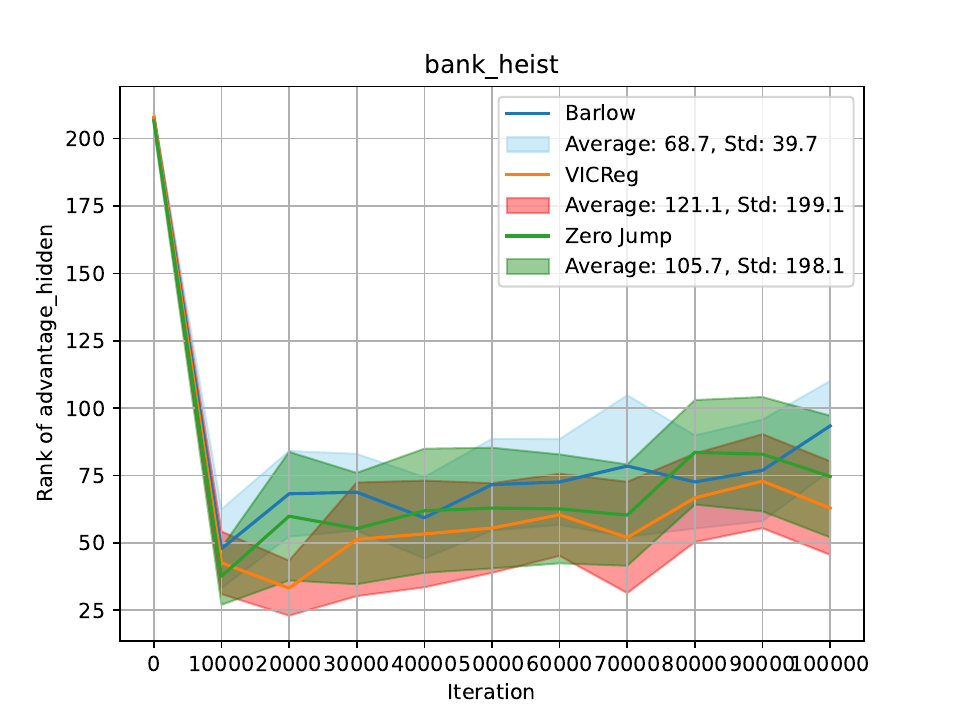}
        \label{fig:advantage_hidden-sub5}
    \end{subfigure}
    \begin{subfigure}[b]{0.2\textwidth}
        \centering
        \includegraphics[width=\textwidth]{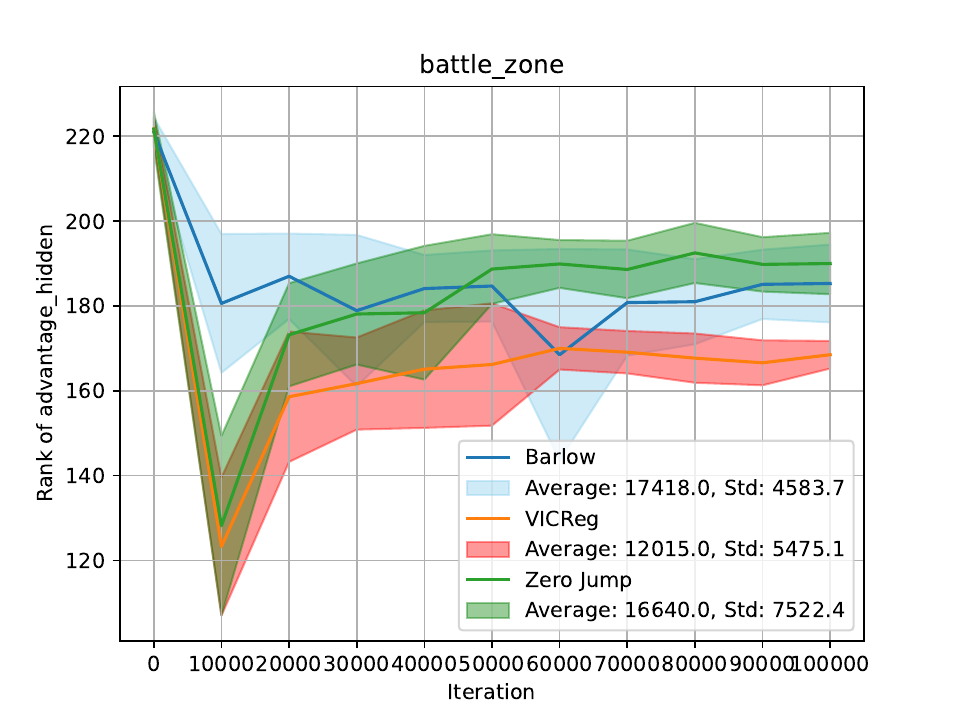}
        \label{fig:advantage_hidden-sub6}
    \end{subfigure}
    \begin{subfigure}[b]{0.2\textwidth}
        \centering
        \includegraphics[width=\textwidth]{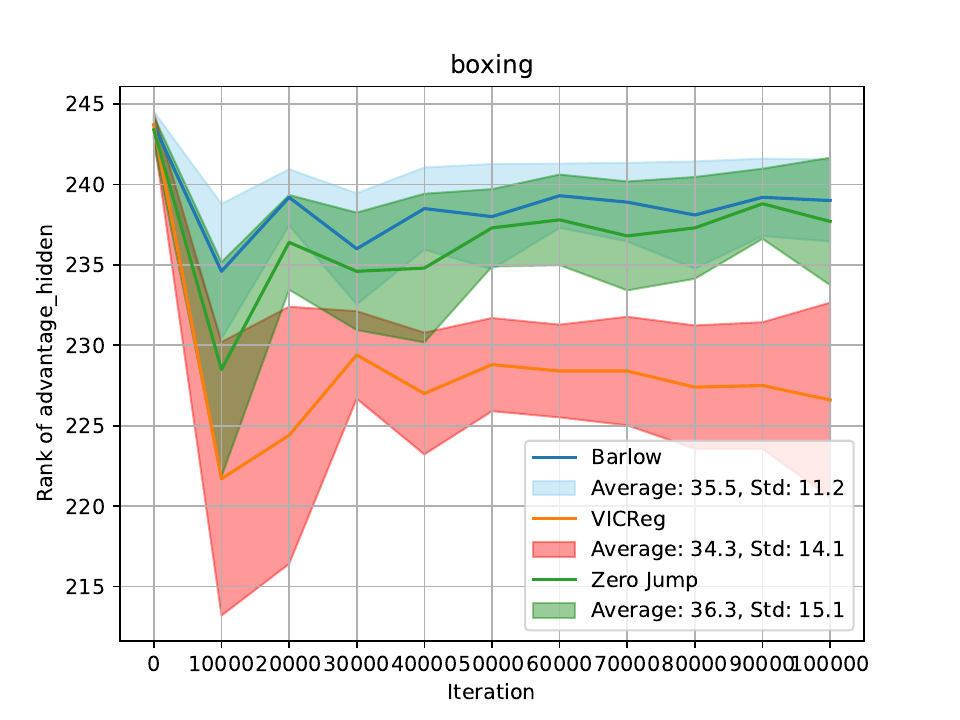}
        \label{fig:advantage_hidden-sub7}
    \end{subfigure}
    \begin{subfigure}[b]{0.2\textwidth}
        \centering
        \includegraphics[width=\textwidth]{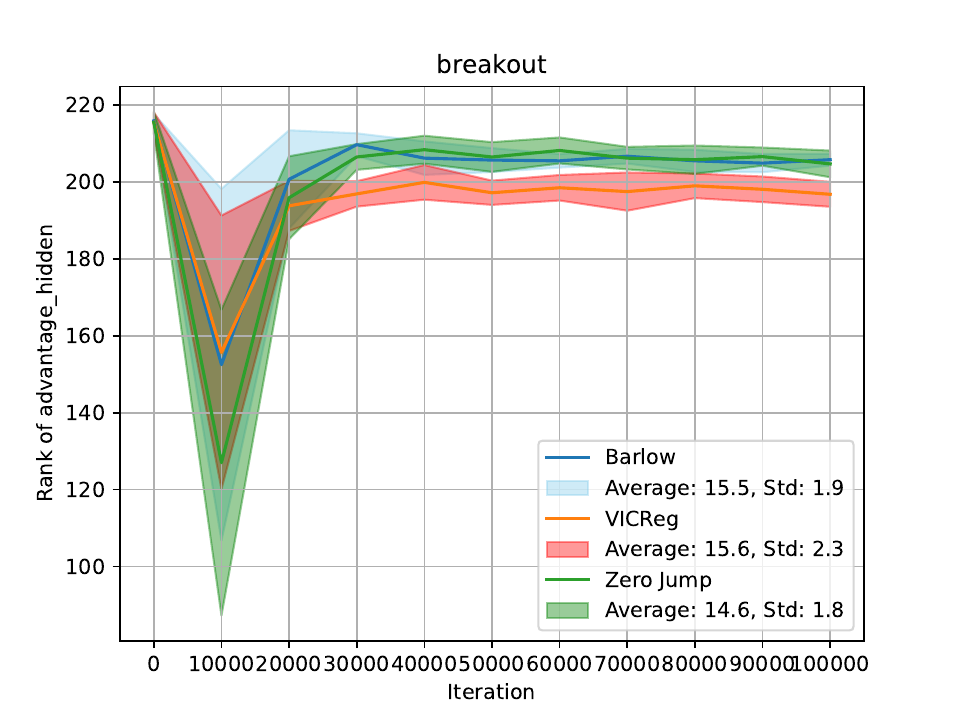}
        \label{fig:advantage_hidden-sub8}
    \end{subfigure}

    \begin{subfigure}[b]{0.2\textwidth}
        \centering
        \includegraphics[width=\textwidth]{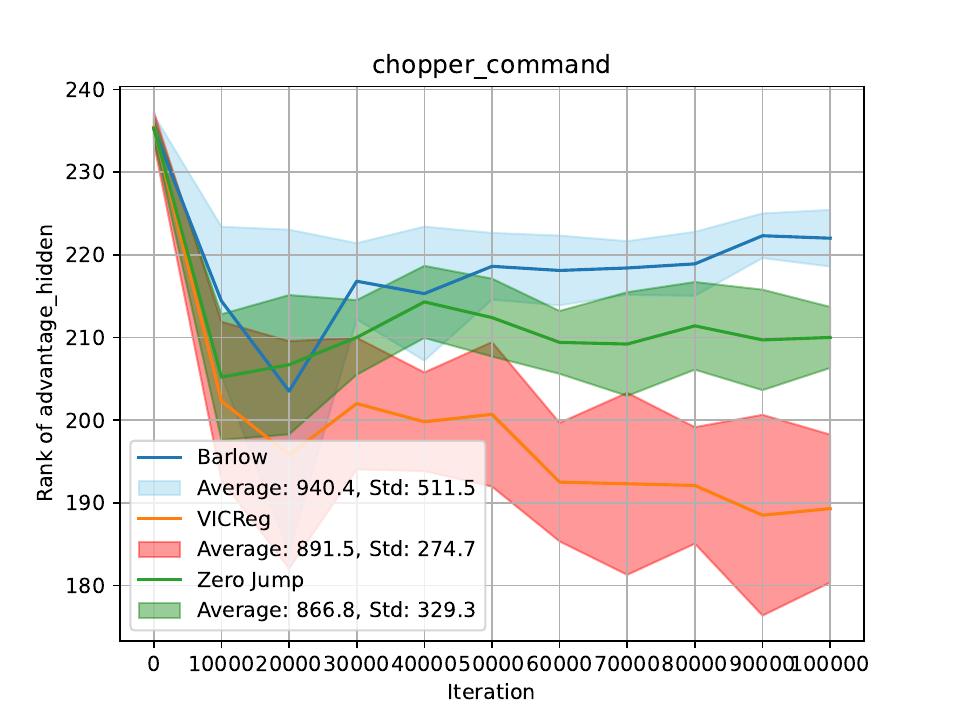}
        \label{fig:advantage_hidden-sub9}
    \end{subfigure}
    \begin{subfigure}[b]{0.2\textwidth}
        \centering
        \includegraphics[width=\textwidth]{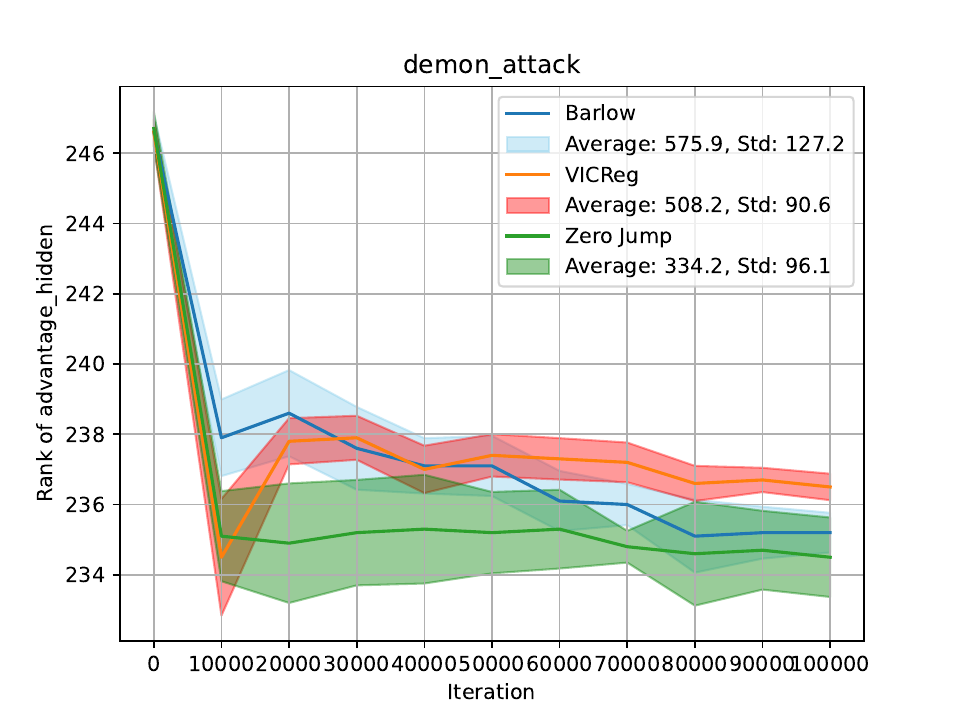}
        \label{fig:advantage_hidden-sub10}
    \end{subfigure}
    \begin{subfigure}[b]{0.2\textwidth}
        \centering
        \includegraphics[width=\textwidth]{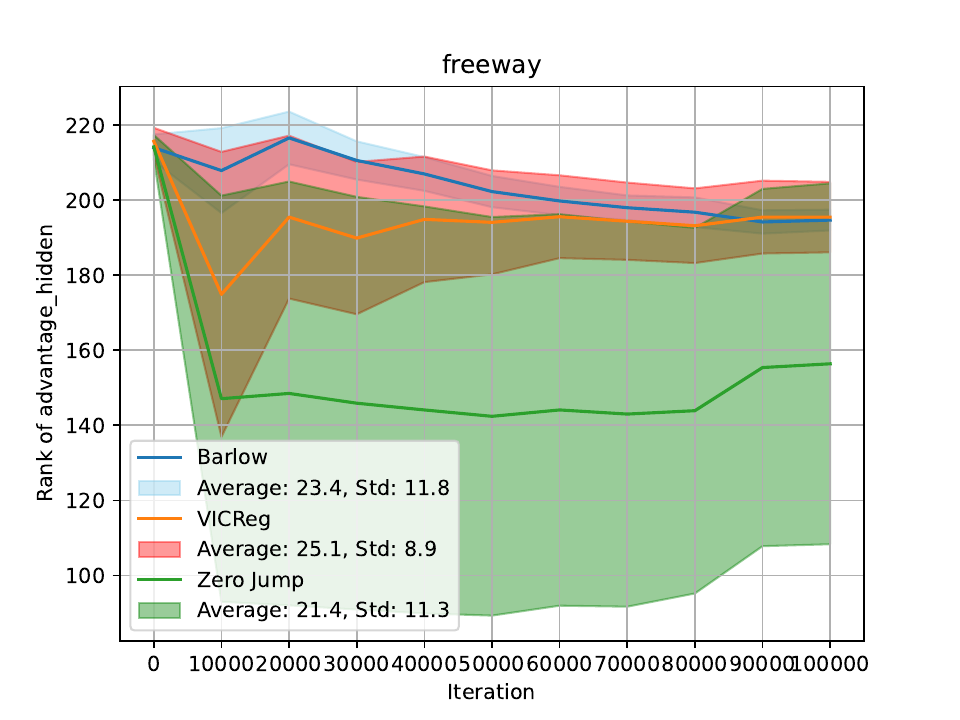}
        \label{fig:advantage_hidden-sub11}
    \end{subfigure}
    \begin{subfigure}[b]{0.2\textwidth}
        \centering
        \includegraphics[width=\textwidth]{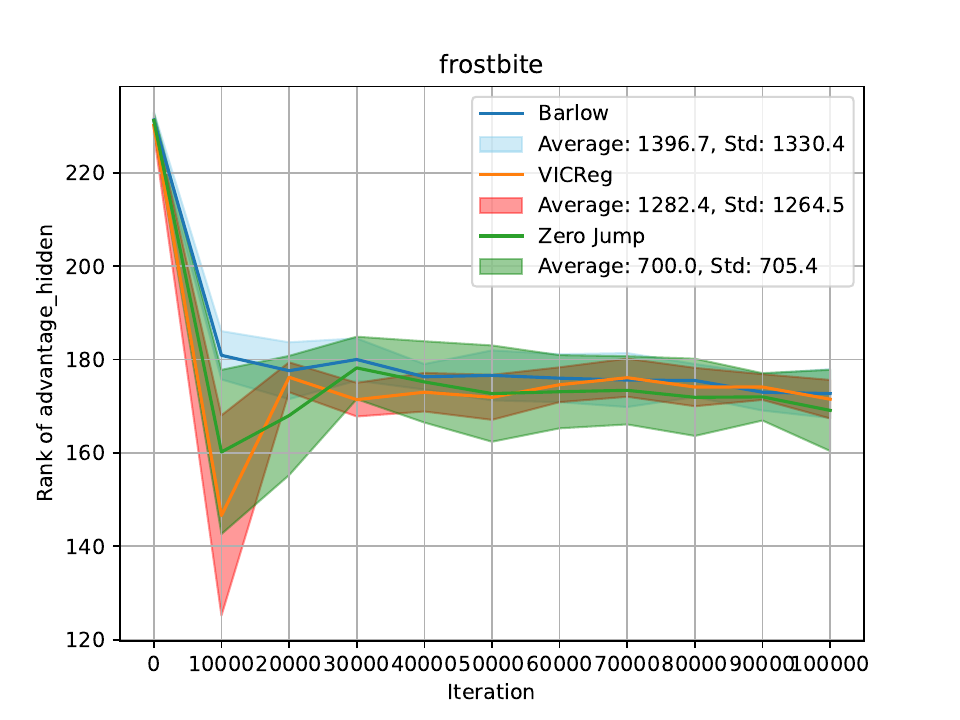}
        \label{fig:advantage_hidden-sub12}
    \end{subfigure}

    \begin{subfigure}[b]{0.2\textwidth}
        \centering
        \includegraphics[width=\textwidth]{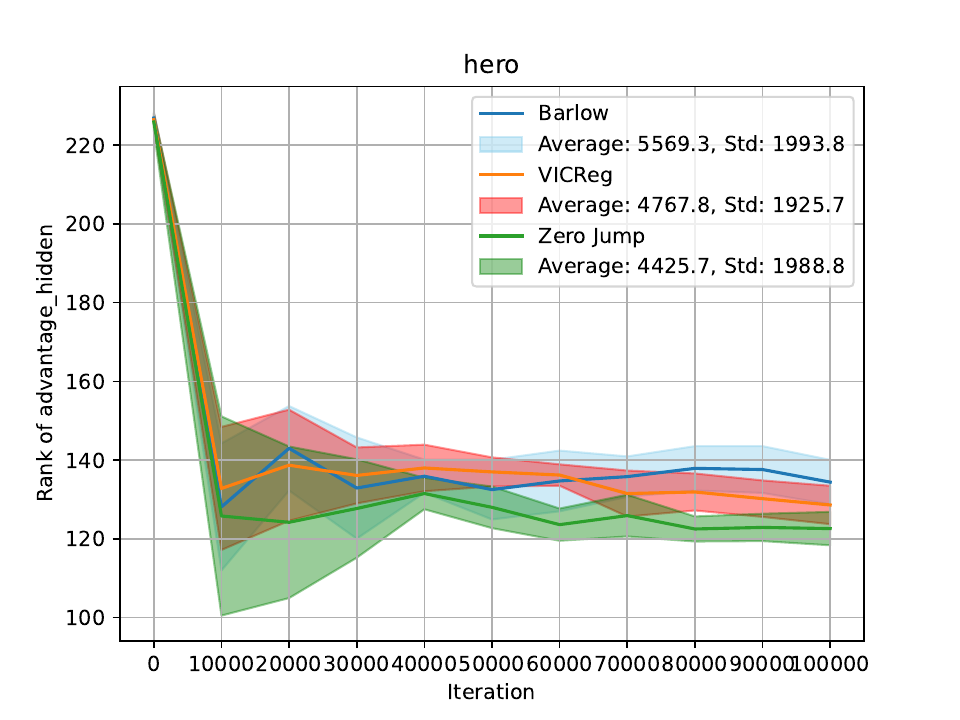}
        \label{fig:advantage_hidden-sub13}
    \end{subfigure}
    \begin{subfigure}[b]{0.2\textwidth}
        \centering
        \includegraphics[width=\textwidth]{figures/rank_figures/advantage_hidden/hero-advantage_hidden.pdf}
        \label{fig:advantage_hidden-sub14}
    \end{subfigure}
    \begin{subfigure}[b]{0.2\textwidth}
        \centering
        \includegraphics[width=\textwidth]{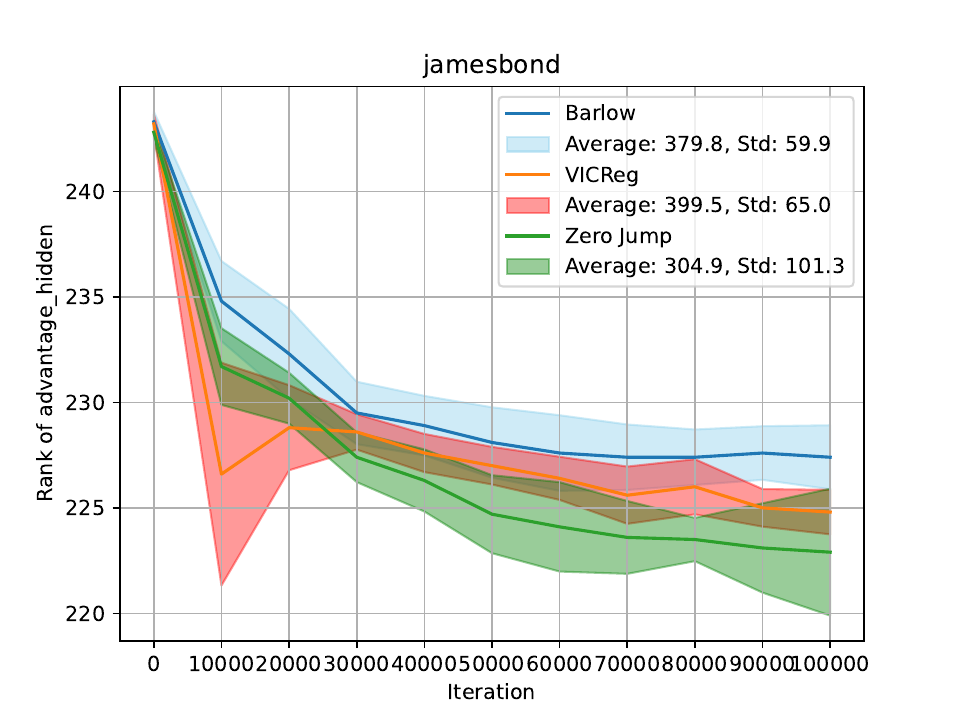}
        \label{fig:advantage_hidden-sub15}
    \end{subfigure}
    \begin{subfigure}[b]{0.2\textwidth}
        \centering
        \includegraphics[width=\textwidth]{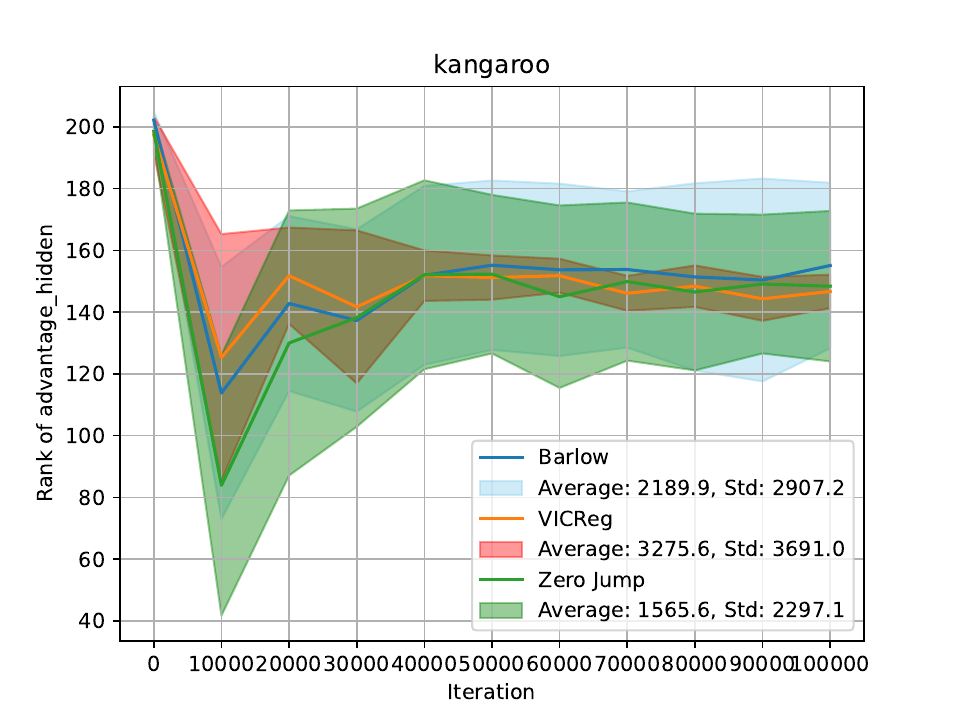}
        \label{fig:advantage_hidden-sub16}
    \end{subfigure}

    \begin{subfigure}[b]{0.2\textwidth}
        \centering
        \includegraphics[width=\textwidth]{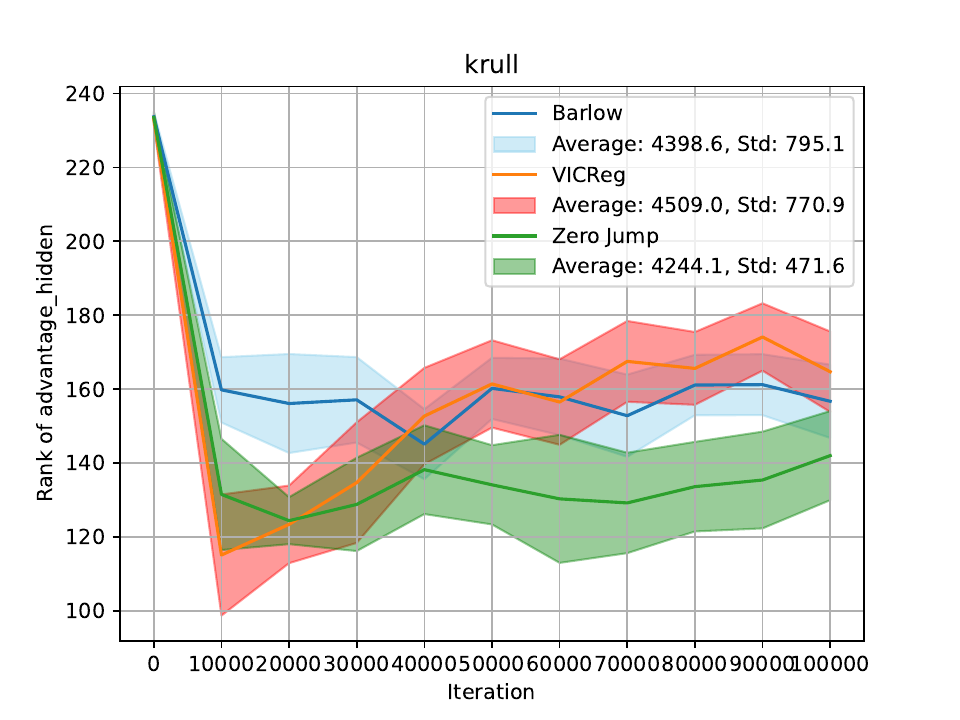}
        \label{fig:advantage_hidden-sub17}
    \end{subfigure}
    \begin{subfigure}[b]{0.2\textwidth}
        \centering
        \includegraphics[width=\textwidth]{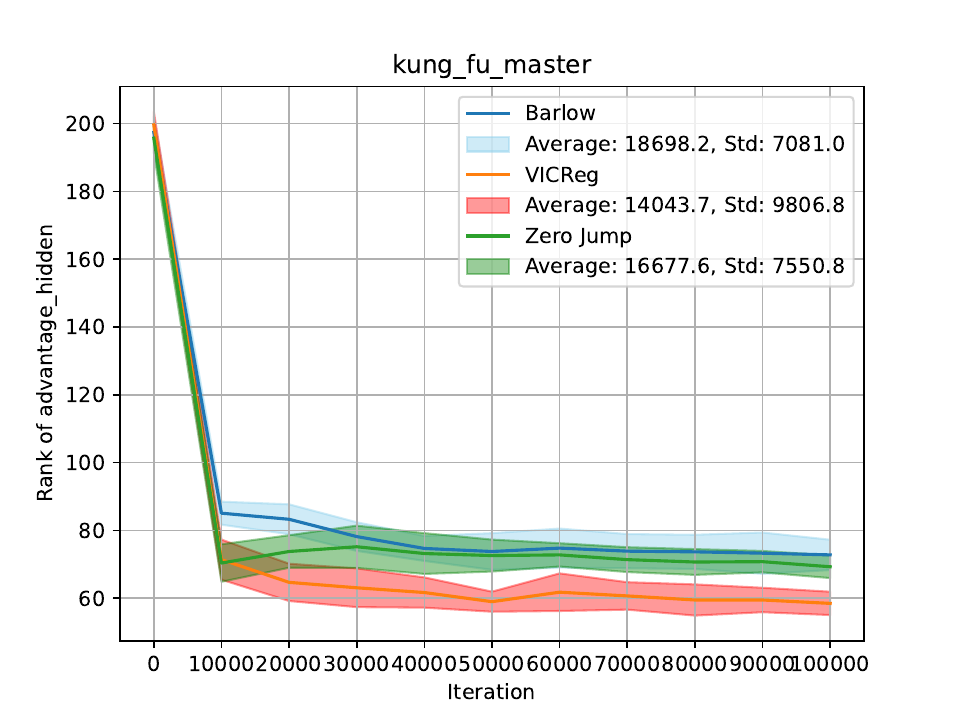}
        \label{fig:advantage_hidden-sub18}
    \end{subfigure}
    \begin{subfigure}[b]{0.2\textwidth}
        \centering
        \includegraphics[width=\textwidth]{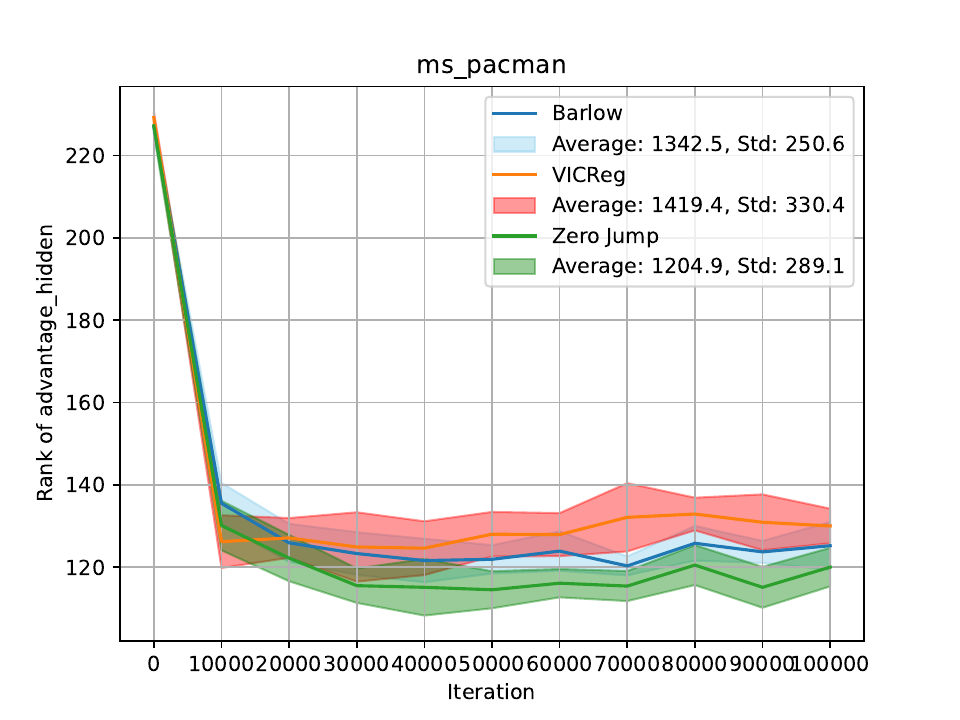}
        \label{fig:advantage_hidden-sub19}
    \end{subfigure}
    \begin{subfigure}[b]{0.2\textwidth}
        \centering
        \includegraphics[width=\textwidth]{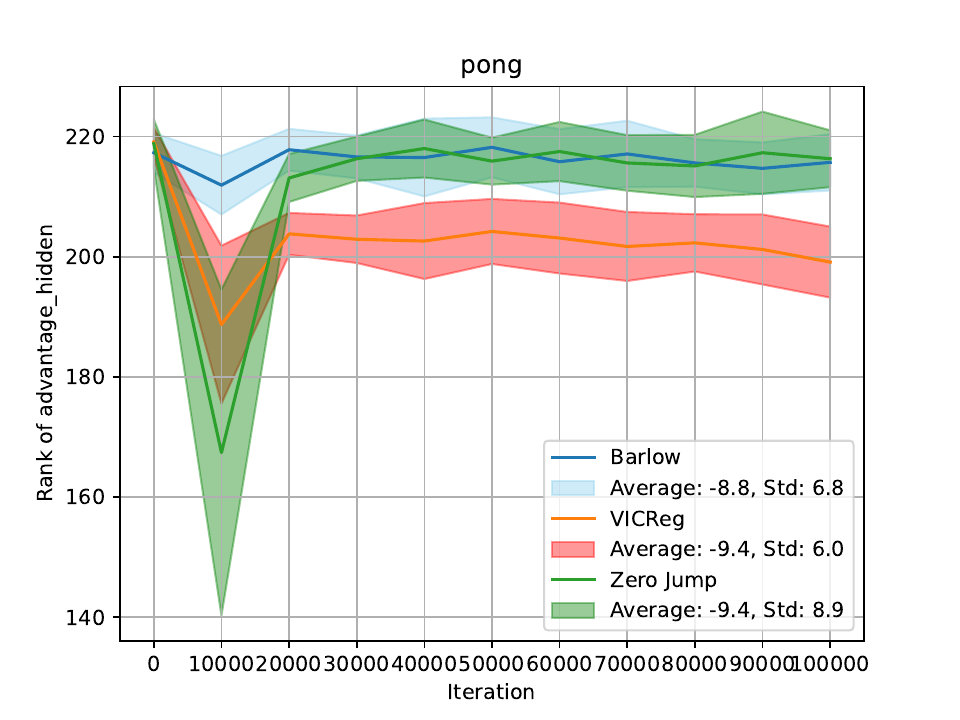}
        \label{fig:advantage_hidden-sub20}
    \end{subfigure}

    \begin{subfigure}[b]{0.2\textwidth}
        \centering
        \includegraphics[width=\textwidth]{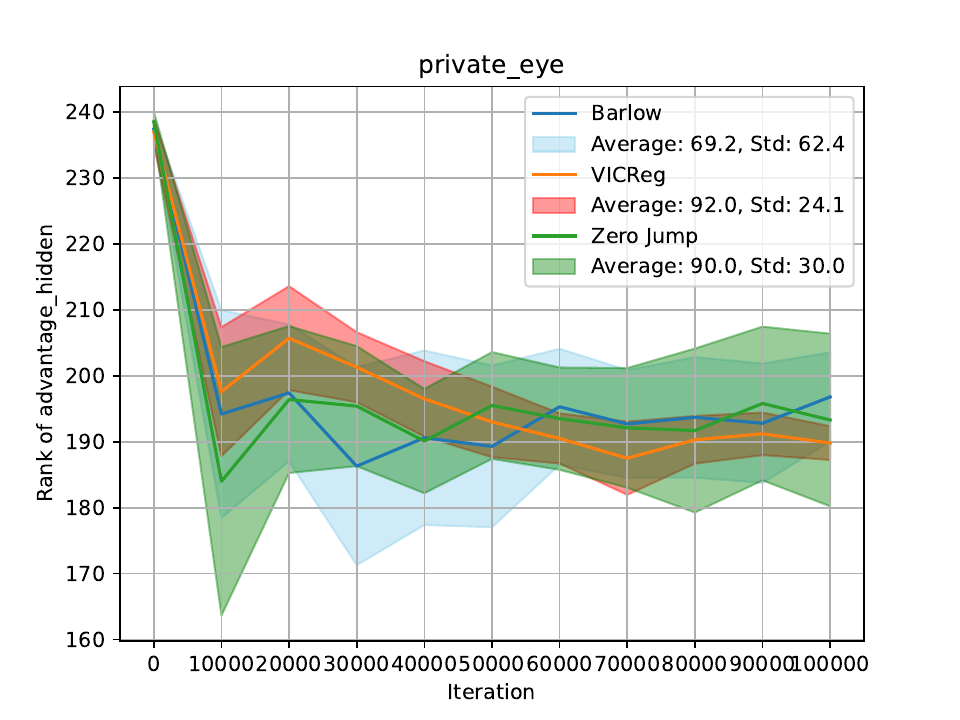}
        \label{fig:advantage_hidden-sub21}
    \end{subfigure}
    \begin{subfigure}[b]{0.2\textwidth}
        \centering
        \includegraphics[width=\textwidth]{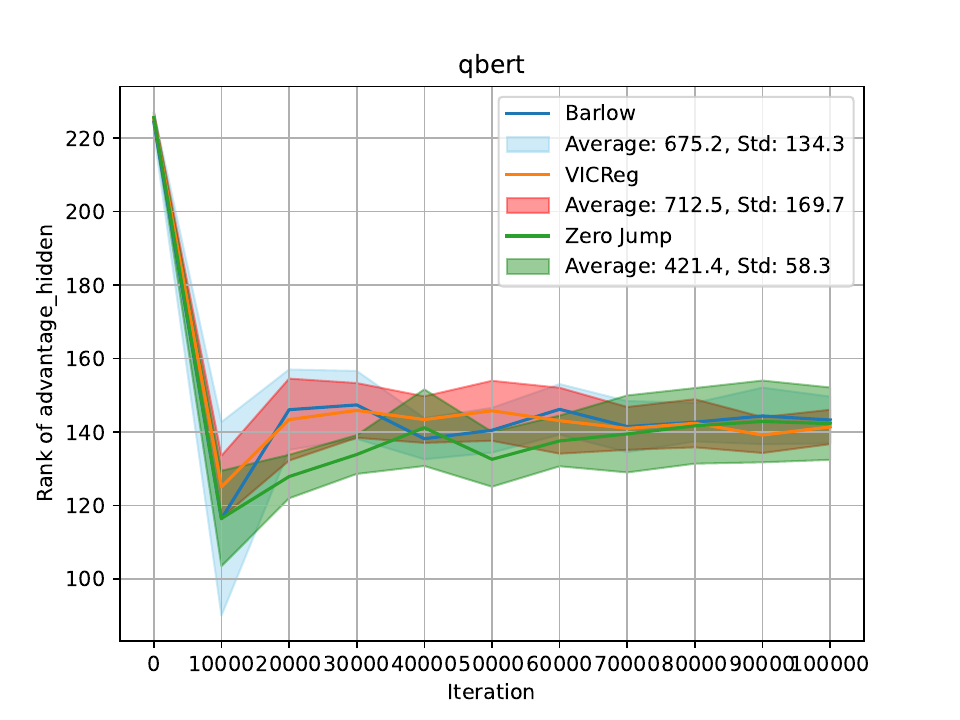}
        \label{fig:advantage_hidden-sub22}
    \end{subfigure}
    \begin{subfigure}[b]{0.2\textwidth}
        \centering
        \includegraphics[width=\textwidth]{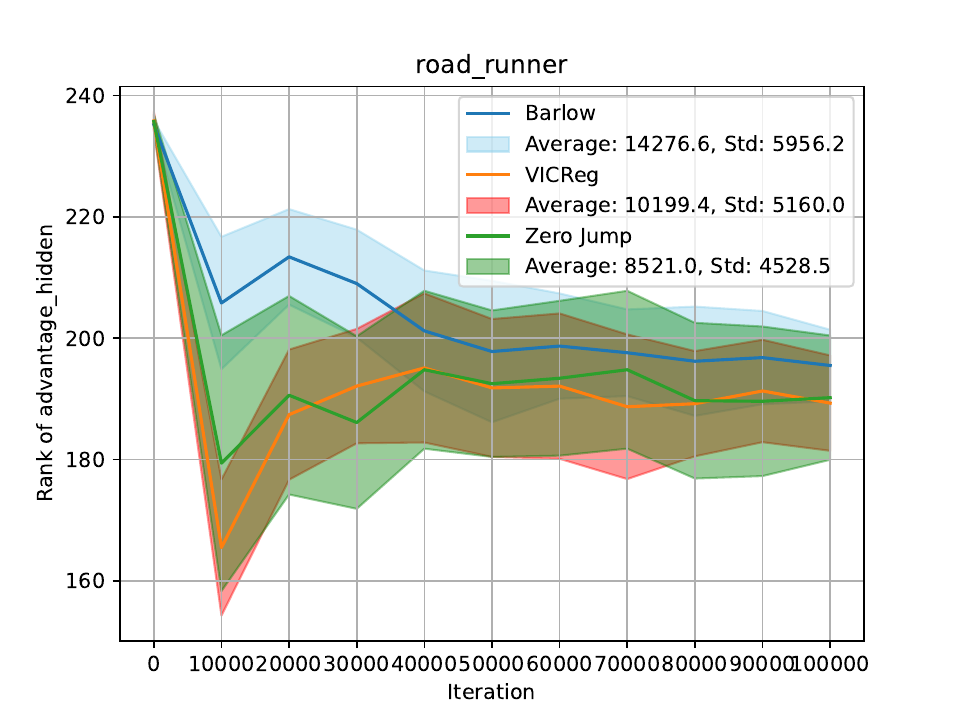}
        \label{fig:advantage_hidden-sub23}
    \end{subfigure}
    \begin{subfigure}[b]{0.2\textwidth}
        \centering
        \includegraphics[width=\textwidth]{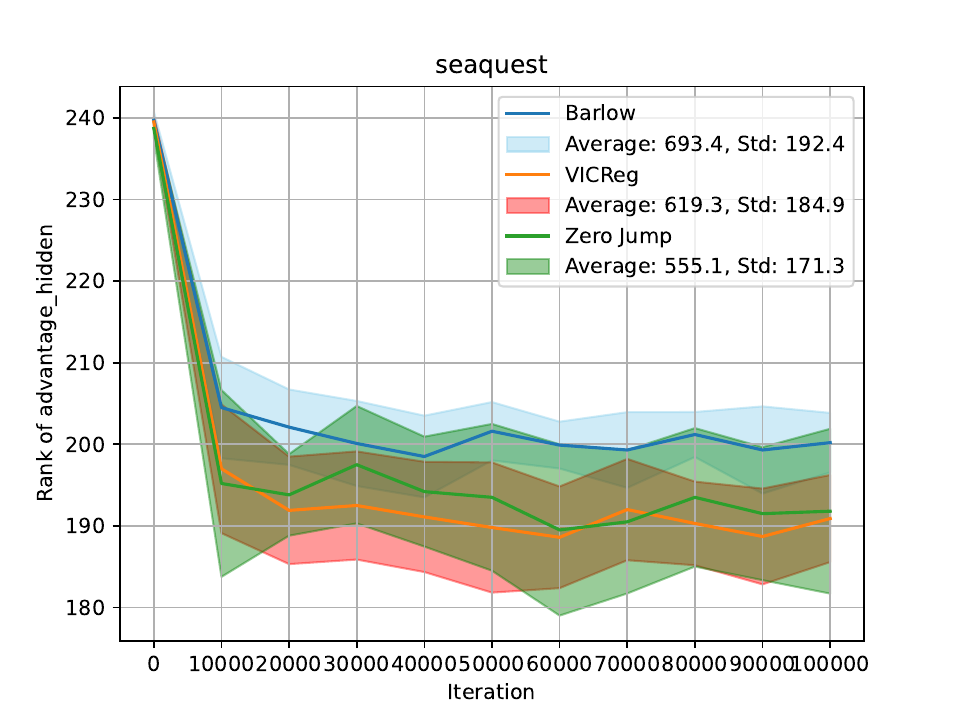}
        \label{fig:advantage_hidden-sub24}
    \end{subfigure}

    \begin{subfigure}[b]{0.2\textwidth}
        \centering
        \includegraphics[width=\textwidth]{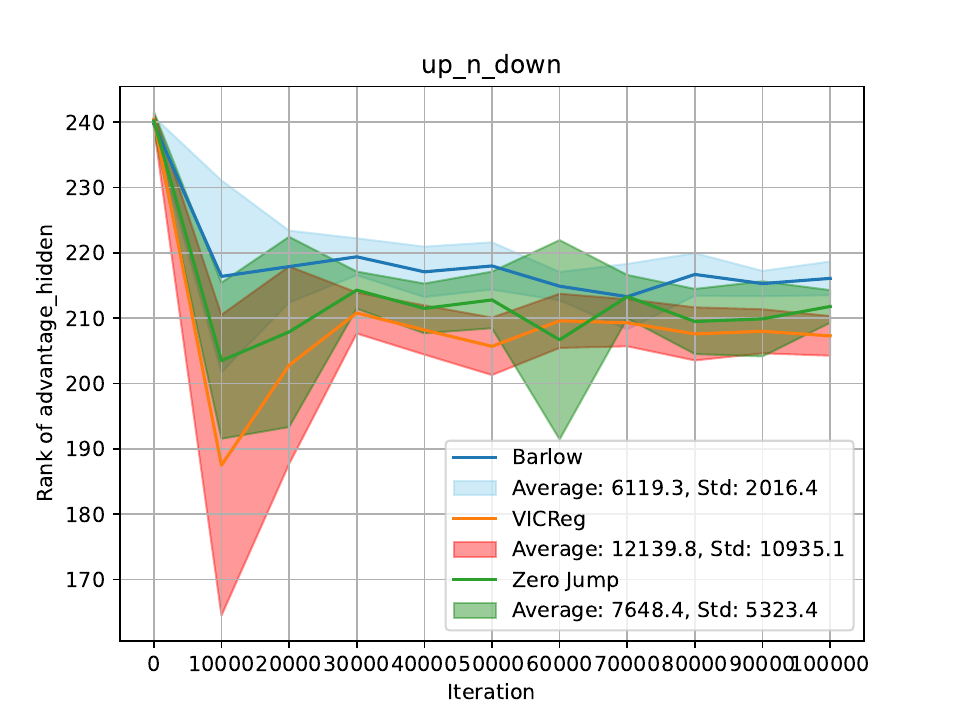}
        \label{fig:advantage_hidden-sub25}
    \end{subfigure}

    \begin{subfigure}[b]{0.2\textwidth}
    \end{subfigure}
    \begin{subfigure}[b]{0.2\textwidth}
    \end{subfigure}
    \begin{subfigure}[b]{0.2\textwidth}
    \end{subfigure}

    \label{fig:advantage_hidden-rank}
\end{figure*}

To investigate how SSL objectives might mitigate rank collapse, we computed the rank of the convolution output and the outputs of the penultimate layers from the advantage and value heads of three different agents: SPR-VICReg, SPR-Barlow, and ZeroJump (SPR without a transition model), scores in Table~\ref{tab:nonterm-noncontr}. Our observations indicate that, although there are some rank differences among the agents, they often converge to the same rank, and these differences do not correlate with the performance scores. Figure \ref{fig:val-rank}, \ref{fig:advantage_hidden-rank} and \ref{fig:conv_out-rank} include ranks across all games.

To explore this further, we examined the proportion of dormant neurons, which are neurons that have near-zero activations.~\citet{sokar2023dormantneuronphenomenondeep} showed that deep reinforcement learning agents experience a rise in the number of dormant neurons within their networks. Additionally, a higher prevalence of dormant neurons is associated with poorer performance.

\begin{figure*}[h]\label{fig:dormant-neuron}
    \centering
    \caption{Fraction of dormant neurons averaged across 10 different runs for every game.}

    \includegraphics[width=0.95\textwidth]{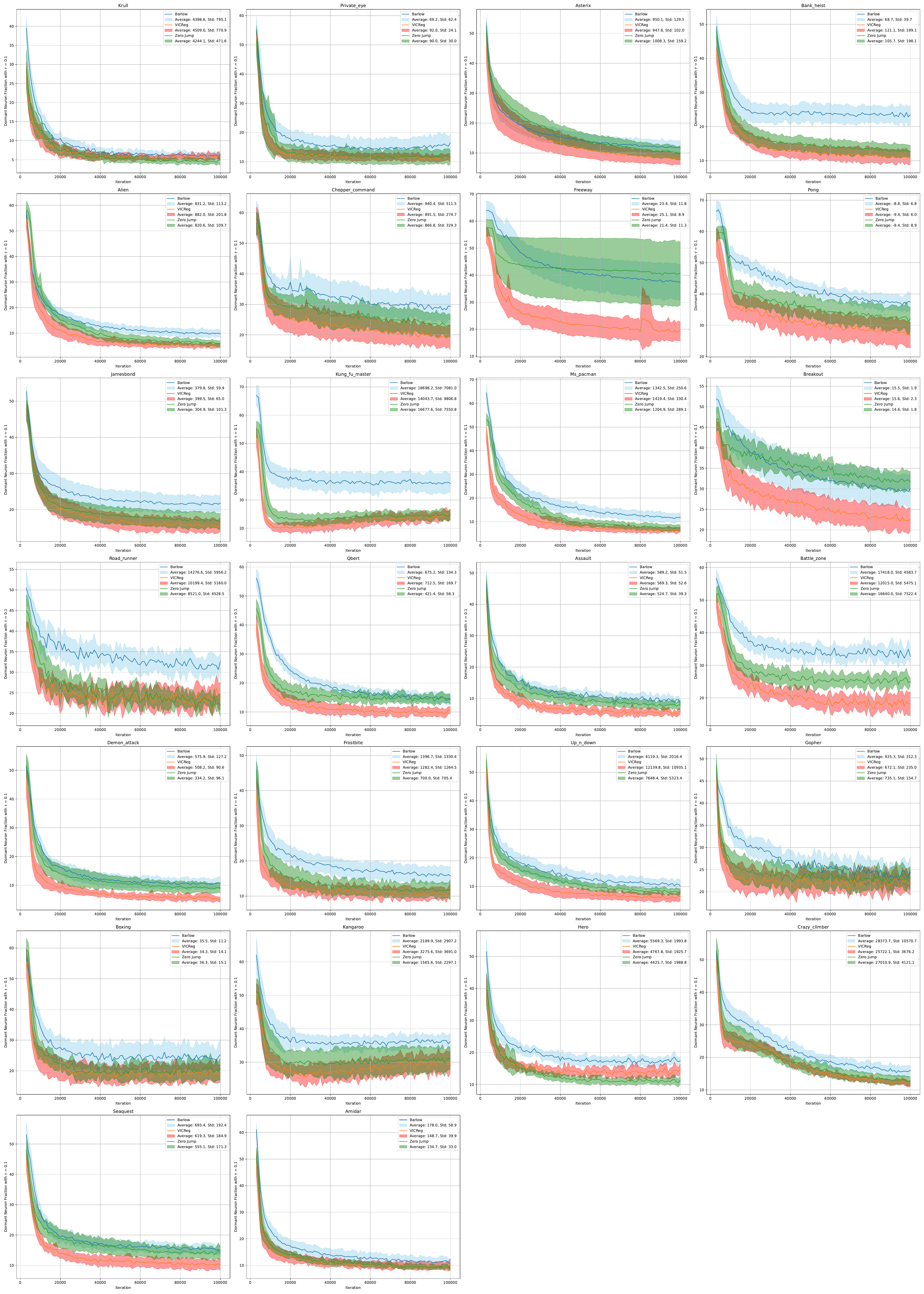} 
    \label{fig:dormant}
\end{figure*}

We also do not observe a clear pattern in the fractions of dormant neurons, in Figure \ref{fig:dormant} that could account for the disparities in performance scores, similar to what was seen in the case of neuron ranks. Unlike rank-based observations, where patterns may emerge, the distribution of dormant neurons does not offer an explanation for the differences in the scores across models. This suggests that the relationship between neuron activity and performance metrics might be more complex and not directly attributable to the proportion of inactive neurons.

\section{Experimental Details}
We retain all hyperparameters of SPR, SR-SPR, and BBF, except for SPR-Barlow and SPR-VICReg, where we adjust the SPR loss weight and increase the batch size from 32 to 64. The official repositories of the models are used, and all experiments are conducted on a Tesla T4 GPU.

\newpage
\section{Full Results on Atari 100k}\label{app:full-results}
\begin{table*}[h]
    \caption{Returns on the 26 games of Atari 100k after 2 hours of real-time experience, and human-normalized aggregate metrics. (VR: VICReg, results with 5 integral digits are rounded to the first integer to fit the table)}
    \label{tab:atari_results_full}
\begin{center}
\small
\begin{tabular}{lrrrrrrrrr}
\toprule
Game                 &  Rand.    &  Human          & Naked              & Non  & Prio          & VR-L    &VR-H      & Barlow     & SPR          \\
\midrule
Alien                &  227.8     &  7127.7         &  868.9             &  881.7      & 872.7              & 902.9         & 922.4            & 891.8      & 841.9             \\
Amidar               &  5.8       &  1719.5         &  165.6             &  179.1      & 164.2              & 181.1         & 176.4            & 177.1      & 179.7                     \\
Assault              &  222.4     &  742.0          &  544.5             &  564.6      & 589.2              & 536.4         & 575.7            & 581.4      & 565.6              \\
Asterix              &  210     &  8503.3         &  972.0             &  951.0      & 977.8              & 955.4         & 1021.7           & 981.2      & 962.5                    \\
BankHeist            &  14.2      &  753.1          &  61.6              &  70.1       & 60.2               & 79.9          & 82.9             & 73.5       & 345.4                      \\
BattleZone           &  2360    &  37188        &  7552.4            &  9424.2     & 13102            & 12557       & 14892            & 14954    & 14834                      \\
Boxing               &  0.1       &  12.1           &  27.3              &  30.4       & 36.4               & 31.3          & 33.9             & 35.1       & 35.7             \\
Breakout             &  1.7       &  30.5           &  16.7              &  18.0       & 18.2               & 16.9          & 16.3             & 17.0       & 19.6                   \\
ChopComm       &  811     &  7387.8         &  906.8             &  949.8      & 901.0              & 832.9         & 929.9            & 938.9      & 946.3          \\
CrzyClmbr         &  10781   &  35829        & 30056            &  32667    & 35829            & 27035       & 29023          & 29229    & 36701             \\
DemonAtt          &  152.1     &  1971.0         & 514.7              &  511.0      & 522.9              & 461.2         & 547.2            & 519.2      & 517.6                \\
Freeway              &  0.0       &  29.6           &  17.4              &  13.71      & 16.3               & 28.0          & 27.7             & 29.5       & 19.3            \\
Frostbite            &  65.2      &  4334.7         &  1137.2            &  1010.9     & 1014.2             & 1353.0        & 1181.4           & 1191.3     & 1170.7                \\
Gopher               &  257.6     &  2412.5         &  585.0             &  660.1      & 548.4              &  737.9        & 713.5            & 691.2      & 660.6       \\
Hero                 &  1027    &  30826        &  6937.8            &  6497.8     & 5686.6             &  5495.1       & 5559.6           & 5746.8     & 5858.6              \\
Jamesbond            &  29      &  302.8          &  327.2             &  359.9      & 349.1              &  357.6        & 384.3            & 404.2      & 366.5      \\
Kangaroo             &  52      &  3035.0         &  2970.9            &  2812.1     & 3016.5             &  2290.6       & 1998.3           & 1771.2     & 3617.4            \\
Krull                &  1598    &  2665.5         &  3980.4            &  4061.8     & 4213.1             &  4166.6       & 4513.9           & 4363.2     & 3681.6             \\
KFMaster         &  258.5     &  22736        &  13126           &  14595    & 15757            &  1488.4       & 15548          & 15998    & 14783            \\
MsPacman             &  307.3     &  6951.6         &  1262.1            &  1162.6     & 1324.6             &  1366.8       & 1588.2           & 1388.2     & 1318.4         \\
Pong                 &  -20.7     &  14.6           &  -1.8              &  -6.0       & -7.2               &  -6.3         & -10.1            & -6.7       & -5.4       \\
PrivateEye           &  24.9      &  69571        &  85.6              &  77.0       & 88.0               &  100.9        & 96.6             & 99.6       &  86.0     \\
Qbert                &  163.9     &  13455        &  847.2             &  758.6      & 759.8              &  796.9        & 687.6            & 765.8      &  866.3         \\
RoadRunner           &  11.5      &  7845.0         &  12595           &  12713    & 11211            &  10683      & 9531.5          & 12412    &  12213           \\
Seaquest             &  68.4      &  42055        &  524.0             &  524.2      & 523.2              &  576.3        & 651.0            & 669.1      &  558.1        \\
UpNDown              &  533.4     &  11693        &  9569.3            &  8130.6     & 10331            & 7952.7        & 9415.3           & 10818    &  10859            \\
\midrule
\#Sprhmn(↑)     &  0         &  N/A            &  4                 &  3          & 3                  &  4            & 4                &    4         &  6                   \\
Mean (↑)             &  0.00     &  1.000          &  0.542             &  0.555      & 0.608              &  0.558        & 0.585            & 0.608        &  0.616               \\
Median (↑)           &  0.00     &  1.000          &  0.225             &  0.221      & 0.308              &  0.297        & 0.280            & 0.312        &  0.396                    \\
IQM (↑)              &  0.00     &  1.000          &  0.273             &  0.278      & 0.298              &  0.292        & 0.298            & 0.321        &  0.337                \\
Opt. Gap (↓)   &  1.00     &  0.000          &  0.617             &  0.615      & 0.603              &  0.609        & 0.605            & 0.587        &  0.577                \\
\bottomrule
\end{tabular}
 
\end{center}
\end{table*}



\newpage
\section{Full Results on DMControl 100k}\label{app:contcontrol}

\begin{table*} [h]
    \caption{Returns on the of DMControl 100k, and Max-normalized aggregate metrics.}
    \label{tab:dmc_results_full}
\begin{center}
\small
\begin{tabular}{lrr rrrrrrr}
\toprule
Environment                 & Virtual    &  VICReg+Pred     & Barlow+Pred       & SPR         & VICReg           & Barlow          \\
\midrule
FINGER, SPIN                &  896.2     &  760.6         &  781.0             &  755.9      & 730.0              & 861.8                     \\
CARTPOLE, SWINGUP           &  815.1     &  791.6         &  784.0             &  826.0      & 780.1              & 778.6                              \\
REACHER, EASY               &  827.0     &  790.7         &  589.6             &  671.5      & 736.1              & 526.5                      \\
CHEETAH, RUN                &  489.6     &  504.3         &  461.6             &  435.2      & 493.5              & 478.6                            \\
WALKER, WALK                &  404.7     &  622.8         &  521.7             &  404.7      & 765.               & 182.2                              \\
BALL IN CUP, CATCH          &  835.4     &  891.6         &  622.8             &  835.4      & 937.5              & 924.9                            \\
\midrule
Mean (↑)                    &  0.705     &  0.738          &  0.673            &  0.660      & 0.740              &  0.625                    \\
Median (↑)                  &  0.803     &  0.772          &  0.703            &  0.726      & 0.750              &  0.652                      \\
IQM (↑)                     &  0.744     &  0.773          &  0.670            &  0.677      & 0.750              &  0.656                     \\
Optimality Gap (↓)          &  0.294     &  0.260          &  0.326            &  0.339      & 0.29              &  0.374                     \\
\bottomrule
\end{tabular}
 
\end{center}
\end{table*}

\newpage
\section{Rank and Dormant Neuron Results}

\newpage

\end{document}